\pdfoutput=1

\documentclass[11pt]{article}

\usepackage[preprint]{acl}

\usepackage{times}
\usepackage{latexsym}

\usepackage[T1]{fontenc}

\usepackage[utf8]{inputenc}

\usepackage{microtype}

\usepackage{inconsolata}

\usepackage{makecell}
\usepackage{booktabs}
\usepackage{graphicx}
\usepackage{soul} 
\usepackage{subcaption}
\usepackage{multirow}
\usepackage{colortbl}
\usepackage{pgf}
\usepackage{float}
\usepackage{amsmath}
\usepackage{setspace}
\usepackage{pifont}
\usepackage[normalem]{ulem}

\definecolor{color0}{rgb}{1,1,1}
\definecolor{color5}{rgb}{0.5,0.7,0.8}

\newcommand{\ccv}[1]{
  \pgfmathsetmacro{\opacity}{#1/5}
  \pgfmathsetmacro{\mixfactor}{\opacity*200}
  \edef\temp{\noexpand\cellcolor{color5!\mixfactor!color0}}
  \temp #1
}
\definecolor{myred}{HTML}{FFC8C8}
\definecolor{myblue}{HTML}{C8E7FF}
\definecolor{myblue2}{HTML}{2E9ADF}

\title{Are LLM-based Evaluators Confusing NLG Quality Criteria?}

\author{Xinyu Hu$^{*,1}$, Mingqi Gao$^{*,1}$, Sen Hu$^{2}$, Yang Zhang$^{2}$ \\ {\bf Yicheng Chen}$^{2}$, {\bf Teng Xu}$^{2}$, {\bf Xiaojun Wan$^{1}$}\\
$^{1}$Wangxuan Institute of Computer Technology, Peking University\\
$^{2}$Ant Group\\
\{huxinyu,gaomingqi,wanxiaojun\}@pku.edu.cn\\ \{hs272483,yaoling.zy,yicheng.chen,harvey.xt\}@antgroup.com}

\begin{document}
\maketitle
\def\thefootnote{*}\footnotetext{Equal contribution.}\def\thefootnote{\arabic{footnote}}

\begin{abstract}

Some prior work has shown that LLMs perform well in NLG evaluation for different tasks. However, we discover that LLMs seem to confuse different evaluation criteria, which reduces their reliability. For further verification, we first consider avoiding issues of inconsistent conceptualization and vague expression in existing NLG quality criteria themselves. So we summarize a clear hierarchical classification system for 11 common aspects with corresponding different criteria from previous studies involved. Inspired by behavioral testing, we elaborately design 18 types of aspect-targeted perturbation attacks for fine-grained analysis of the evaluation behaviors of different LLMs. We also conduct human annotations beyond the guidance of the classification system to validate the impact of the perturbations. Our experimental results reveal confusion issues inherent in LLMs, as well as other noteworthy phenomena, and necessitate further research and improvements for LLM-based evaluation.
\end{abstract}

\section{Introduction}

With the emergence of powerful large language models (LLMs) such as ChatGPT, LLM-based evaluators have been widely used for various natural language generation (NLG) tasks~\citep{DBLP:conf/acl/ChiangL23,DBLP:conf/eamt/KocmiF23,DBLP:journals/corr/abs-2303-15621,DBLP:journals/corr/abs-2304-02554}. In evaluation for common NLG tasks such as summarization~\citep{DBLP:journals/tacl/FabbriKMXSR21}, dialogue~\citep{DBLP:conf/acl/MehriE20}, and story generation~\citep{DBLP:conf/inlg/XieCL23}, different aspects of quality (such as fluency and faithfulness) should be considered individually. Traditional evaluation metrics are either incapable of evaluating specific aspects, like BLEU~\cite{DBLP:conf/acl/PapineniRWZ02} and BERTScore~\citep{DBLP:conf/iclr/ZhangKWWA20}, or they can only roughly assess a single aspect, like FactCC \cite{DBLP:conf/emnlp/KryscinskiMXS20}. In contrast, LLMs can be treated as akin to human annotators, with various definitions of aspects contained in the prompt for flexible evaluation. Some studies~\citep{DBLP:journals/corr/abs-2303-04048,DBLP:journals/corr/abs-2308-16797,DBLP:conf/emnlp/LiuIXWXZ23} have shown LLM-based evaluators have achieved comparable performance with humans in many NLG tasks, suggesting LLMs to become promising candidates for automatic evaluation.

\begin{figure}[t]
\centering
\includegraphics[width=0.48\textwidth]{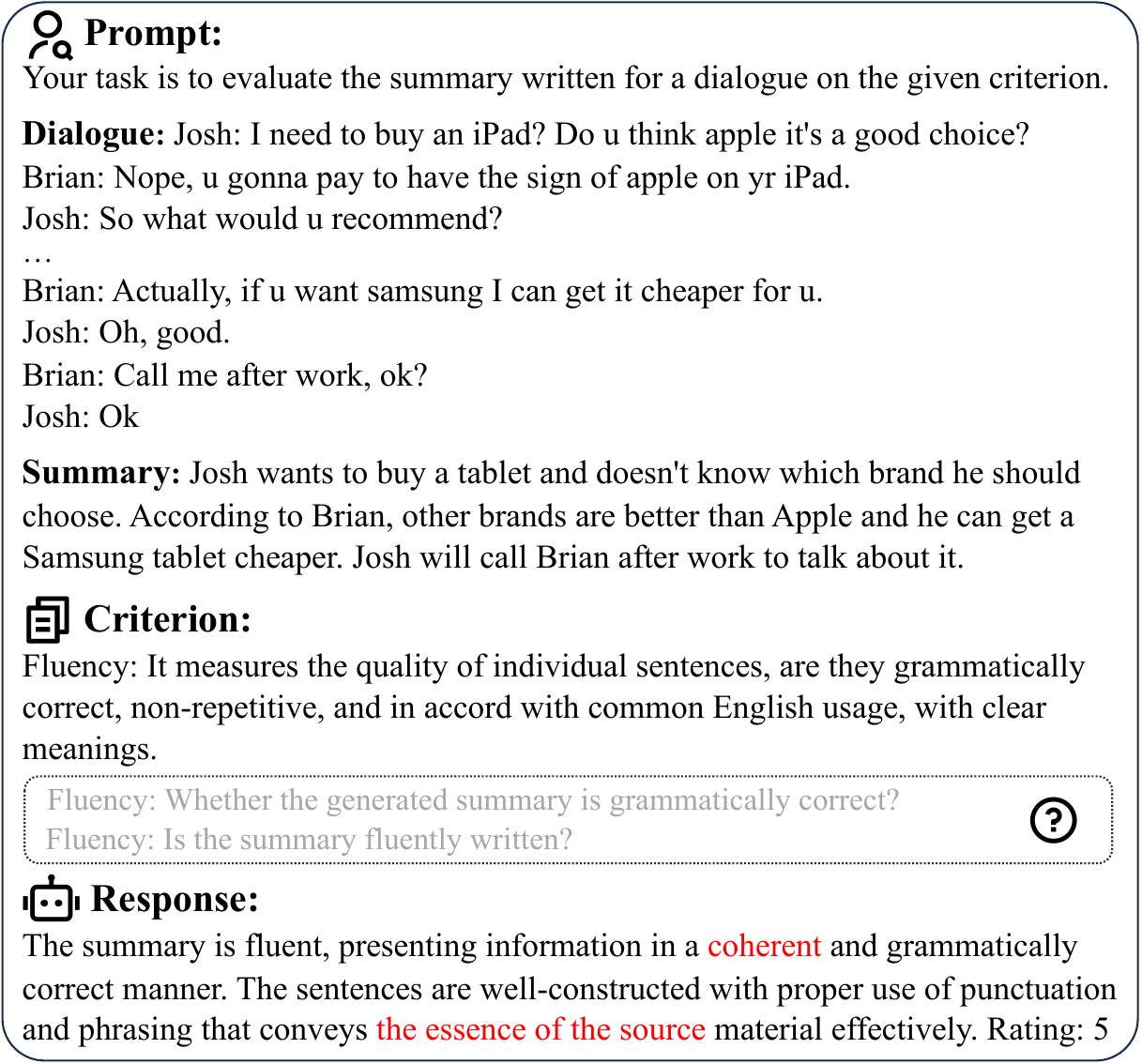}
\caption{An example of prompting LLMs to evaluate the dialogue summarization on criteria for fluency.}
\label{fig:1}
\end{figure}

However, during the explorations of LLM-based NLG evaluation, we observed two noteworthy phenomena that have not been revealed in previous work. First, the evaluation results from LLMs for a given aspect can achieve a higher correlation with human judgments on another clearly different aspect. Second, the correlations between LLM-generated scores across different aspects are significantly higher than those between human judgments accordingly. These lead us to \textbf{question the reliability of LLM evaluations on required aspects, since LLMs seem to confuse different aspects}.

Understanding these issues is inseparable from aspects themselves at first, which stem from human evaluation for NLG tasks and are typically described by terms and definitions, forming corresponding specific criteria. Through semi-structured interviews, \citet{DBLP:conf/naacl/ZhouBTDSO22} revealed that if aspects for evaluation lacked clear conceptualization, human annotators might conflate different aspects, such as fluency and grammaticality. \citet{DBLP:conf/inlg/HowcroftBCGHMMM20} pointed out the long-standing confusion of terms and definitions in human annotations, resulting in incomparable evaluations. Combining our investigation of previous work involving evaluation criteria, we believe that there are two distinct issues. The first is \textbf{inconsistent conceptualization}, where the definition is inconsistent with others for the same aspect but is clearly articulated. The second is \textbf{ambiguous expression}, where the definition is so vague that human annotators aren't sure what it really means. In Figure~\ref{fig:1}, we present an example of evaluation for fluency, where the criteria enclosed by the dashed box are selected from existing work and correspond to these two issues.

Therefore, we should reduce the influence of the issues within the evaluation criteria as much as possible, so as to reveal the actual performance of LLMs on NLG evaluation across aspects. We collect many existing criteria from previous papers involved, and summarize a clear hierarchical classification system for aspects that are most commonly used. For each aspect, we construct five criteria with descriptions of different levels of detail, including default, detailed, and simplified ones, to explore the corresponding effects. Then, inspired by behavioral testing in NLP \citep{DBLP:conf/acl/RibeiroWGS20}, we elaborately design a series of perturbation attacks based on the classification system to conduct targeted analyses on both proprietary LLMs (GPT-3.5 and GPT-4\footnote{\href{https://openai.com}{https://openai.com}}) and specifically fine-tuned LLMs like Prometheus \citep{DBLP:journals/corr/abs-2310-08491}. Different from previous related work, each of our perturbations is designed for a specific aspect to better verify the variances in evaluation for aspects that are related or not. We also engage human annotators to check our perturbations and expected impacts to enhance the reliability of our attack tests. To sum up, our contributions and findings are as follows: 
\begin{itemize}
    \item To the best of our knowledge, we are the first to explore the capabilities of LLMs in distinguishing aspects during NLG evaluation and the impacts of different criteria descriptions, bridging human and LLM-based evaluation.
    \item We summarize a classification system containing 11 common aspects and propose 18 aspect-targeted perturbation attacks, which have been verified by human annotators, to test the fine-grained evaluation behaviors of LLMs.
    \item Our experimental results reveal the confusion across different aspects in LLM-based evaluation, even for the powerful GPT-4, which necessitate attention and in-depth research. The related resources have been released\footnote{\href{https://github.com/PKU-ONELab/LLM-evaluator-reliability}{https://github.com/herrxy/LLM-evaluator-reliability}}, aiming to facilitate future relevant work.
\end{itemize}

\begin{table}
\centering
\small
\setlength{\tabcolsep}{4.5pt}
\begin{tabular}{lccccc}
\toprule
Evaluation Form & Flu. & Coh. & Rel. & Con. & Avg. \\ 
\midrule
Score only & 0.36 & 0.44 & 0.45 & 0.35 & 0.40 \\
Rate-explain & 0.37 & 0.53 & 0.48 & 0.44 & 0.45 \\
Analyze-rate & 0.41 & \textbf{0.58} & \textbf{0.50} & \textbf{0.57} & \textbf{0.52} \\
Analyze-rate (T=0) & 0.37 & 0.53 & 0.36 & 0.47 & 0.43\\
Analyze-rate (1-shot) & 0.31 & 0.42 & 0.33 & 0.42 & 0.37 \\
Analyze-rate (5-shot) & \textbf{0.47} & 0.51 & 0.44 & 0.53 & 0.49\\
\bottomrule
\end{tabular}
\caption{Pearson correlation coefficients between scores generated by GPT-3.5 with different forms of evaluation and human judgments on SummEval.}
\label{tab:1}
\end{table}

\begin{figure}
\centering
\includegraphics[width=0.48\textwidth]{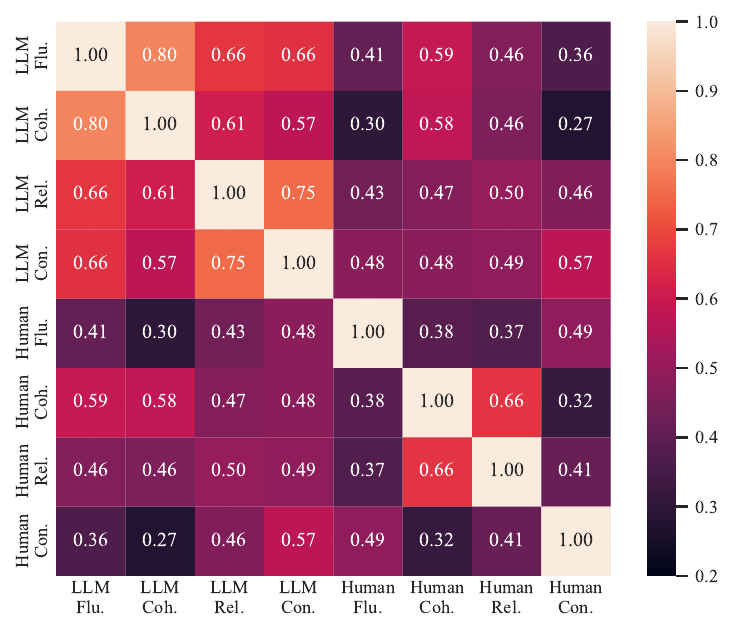}
\caption{Correlation between scores generated by GPT-3.5 or human annotators on four aspects in SummEval.}
\label{fig:2}
\end{figure}

\section{Preliminary Study}

\begin{figure*}
\centering
\includegraphics[width=\textwidth]{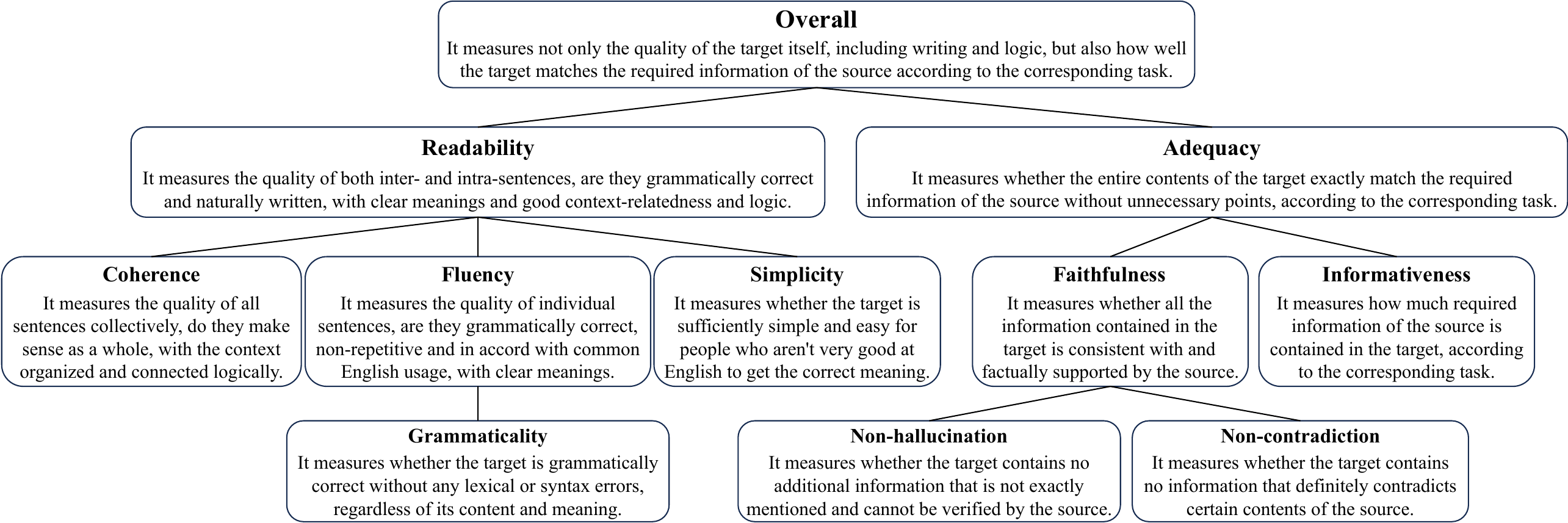}
\caption{Our summarized classification system for commonly-used aspects in NLG evaluation and their definitions.}
\label{fig:5}
\end{figure*}

To explore the NLG evaluation capabilities of LLMs and potential issues, we conduct experiments with GPT-3.5 on the commonly-used summarization evaluation dataset Summeval~\citep{DBLP:journals/tacl/FabbriKMXSR21}, attempting the evaluation forms introduced by~\citet{DBLP:conf/emnlp/ChiangL23}. Their work, as well as other studies~\citep{DBLP:journals/corr/abs-2303-04048, DBLP:conf/acl/ChiangL23, DBLP:conf/emnlp/LiuIXWXZ23}, has explored directly prompting LLMs for NLG evaluation. Furthermore, their evaluations are all zero-shot, so we additionally employ few-shot methods. The main experimental results are presented in Table~\ref{tab:1}. Consistent with the findings of~\citet{DBLP:conf/emnlp/ChiangL23}, requiring the model to analyze before rating (analyze-rate) along with multiple samplings and the temperature set to 1 achieves the best performance. These settings, therefore, are also used in our following experiments. However, it appears that the few-shot method has no effect as expected; instead, it leads to worse performance.

We also present the Pearson correlation coefficients between the evaluation scores of the model and human experts across different aspects in Figure~\ref{fig:2}. Interestingly, we notice some issues of confusion inherent in LLMs. First, the evaluation from the model is likely to achieve a higher correlation with human judgments on another aspect than the current aspect (such as fluency and relevance). On the other hand, most correlations between scores from the model across four aspects are significantly higher than corresponding ones between human judgments. It seems that GPT-3.5 confuses different aspects during evaluation to a certain extent, leading to a convergence in their results. We therefore study some cases of outputs and discover that GPT-3.5 indeed incorporates assessments regarding other aspects, illustrated as the red part in Figure~\ref{fig:1}. We speculate that this may lead to the poor performance of few-shot evaluations. And similar problems can also be found in GPT-4 and Prometheus~\citep{DBLP:journals/corr/abs-2310-08491}, with more discussions and details described in Appendix~\ref{sec:preliminary}. These results suggest the unreliabilities hidden in the LLM-based NLG evaluation, and more targeted research and experiments are required.

\section{Methodology}

\begin{table*}
\centering\small
\renewcommand{\arraystretch}{0.95}
\setlength{\tabcolsep}{2.5pt}
\begin{tabular}{llp{11.2cm}}
\toprule
\textbf{Aspect} & \textbf{Type} & \textbf{Perturbed Text}\\
\addlinespace[2pt] \hline \addlinespace[1.5pt]

\multirow{3}{*}{Original} & \multirow{3}{1.6cm}{-} & Josh wants to buy a tablet and doesn't know which brand he should choose. According to Brian, other brands are better than Apple and he can get a Samsung tablet cheaper. Josh will call Brian after work to talk about it. \\
\addlinespace[1pt] \hline \addlinespace[1pt]

\multirow{10.8}{1.6cm}{Fluency (Flu.)} & \multirow{4.2}{1.6cm}{Repetition} & Josh wants to buy a tablet and doesn't know which brand he should choose \sethlcolor{yellow}\hl{and make the selection of}. According to Brian, other brands are better than Apple and he can get a Samsung tablet cheaper \sethlcolor{yellow}\hl{at a lower price}. Josh will call \sethlcolor{yellow}\hl{and ring up} Brian after work to talk about it. \\ \addlinespace[1pt] \cline{2-3} \addlinespace[1pt]

 & \multirow{3.4}{1.6cm}{Passive Voice}  & Josh wants to buy a tablet and doesn't know which brand \sethlcolor{yellow}\hl{should be chosen} by him. According to Brian, other brands \sethlcolor{yellow}\hl{are considered} better than Apple, and a Samsung tablet \sethlcolor{yellow}\hl{can be got} cheaper by him. A call \sethlcolor{yellow}\hl{will be made} to Brian by Josh after work to talk about it. \\ \addlinespace[1pt] \cline{2-3} \addlinespace[1pt]

 & \multirow{3.3}{1.6cm}{Inversion} & Josh wants to buy a tablet, and \sethlcolor{yellow}\hl{which brand he should choose, he doesn't know}. \sethlcolor{yellow}\hl{Better than} \sethlcolor{yellow}\hl{Apple are other brands}, according to Brian, and he can get a Samsung tablet cheaper. \sethlcolor{yellow}\hl{Brian Josh will call} after work to talk about it. \\
\addlinespace[1pt] \hline \addlinespace[1pt]

\multirow{6.4}{1.6cm}{Coherence (Coh.)} & \multirow{3.2}{1.6cm}{Improper Connective} & Josh wants to buy a tablet and doesn't know which brand he should choose. \sethlcolor{yellow}\hl{Therefore}, according to Brian, other brands are better than Apple and he can get a Samsung tablet cheaper. \sethlcolor{yellow}\hl{However}, Josh will call Brian after work to talk about it. \\ \addlinespace[1pt] \cline{2-3} \addlinespace[1pt]

 & \multirow{3.2}{1.6cm}{Sentence Exchange} & \sethlcolor{yellow}\hl{Josh will call Brian after work to talk about it.} According to Brian, other brands are better than Apple and he can get a Samsung tablet cheaper. \sethlcolor{yellow}\hl{Josh wants to buy a tablet and doesn't know which brand he should choose.} \\ 
\addlinespace[1pt] \hline \addlinespace[1pt]

\multirow{10}{2.5cm}{Grammaticality (Gram.)} & \multirow{3.25}{1.6cm}{Incorrect Verb Form} & Josh want to \sethlcolor{yellow}\hl{buying} a tablet and doesn't \sethlcolor{yellow}\hl{knows} which brand he should choose. According to Brian, other brands \sethlcolor{yellow}\hl{is} better than Apple and he can \sethlcolor{yellow}\hl{gets} a Samsung tablet cheaper. Josh will \sethlcolor{yellow}\hl{called} Brian after work to \sethlcolor{yellow}\hl{talks} about it. \\ \addlinespace[1pt] \cline{2-3} \addlinespace[1pt]

 & \multirow{3.3}{1.6cm}{Word Exchange} & Josh wants to buy a \sethlcolor{yellow}\hl{and tablet} doesn't know which brand he should choose. According to Brian, \sethlcolor{yellow}\hl{brands other} are better than Apple and he can get a Samsung tablet cheaper. Josh will call Brian \sethlcolor{yellow}\hl{work after} to talk about it. \\ \addlinespace[1pt] \cline{2-3} \addlinespace[1pt]
 
 & \multirow{3.3}{1.6cm}{Spelling Mistake} & Josh \sethlcolor{yellow}\hl{wantts} to buy a tablet and doesn't \sethlcolor{yellow}\hl{kno} which brand he should choose. According to Brian, \sethlcolor{yellow}\hl{othe} brands are better than Apple and he can get a Samsung tablet \sethlcolor{yellow}\hl{cheapr}. Josh \sethlcolor{yellow}\hl{wwill} call Brian \sethlcolor{yellow}\hl{atfer} work to talk about it. \\
\addlinespace[1pt] \hline \addlinespace[1pt]

\multirow{7.2}{1.6cm}{Simplicity (Sim.)} & \multirow{3.3}{1.6cm}{Uncommon Phrase} & Josh wants to \sethlcolor{yellow}\hl{procure} a tablet and \sethlcolor{yellow}\hl{remains uncertain about} which brand he \sethlcolor{yellow}\hl{ought to} choose. \sethlcolor{yellow}\hl{As per} Brian, other brands are better than Apple and he can get a Samsung tablet \sethlcolor{yellow}\hl{at a more economical rate}. Josh will \sethlcolor{yellow}\hl{telephone} Brian after work to \sethlcolor{yellow}\hl{interflow} about it. \\ \addlinespace[1pt] \cline{2-3} \addlinespace[1pt]

 & \multirow{4}{1.6cm}{Complex Sentence} & Josh, who wants to buy a tablet, doesn't know which brand he should choose. According to Brian, who thinks that other brands are better than Apple, he can get a tablet whose brand is Samsung, which is cheaper. Josh will call someone who is Brian after work to talk about it. \\
\addlinespace[1pt] \hline \addlinespace[1pt]

\multirow{7.5}{2.5cm}{Informativeness (Inf.)} & \multirow{2}{1.6cm}{Abbreviation} & Josh wants to buy a tablet and doesn't decide the brand. Brian suggests non-Apple brands. Josh will discuss it with Brian. \\
\addlinespace[1pt] \cline{2-3} \addlinespace[1pt]

 & \multirow{3.2}{1.6cm}{Hypernym} & Josh wants to buy a \sethlcolor{yellow}\hl{device} and doesn't know which brand he should choose. According to Brian, other brands are better than Apple and he can get a \sethlcolor{yellow}\hl{Korean-brand device} cheaper. Josh will \sethlcolor{yellow}\hl{contact} Brian after work to talk about it. \\
\addlinespace[1pt] \cline{2-3} \addlinespace[1pt]

 & \multirow{2.1}{1.6cm}{Sentence Deletion} & Josh wants to buy a tablet and doesn't know which brand he should choose. According to Brian, other brands are better than Apple and he can get a Samsung tablet cheaper. \sethlcolor{yellow}\hl{ } \\
\addlinespace[1pt] \hline \addlinespace[1pt]

\multirow{8.2}{2.5cm}{Non-hallucination (Hal.) } & \multirow{4}{1.6cm}{Complement} & Josh wants to buy a tablet and doesn't know which brand he should choose. According to  Brian, \sethlcolor{yellow}\hl{who has extensive experience in tech gadget reviews,} other brands are better than Apple and he can get a Samsung tablet cheaper, \sethlcolor{yellow}\hl{known for its high-resolution display and long battery life.} Josh will call Brian after work, \sethlcolor{yellow}\hl{around 6 PM,} to talk about it. \\ \addlinespace[1pt] \cline{2-3} \addlinespace[1pt]

 & \multirow{4}{1.6cm}{Continuation} & Josh wants to buy a tablet and doesn't know which brand he should choose. According to Brian, other brands are better than Apple and he can get a Samsung tablet cheaper. Josh will call Brian after work to talk about it. \sethlcolor{yellow}\hl{He's hoping that Brian can provide some insight into the pros and cons of various products that fit within his budget in detail.} \\
\addlinespace[1pt] \hline \addlinespace[1pt]

\multirow{10}{2.5cm}{Non-contradiction (Cont.)} & \multirow{3.3}{1.6cm}{Different Entity} & Josh wants to buy a \sethlcolor{yellow}\hl{smartphone} and doesn't know which brand he should choose. According to Brian, other brands are better than \sethlcolor{yellow}\hl{Sony}, and he can get a Samsung \sethlcolor{yellow}\hl{smartphone} cheaper. Josh will call Brian after \sethlcolor{yellow}\hl{school} to talk about it. \\ \addlinespace[1pt] \cline{2-3} \addlinespace[1pt]

 & \multirow{3.3}{1.6cm}{Conflicting Fact} & Josh wants to buy a tablet and \sethlcolor{yellow}\hl{roughly knows} which brand he should choose. According to Brian, \sethlcolor{yellow}\hl{Apple is the best brand and he should avoid Samsung tablets at all costs}. Josh will call Brian \sethlcolor{yellow}\hl{at once} to talk about it. \\ \addlinespace[1pt] \cline{2-3} \addlinespace[1pt]

 & \multirow{3.1}{1.6cm}{Negation} & Josh wants to buy a tablet and doesn't know which brand he should choose. According to Brian, other brands are better than Apple and he can get a Samsung tablet cheaper. \sethlcolor{yellow}\hl{Nevertheless,} Josh will \sethlcolor{yellow}\hl{not} call Brian after work to talk about it. \\

\addlinespace[-0.5pt] \bottomrule
\end{tabular}
\caption{Our designed 18 perturbations and corresponding examples, where the modifications from the original reference text (the summary in Figure~\ref{fig:1}) have been highlighted.}
\label{tab:2}
\end{table*}

\begin{table*}
\centering\small
\setlength{\tabcolsep}{5pt}
\begin{tabular}{lp{13cm}}
\toprule
 \textbf{Type} & \textbf{Criterion (Term and Definition)}\\
\addlinespace[1.5pt] \hline \addlinespace[2pt]

Simplified & Fluency: It measures whether individual sentences are grammatically correct and well-written. \\
\addlinespace[1pt] \hline \addlinespace[2pt]

\multirow{4}{2cm}{Detailed} & Fluency: It measures the quality of individual sentences, are they grammatically correct, non-repetitive, and in accord with common English usage, with clear meanings. Consider whether there are misspellings, tense errors, missing determiners, or more severe problems, such as duplication, unfamiliar phrases, complex syntactic structures, and missing components. \\
\addlinespace[1pt] \hline \addlinespace[2pt]

Term & Fluency: It measures whether the target is fluent. \\
\addlinespace[1pt] \hline \addlinespace[2pt]

\multirow{8}{2cm}{List} & Fluency: It measures the quality of individual sentences, are they grammatically correct, non-repetitive, and in accord with common English usage, with clear meanings. \\
& Score 5: Entirely fluent, grammatically correct, and well-written. \\
& Score 4: Only containing some minor non-fluent parts or grammatical errors that basically have no effect. \\
& Score 3: Fluent in general, with some obvious grammatical errors and unfamiliar phrases. \\
& \makecell[l]{Score 2: There are major grammatical errors, duplication, unfamiliar phrases and syntactic structures,\\ \addlinespace[-1pt] and missing components, but some fluent segments.} \\
& Score 1: Not fluent at all, full of meaningless fragments and unclear contents. \\
\addlinespace[-1pt] \bottomrule
\end{tabular}
\caption{Examples of different criterion descriptions for fluency with definitions of different levels of detail.}
\label{tab:4}
\end{table*}

Our findings in the preliminary study lead us to question whether LLMs can understand and execute the evaluation requirements represented by different criteria well. To conduct more in-depth explorations, we propose the fine-grained perturbation test inspired by behavioral testing~\citep{DBLP:conf/acl/RibeiroWGS20}, hoping to reveal their more actual capacities for NLG evaluation. In particular, instead of relatively coarse-grained perturbations in previous work, our perturbation attacks have been crafted to specifically target certain evaluation aspects without affecting evaluations for other unrelated ones. We formulate our approach as follows:

We first collect a set of different common criteria denoted as $C=\{c_i, i=1,2,\ldots m\}$, conceptually involving inclusive and non-inclusive relationships. And each of our perturbation attacks $p_j, j=1,2,\ldots n$ is applied to the original text $x$ to generate the corresponding perturbed text $p_j(x)$. Meanwhile, each perturbation is designed and expected to only reduce the text quality regarding the criteria for the originally targeted aspect and others whose scopes cover it. We define the set of criteria affected by the perturbation $p_j$ as $C^j_T$, and the rest in $C$ are defined as $C^j_F$. And the distinction between these two groups is made more reliable based on our classification system and human annotations. Then, to conduct the test, we prompt the model to evaluate all the perturbed texts, as well as the original text, in the form of scoring to check the expected two different evaluation behaviors:

\vspace{-\baselineskip}
{
\small
\begin{spacing}{0.75}
\begin{multline*}
\small
S_T = \{\frac{1}{N} \sum_{k=1}^N (M_s(x_k, v_k, c_i)-M_s(p_j(x_k), v_k, c_i)) \\
\mid \forall i, j,\ s.t.\ c_i \in C^j_T \}
\end{multline*}
\end{spacing}
}
\noindent where $N$ denotes the number of original texts pending perturbations in our test, and $M_s$ serves as the model's scoring based on provided information, which includes additional necessary content $v$ aside from texts and criteria to evaluate, such as task instructions. $S_T$ represents the set of those that should be affected, where each item $s^j_i$ represents the change in evaluation scores after the perturbation $p_j$ regarding the criterion $c_i$, which should be significant. On the other hand, $S_F$ is defined in a similar manner, but each item of it is expected to be zero, showing no impacts of perturbations. We will describe important components of our approach in more detail in the following sections.

\subsection{Classification System for Aspects}

As mentioned in~\citet{DBLP:conf/inlg/HowcroftBCGHMMM20,DBLP:conf/naacl/ZhouBTDSO22}, there is inconsistent and unclear conceptualization in existing evaluation aspects, which makes it difficult to understand the requirements and relationships among them. In light of this, we carefully collect and read about 300 papers that involve various aspects for NLG evaluation, and select those most commonly used. Then, we integrate their definitions used in the corresponding work and construct our default criteria as unambiguously as possible. Furthermore, they can be organized as a tree-like classification system, as shown in Figure~\ref{fig:5}, thanks to the relatively clear relationships within our definitions.

\subsection{Perturbation Attacks}

For each fundamental aspect in Figure~\ref{fig:5}, we design several targeted perturbation attacks, as displayed with the corresponding examples in Table~\ref{tab:2}. Since fluency involves more considerations other than grammaticality, we also propose additional perturbations for it, like adding repetitive content. The perturbations are crafted and expected to affect only the current aspect and those located at its ancestor nodes as much as possible. Compared to the prior perturbation research, where the texts for the attack are constructed for universal checking using simple templates or rules, our perturbations are more fine-grained and require better controls during generation, like adding complements that should be related and not contradictory for non-hallucination. Therefore, we manually generate some high-quality examples as demonstrations, along with corresponding instructions, and then prompt the powerful GPT-4 to construct the perturbed texts in 10-shot settings. We have conducted a sampling and manual inspection of them to ensure their quality and reliability, and more details are described in Appendix~\ref{sec:perturbation}.

\subsection{Different Descriptions of Criteria}

Since different definitions are often used in practice for the same aspect, forming different corresponding criteria, we also intend to study the impact of different levels of detail in definitions, as shown in Table~\ref{tab:4}. We take fluency as an example, and there are four different types beside our default definitions in Figure~\ref{fig:5}: simplified, detailed, term, and list. Moreover, to better analyze the existing issues of quality criteria, we select several typical criteria that have been used for NLG evaluation from the existing literature for each aspect. The resources and data mentioned above, including the collections of criteria, prompts for data construction, perturbed texts, and relevant experimental results, are released to facilitate the development of future research on NLG evaluation.

\section{Data and Test Settings}

\paragraph{Datasets.} We select three common NLG tasks: summarization (including news and dialogue), paraphrase, and table-to-text generation, for our experiments and tests. The construction of perturbation attacks requires high-quality original texts to ensure significant declines in the qualities of different aspects. However, previous studies typically employed references directly from the corresponding datasets, which are always generated by some rules instead of being written by humans, leading to unsatisfactory quality~\citep{DBLP:conf/emnlp/KryscinskiKMXS19, DBLP:journals/corr/abs-2309-09558, DBLP:conf/emnlp/SottanaLZY23}. Therefore, we carefully prompt the powerful GPT-4 to obtain better references based on the original data. We finally sample 1000 pieces of data, each of which is subjected to 18 different perturbations we propose.

\paragraph{LLMs in Tests.} Our tests cover both proprietary LLMs (GPT-3.5 and GPT-4) and open-source LLMs (Prometheus). GPT-3.5 and GPT-4 have been mentioned in many existing studies as performing well in flexible NLG evaluation. On the other hand, some research has recently shifted toward fine-tuning specialized open-source LLMs for evaluation, aiming to avoid the deficiencies of prompting LLMs—such as high costs and unstable reproducibility. However, most of them do not support evaluation with specified criteria, and among those remaining, only Prometheus~\citep{DBLP:journals/corr/abs-2310-08491} fine-tuned on Llama-2-Chat-13B~\citep{DBLP:journals/corr/abs-2307-09288} has released their model.

\paragraph{Human Judgement} To more reliably distinguish whether different criteria would be affected by specific perturbation attacks, we conduct human annotations and judgments beyond the guidance of the classification system. Due to the high cost and time-consuming nature of human annotations, it is not feasible to manually judge all the data. So we sample a portion of the data and recruit 40 annotators (each of whom is proficient in English and possesses certifications) to ensure that each piece of data is annotated four times. Overall, the more detailed the aspect definitions are, the more the corresponding human judgments match our expectations based on the classification system, as well as higher annotation consistency. In particular, definitions of detailed type achieve the highest match rate of 94.4\%, with full results shown in Table~\ref{tab:7}.

More details including related discussions and prompts used are described in Appendix~\ref{sec:data}.

\section{Experiments}

\begin{table*}[!ht]
\centering
\small
\setlength{\tabcolsep}{2.5pt}
\begin{tabular}{ll>{\centering\arraybackslash}m{0.88cm}>{\centering\arraybackslash}m{0.88cm}>{\centering\arraybackslash}m{0.88cm}>{\centering\arraybackslash}m{0.88cm}>{\centering\arraybackslash}m{0.88cm}>{\centering\arraybackslash}m{0.88cm}>{\centering\arraybackslash}m{0.88cm}>{\centering\arraybackslash}m{0.88cm}>{\centering\arraybackslash}m{0.88cm}>{\centering\arraybackslash}m{0.88cm}>{\centering\arraybackslash}m{0.88cm}}
\toprule
\multicolumn{2}{c}{\textbf{Perturbation Attack}} & \textbf{Flu.} & \textbf{Coh.} & \textbf{Gram.} & \textbf{Sim.} & \textbf{Read.} & \textbf{Fai.} & \textbf{Cont.} & \textbf{Hal.} & \textbf{Inf.} & \textbf{Ade.} & \textbf{All.} \\
\addlinespace[1.5pt] \hline \addlinespace[1.5pt]

\multirow{3}{*}{Flu.} & Repetition & \cellcolor{myblue2!6.79}0.20 & \cellcolor{myblue2!2.99}\underline{0.06} & \cellcolor{myblue2!6.52}\underline{0.19} & \cellcolor{myblue2!8.7}\underline{0.27} & \cellcolor{myblue2!6.25}0.18 & \cellcolor{myblue2!4.89}\underline{0.13} & \cellcolor{myblue2!5.43}\underline{0.15} & \cellcolor{myblue2!7.88}\underline{0.24} & \cellcolor{myblue2!2.99}\underline{0.06} & \cellcolor{myblue2!4.08}\underline{0.10} & \cellcolor{myblue2!4.08}0.10 \\
 & Passive Voice & \cellcolor{myblue2!14.4}\uwave{0.48} & \cellcolor{myblue2!6.79}\underline{0.20} & \cellcolor{myblue2!16.3}\underline{0.55} & \cellcolor{myblue2!8.7}\underline{0.27} & \cellcolor{myblue2!13.32}\uwave{0.44} & \cellcolor{myblue2!4.35}\underline{0.11} & \cellcolor{myblue2!4.35}\underline{0.11} & \cellcolor{myblue2!3.26}\underline{0.07} & \cellcolor{myblue2!5.16}\underline{0.14} & \cellcolor{myblue2!4.89}\underline{0.13} & \cellcolor{myblue2!6.79}0.20 \\
 & Inversion & \cellcolor{myblue2!36.68}\uwave{1.30} & \cellcolor{myblue2!19.29}\underline{0.66} & \cellcolor{myblue2!41.58}\underline{1.48} & \cellcolor{myblue2!17.93}\underline{0.61} & \cellcolor{myblue2!36.96}\uwave{1.31} & \cellcolor{myblue2!9.78}\underline{0.31} & \cellcolor{myblue2!10.33}\underline{0.33} & \cellcolor{myblue2!5.71}\underline{0.16} & \cellcolor{myblue2!11.14}\underline{0.36} & \cellcolor{myblue2!11.68}\underline{0.38} & \cellcolor{myblue2!18.48}\uwave{0.63} \\
\addlinespace[1.5pt] \hline \addlinespace[1.5pt]

\multirow{2}{*}{Coh.} & Improper Connective & \cellcolor{myblue2!4.89}\underline{0.13} & \cellcolor{myblue2!2.72}\uwave{0.05} & \cellcolor{myblue2!4.35}\underline{0.11} & \cellcolor{myblue2!5.71}\underline{0.16} & \cellcolor{myblue2!4.89}\uwave{0.13} & \cellcolor{myblue2!4.89}\underline{0.13} & \cellcolor{myblue2!6.52}\underline{0.19} & \cellcolor{myblue2!9.24}\underline{0.29} & \cellcolor{myblue2!3.53}\underline{0.08} & \cellcolor{myblue2!3.8}\underline{0.09} & \cellcolor{myblue2!3.53}\uwave{0.08} \\
 & Sentence Exchange & \cellcolor{myblue2!5.98}\underline{0.17} & \cellcolor{myblue2!5.43}\uwave{0.15} & \cellcolor{myblue2!5.98}\underline{0.17} & \cellcolor{myblue2!2.99}\underline{0.06} & \cellcolor{myblue2!5.71}\uwave{0.16} & \cellcolor{myblue2!4.62}\underline{0.12} & \cellcolor{myblue2!5.16}\underline{0.14} & \cellcolor{myblue2!5.71}\underline{0.16} & \cellcolor{myblue2!5.16}\underline{0.14} & \cellcolor{myblue2!4.62}\underline{0.12} & \cellcolor{myblue2!4.62}\uwave{0.12} \\ 
\addlinespace[1.5pt] \hline \addlinespace[1.5pt]

\multirow{3}{*}{Gram.} & Incorrect Verb Form & \cellcolor{myblue2!67.39}\uwave{2.43} & \cellcolor{myblue2!27.99}\underline{0.98} & \cellcolor{myblue2!82.07}\uwave{2.97} & \cellcolor{myblue2!28.53}\underline{1.00} & \cellcolor{myblue2!63.32}\uwave{2.28} & \cellcolor{myblue2!16.58}\underline{0.56} & \cellcolor{myblue2!27.72}\underline{0.97} & \cellcolor{myblue2!13.32}\underline{0.44} & \cellcolor{myblue2!14.4}\underline{0.48} & \cellcolor{myblue2!17.66}\underline{0.60} & \cellcolor{myblue2!36.68}\uwave{1.30} \\
 & Word Exchange & \cellcolor{myblue2!82.07}\uwave{2.97} & \cellcolor{myblue2!48.64}\underline{1.74} & \cellcolor{myblue2!92.66}\uwave{3.36} & \cellcolor{myblue2!43.21}\underline{1.54} & \cellcolor{myblue2!82.07}\uwave{2.97} & \cellcolor{myblue2!23.1}\underline{0.80} & \cellcolor{myblue2!37.23}\underline{1.32} & \cellcolor{myblue2!17.39}\underline{0.59} & \cellcolor{myblue2!24.46}\underline{0.85} & \cellcolor{myblue2!26.9}\underline{0.94} & \cellcolor{myblue2!52.45}\uwave{1.88} \\
 & Spelling Mistake & \cellcolor{myblue2!90.22}\uwave{3.27} & \cellcolor{myblue2!32.88}\underline{1.16} & \cellcolor{myblue2!100.54}\uwave{3.65} & \cellcolor{myblue2!36.68}\underline{1.30} & \cellcolor{myblue2!84.51}3.06 & \cellcolor{myblue2!20.11}\underline{0.69} & \cellcolor{myblue2!34.51}\underline{1.22} & \cellcolor{myblue2!14.4}\underline{0.48} & \cellcolor{myblue2!17.93}\underline{0.61} & \cellcolor{myblue2!22.01}\underline{0.76} & \cellcolor{myblue2!47.55}\uwave{1.70} \\
\addlinespace[1.5pt] \hline \addlinespace[1.5pt]

\multirow{2}{*}{Sim.} & Uncommon Phrase & \cellcolor{myblue2!6.79}\underline{0.20} & \cellcolor{myblue2!2.45}\underline{0.04} & \cellcolor{myblue2!5.98}\underline{0.17} & \cellcolor{myblue2!23.1}\uwave{0.80} & \cellcolor{myblue2!6.25}\uwave{0.18} & \cellcolor{myblue2!3.8}\underline{0.09} & \cellcolor{myblue2!3.26}\underline{0.07} & \cellcolor{myblue2!3.53}\underline{0.08} & \cellcolor{myblue2!3.8}\underline{0.09} & \cellcolor{myblue2!3.8}\underline{0.09} & \cellcolor{myblue2!4.08}0.10 \\
 & Complex Sentence & \cellcolor{myblue2!19.29}\underline{0.66} & \cellcolor{myblue2!7.61}\underline{0.23} & \cellcolor{myblue2!16.58}\underline{0.56} & \cellcolor{myblue2!17.12}\uwave{0.58} & \cellcolor{myblue2!18.21}0.62 & \cellcolor{myblue2!5.71}\underline{0.16} & \cellcolor{myblue2!5.71}\underline{0.16} & \cellcolor{myblue2!5.71}\underline{0.16} & \cellcolor{myblue2!6.79}\underline{0.20} & \cellcolor{myblue2!6.25}\underline{0.18} & \cellcolor{myblue2!9.24}0.29 \\
\addlinespace[1.5pt] \hline \addlinespace[1.5pt]
 
\multirow{3}{*}{Inf.} & Abbreviation & \cellcolor{myblue2!1.36}\underline{-0.03} & \cellcolor{myblue2!1.63}\underline{0.01} & \cellcolor{myblue2!1.36}\underline{-0.05} & \cellcolor{myblue2!1.36}\underline{-0.01} & \cellcolor{myblue2!1.36}\underline{0.00} & \cellcolor{myblue2!1.36}\underline{-0.02} & \cellcolor{myblue2!1.36}\underline{-0.01} & \cellcolor{myblue2!1.36}\underline{-0.02} & \cellcolor{myblue2!2.45}\uwave{0.04} & \cellcolor{myblue2!1.63}\uwave{0.01} & \cellcolor{myblue2!1.63}\uwave{0.01} \\
 & Hypernym & \cellcolor{myblue2!9.24}\underline{0.29} & \cellcolor{myblue2!6.25}\underline{0.18} & \cellcolor{myblue2!7.61}\underline{0.23} & \cellcolor{myblue2!4.08}\underline{0.10} & \cellcolor{myblue2!9.24}\underline{0.29} & \cellcolor{myblue2!5.43}\underline{0.15} & \cellcolor{myblue2!5.71}\underline{0.16} & \cellcolor{myblue2!6.52}\underline{0.19} & \cellcolor{myblue2!5.71}\uwave{0.16} & \cellcolor{myblue2!5.98}\uwave{0.17} & \cellcolor{myblue2!7.88}\uwave{0.24} \\
 & Sentence Deletion & \cellcolor{myblue2!2.72}\underline{0.05} & \cellcolor{myblue2!3.8}\underline{0.09} & \cellcolor{myblue2!1.9}\underline{0.02} & \cellcolor{myblue2!1.63}\underline{0.01} & \cellcolor{myblue2!2.72}\underline{0.05} & \cellcolor{myblue2!2.72}\underline{0.05} & \cellcolor{myblue2!2.72}\underline{0.05} & \cellcolor{myblue2!1.36}\underline{-0.02} & \cellcolor{myblue2!4.35}\uwave{0.11} & \cellcolor{myblue2!2.99}\uwave{0.06} & \cellcolor{myblue2!4.08}\uwave{0.10} \\
\addlinespace[1.5pt] \hline \addlinespace[1.5pt]
 
\multirow{2}{*}{Hal.} & Complement & \cellcolor{myblue2!8.7}\underline{0.27} & \cellcolor{myblue2!7.34}\underline{0.22} & \cellcolor{myblue2!12.5}\underline{0.41} & \cellcolor{myblue2!31.25}\underline{1.10} & \cellcolor{myblue2!10.33}\underline{0.33} & \cellcolor{myblue2!28.8}\uwave{1.01} & \cellcolor{myblue2!33.42}\underline{1.18} & \cellcolor{myblue2!60.33}\uwave{2.17} & \cellcolor{myblue2!7.61}\underline{0.23} & \cellcolor{myblue2!18.48}\uwave{0.63} & \cellcolor{myblue2!11.96}\uwave{0.39} \\
 & Continuation & \cellcolor{myblue2!2.17}\underline{0.03} & \cellcolor{myblue2!1.36}\underline{0.00} & \cellcolor{myblue2!2.45}\underline{0.04} & \cellcolor{myblue2!5.98}\underline{0.17} & \cellcolor{myblue2!2.17}\underline{0.03} & \cellcolor{myblue2!6.52}\uwave{0.19} & \cellcolor{myblue2!7.61}\underline{0.23} & \cellcolor{myblue2!28.8}\uwave{1.01} & \cellcolor{myblue2!1.9}\underline{0.02} & \cellcolor{myblue2!4.08}\uwave{0.10} & \cellcolor{myblue2!2.45}\uwave{0.04} \\
\addlinespace[1.5pt] \hline \addlinespace[1.5pt]
 
\multirow{3}{*}{Cont.} & Different Entity & \cellcolor{myblue2!54.89}\underline{1.97} & \cellcolor{myblue2!53.26}\underline{1.91} & \cellcolor{myblue2!58.7}\underline{2.11} & \cellcolor{myblue2!44.29}\underline{1.58} & \cellcolor{myblue2!56.52}\underline{2.03} & \cellcolor{myblue2!71.2}\uwave{2.57} & \cellcolor{myblue2!80.43}\uwave{2.91} & \cellcolor{myblue2!73.1}2.64 & \cellcolor{myblue2!49.46}\uwave{1.77} & \cellcolor{myblue2!64.13}\uwave{2.31} & \cellcolor{myblue2!61.14}\uwave{2.20} \\
 & Conflicting Fact & \cellcolor{myblue2!77.45}\underline{2.80} & \cellcolor{myblue2!80.43}\underline{2.91} & \cellcolor{myblue2!76.9}\underline{2.78} & \cellcolor{myblue2!70.38}\underline{2.54} & \cellcolor{myblue2!81.52}\underline{2.95} & \cellcolor{myblue2!96.2}\uwave{3.49} & \cellcolor{myblue2!101.36}\uwave{3.68} & \cellcolor{myblue2!94.02}3.41 & \cellcolor{myblue2!69.84}\uwave{2.52} & \cellcolor{myblue2!87.77}\uwave{3.18} & \cellcolor{myblue2!85.33}\uwave{3.09} \\
 & Negation & \cellcolor{myblue2!8.15}\underline{0.25} & \cellcolor{myblue2!7.88}\underline{0.24} & \cellcolor{myblue2!7.34}\underline{0.22} & \cellcolor{myblue2!4.62}\underline{0.12} & \cellcolor{myblue2!8.15}\underline{0.25} & \cellcolor{myblue2!12.77}\uwave{0.42} & \cellcolor{myblue2!17.93}\uwave{0.61} & \cellcolor{myblue2!12.5}0.41 & \cellcolor{myblue2!7.88}\uwave{0.24} & \cellcolor{myblue2!9.78}\uwave{0.31} & \cellcolor{myblue2!9.78}\uwave{0.31} \\

\addlinespace[-0.0pt]\bottomrule
\end{tabular}
\caption{The variances of evaluation scores from GPT-3.5 between original texts and different perturbed texts. The abbreviations in the first line represent Fluency, Coherence, Grammaticality, Simplicity, Readability, Faithfulness, Non-contradiction, Non-hallucination, Informativeness, Adequacy, and Overall, respectively.}
\label{tab:5}
\end{table*}

We primarily display the experiments and analyze the results with GPT-3.5, and the performance of other LLMs is described in Section 5.4. And to minimize the interference from criteria themselves, we first analyze the results with the detailed type of aspect definitions, which have also been confirmed through human judgments, to align most closely with our expectations. The main results are shown in Table~\ref{tab:5}, each item of which represents the average variations between the evaluation scores of pre- and post-perturbation attacks, respectively. Moreover, those items with the consistent judgments as shown in Table~\ref{tab:7} can be categorized into two groups: $S_T$ (with wavy lines) and $S_F$ (with underlines), as defined in Section 3, which correspond to the directional expectation test for impactful attacks and the invariance test for non-impactful attacks, respectively. Furthermore, we explore the effects of different levels of detail in aspect definitions. The complete experimental results for three LLMs can be found in Appendix~\ref{sec:experiments}.

\subsection{Directional Expectation Test}

The results show that the perturbations for Coherence and Informativeness almost did not lead to any degradation, with the changes in evaluation scores of pre- and post-perturbation all less than 0.2. However, definite but different human judgments that they should affect respective aspects indicate that the model lacks understanding of these two aspects. As for Fluency, the impact of perturbations intensified progressively from repetition to passive voice and then to inversion, consistent with intuition since the degree of sentence alteration increases. Specifically, despite our explicit mention that redundant information should be considered in evaluations regarding Fluency, both GPT-3.5 and human annotators fail to adhere to the instruction. Through discussions with human annotators, we find that repetition issues are common and easy to ignore, which may lead to such verbosity bias in LLMs (also observed by~\citet{DBLP:journals/corr/abs-2306-05685}) through these issues within training data. On the other hand, all perturbations for Grammaticality and Non-contradiction except for negation, as well as complement for Non-hallucination successfully show noticeable and expected decreases (greater than 2). And the remaining ones, like those for Simplicity, are not pronounced, with score variations ranging between 0.5 and 1.

\subsection{Invariance Test}

Conversely, in situations where the evaluation should not be affected, the primary deviations from expectations and human judgments exist in Grammaticality and Non-contradiction, particularly the latter. Grammatical issues influence all criteria, yet there is a clear hierarchy. The undeserved impacts on Coherence and Simplicity—aspects also included in Readability—are greater than those under Adequacy which are more unrelated. And they also seem lesser compared to criteria that are indeed expected to be affected, such as Fluency. It indicates that while GPT-3.5 struggles to disregard grammatical errors when assessing irrelevant criteria, it can still differentiate to some extent. However, two perturbations for Non-contradiction cause almost indistinguishable degradations in all criteria, even those under Readability that do not require the source content. In comparison, Non-hallucination, also part of Faithfulness, does not result in similar behaviors. This suggests that GPT-3.5 may be overly sensitive to conflicting points between the target and source content, while being more restrained in judging unverifiable information.

\subsection{Different Definition Types}

\begin{figure*}[!ht]
  \centering
  \begin{subfigure}[b]{0.425\textwidth}
    \centering
    \includegraphics[width=\textwidth]{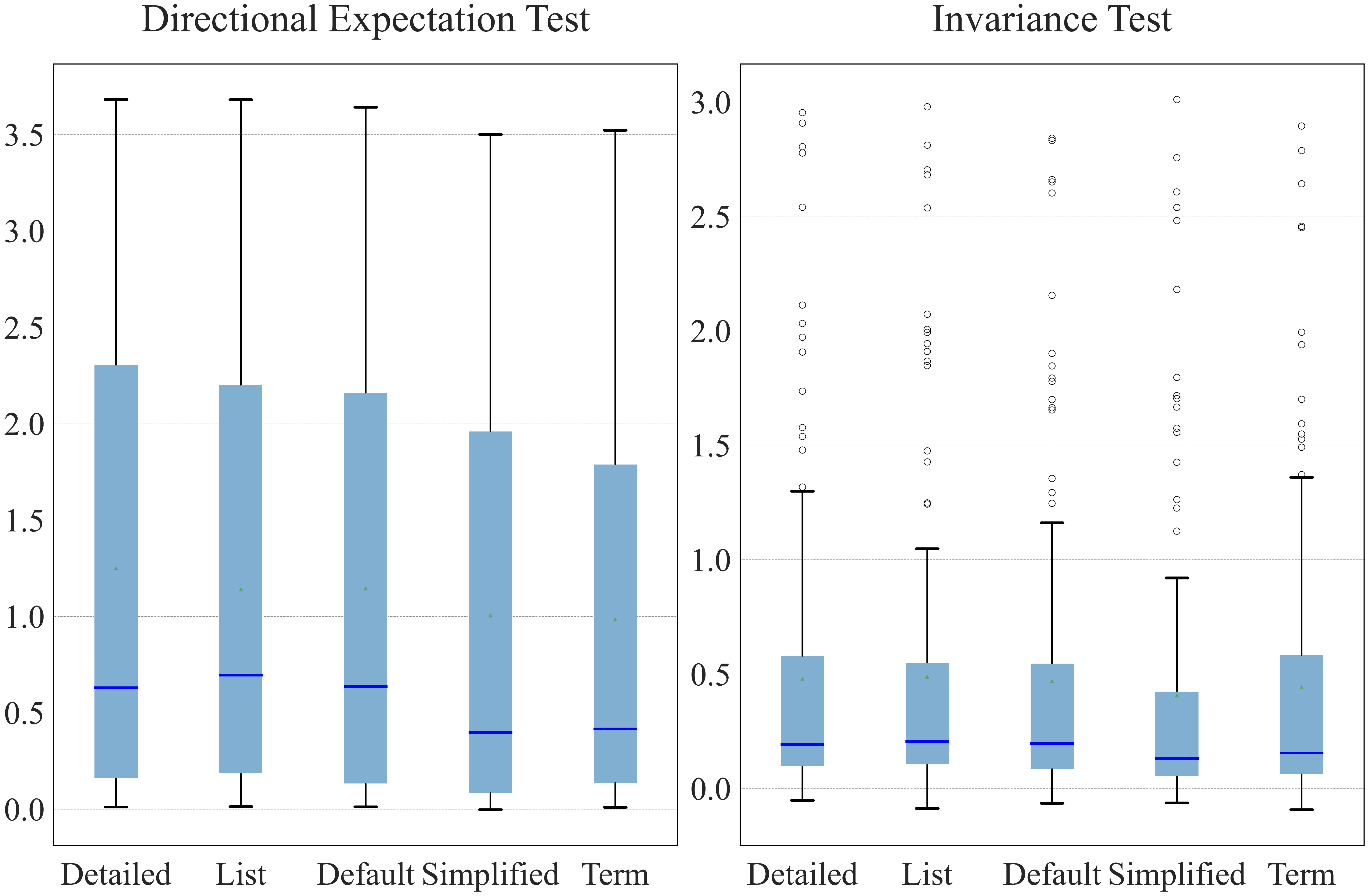}
    \caption{Distributions of variances in evaluation scores.}
    \label{fig:8-1} 
  \end{subfigure}
  \begin{subfigure}[b]{0.555\textwidth}
    \centering
    \includegraphics[width=\textwidth]{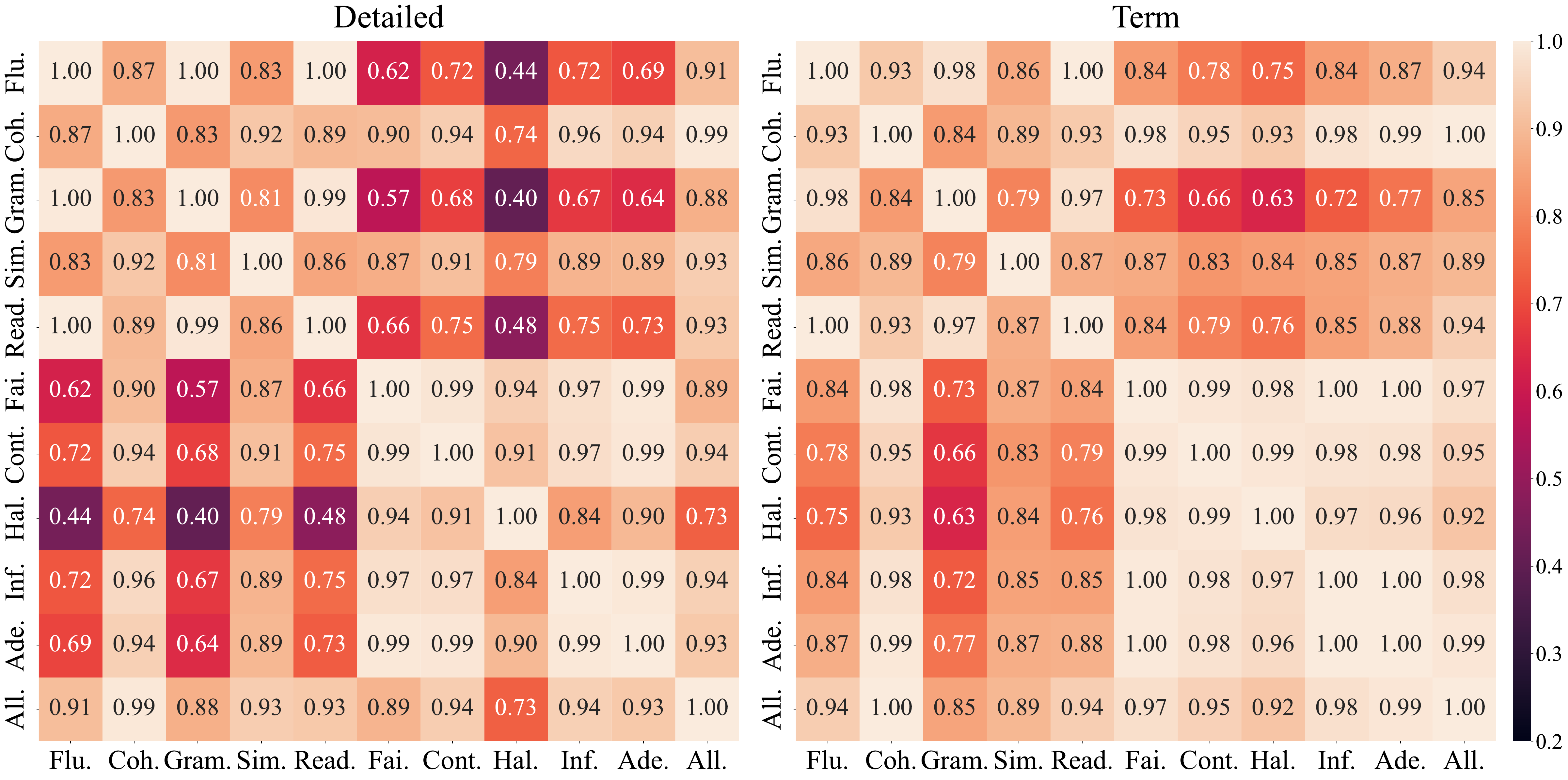}
    \caption{Correlations between evaluation scores across different aspects.}
    \label{fig:8-2} 
  \end{subfigure}
  \hspace{-0.01in}
  \caption{Boxplots for items of two tests and correlation matrices for description types of detailed and term.}
  \label{fig:8} 
\end{figure*}

Considering the strong instruction-following capabilities of current LLMs, it is intuitive that the more detailed the description of criteria, the more accurate the evaluation from the model should be. However, the correlations between the results of five different types of criterion descriptions are quite high, as shown in Figure~\ref{fig:9}. The almost same evaluation behaviors suggest that GPT-3.5 may rely primarily on terminology to understand and assess each criterion. And we speculate that our written definitions are close to the inherent understanding of the model for corresponding terms, which consequently leads to such a phenomenon. The related knowledge is likely derived from a wide range of pre-training corpora. In contrast, human annotators, who lack extensive NLG evaluation experience, indeed exhibit different performance when given these different types of descriptions, as shown in Appendix~\ref{sec:data}.

Furthermore, for deeper comparison, we display the score distributions of $S_T$ and $S_F$ with different description types in Figure~\ref{fig:8-1}. It seems that exhaustive descriptions can still help the model make more clear judgments to some extent, since the variations in the evaluation of $S_T$ are more significant. However, the changes in scores in the invariant situation are somewhat erratic, proving that the confusion issues are unrelated to whether descriptions are detailed or not. In addition, we also calculate the correlation of evaluation scores for different aspects for each description type. We present the results of the detailed and term in Figure~\ref{fig:8-2}, with the complete results displayed in Appendix~\ref{sec:experiments}. It is evident that the less detailed the description is, the more similar the evaluations for different aspects are, indicating more severe confusion.

\subsection{Different LLMs}

We have also conducted the same experiments on GPT-4 and Prometheus for comparative analysis of different types of LLMs. Due to the large scale of our perturbation attacks and the high cost of prompting GPT-4, we sampled one-fifth of the data for the test for GPT-4. All of the results are shown in Appendix~\ref{sec:experiments} and corresponding figures. We find that both the more powerful GPT-4 and the specially fine-tuned Prometheus also have the issues present in GPT3.5 described before. GPT-4 performs better in the directional expectation test compared to GPT-3.5, but surprisingly, it exhibits worse performance in the invariance test, especially showing severe sensitivity to grammatical and conflict-related perturbations about Grammaticality and Non-contradiction. On the other hand, Prometheus performs the worst in both tests, basically failing to differentiate between various aspects, which may be due to its small model size and the training data constructed by GPT-4.

\section{Discussions}

To investigate the failures of LLMs in our attack tests, we conduct some extended experiments with detailed aspect definitions. We retain only the definition or term, or even use the empty criterion, with the results presented in Figure~\ref{fig:63} and \ref{fig:64} in the appendix, which still exhibit convergence. It indicates that the improper sensitivity of LLMs to grammaticality and contradiction is likely derived from the default evaluation behaviors inherent in LLMs. They will be cumulative, regardless of whether the current criteria are unclear or unrelated to those two aspects. Moreover, the detailed aspect definitions indeed have effects for aspects whose terms are not commonly-used in NLG evaluation, like non-hallucination.

Furthermore, we have attempted different empirical methods to intervene in LLM-based evaluation to mitigate these issues. Specifically, the most intuitive solution is using explicit instructions to require LLMs to ensure the evaluation relies solely on the given criteria. However, there are only slight and unstable improvements. So we further considered the ideas of Chain of Thought (CoT) to decompose the evaluation process. One method was inspired by Multidimensional Quality Metrics (MQM), which first identified relevant issues based on the criteria, then conducted the evaluation based on the identified issues and their severity. Another related method required LLMs to provide the preliminary evaluation, then utilized LLMs themselves to check if the evaluation strictly adhered to the given criteria, and finally offered an improved version. Moreover, we also attempted to provide more comprehensive background knowledge for LLMs, such as by including other aspect definitions in the prompt. Although these methods show varying degrees of improvement, none of them has a generally significant effect, and we believe that they do not fundamentally solve the reliability issues in LLM-based evaluation, which are quite stubborn and challenging and necessitate more systematic research.

To cover a more diverse range of descriptions of quality criteria, we also conduct human evaluation and LLM-based evaluation with descriptions selected from existing papers. We find that ambiguous expressions play a similar role in both human evaluations and LLM-based evaluation as less informative descriptions designed by us. Inconsistent conceptualizations (e.g. a description mixing Fluency and Grammaticality) can alter human judgments on related aspect-targeted perturbations, and a similar but weaker effect exists in LLM-based evaluation.

\section{Related Works}

\paragraph{Diagnostic Tests for NLG Evaluation Metrics.} Recent studies have highlighted issues with NLG evaluation metrics through synthetic perturbations, showing their scores often diverge from human judgments \citep{DBLP:conf/emnlp/SaiDSMK21} and some of their blind spots \citep{DBLP:conf/acl/HeZ0KCGT23}. Moreover, some studies focused on diagnostic tests for single tasks or specific aspects, such as translation \citep{DBLP:conf/emnlp/KarpinskaRTSGI22}, summarization \citep{DBLP:conf/emnlp/ErnstSD023}, story generation \citep{DBLP:conf/emnlp/XieLCL23}, and factuality \citep{DBLP:conf/emnlp/Chen0Q21}. Notably, \citet{DBLP:journals/corr/abs-2312-15407} explored the robustness of LLM-based dialogue evaluators using perturbation strategies, while \citet{DBLP:journals/corr/abs-2305-14658} highlighted their inability to judge closed-ended responses without references under adversarial conditions. Neither study addressed varying expressions or distinctions among evaluation aspects.

\paragraph{Analyzing the limitations of LLM-based evaluators.} \citet{DBLP:conf/emnlp/ShenCNYB23} pointed out the order of the two texts affects evaluation results when ChatGPT and GPT-4 are used as comparison-based evaluators. LLM-based evaluators also prefer longer responses \citep{DBLP:journals/corr/abs-2306-05685} and responses generated by themselves \citep{DBLP:conf/emnlp/LiuIXWXZ23}. \citet{DBLP:conf/emnlp/ShenCNYB23} discovered that the performance of ChatGPT on summarization evaluation varies on different systems and aspects. \citet{DBLP:journals/corr/abs-2309-07462} stated that LLM-based evaluators may have more biases in non-Latin languages. It is worth mentioning that \citet{DBLP:conf/emnlp/XuWPSFWL23} use GPT-4 to identify multiple failure modes in the explanations generated by the trained evaluator, though the failure modes cannot be used directly for aspect-specific evaluation.

\paragraph{NLG Quality Criteria.} In human evaluation, \citet{DBLP:conf/inlg/BelzMH20} proposed a classification system based on the property of quality criteria to support comparability. \citet{DBLP:conf/inlg/HowcroftBCGHMMM20} demonstrate that different descriptions of quality criteria can be mapped to normalized criteria. In LLM-based NLG evaluation, researchers have attempted to automatically generate quality standards more suitable for LLMs. \citet{DBLP:journals/corr/abs-2309-13308} let LLMs draft expressions of quality criteria based on examples with human ratings. \citet{DBLP:journals/corr/abs-2309-13633} utilized LLMs to review user-defined criteria and offered suggestions for disambiguation, merging, and splitting. Furthermore, some studies aim to improve LLMs' ability to evaluate specific aspects through chain-of-thoughts \citep{DBLP:journals/corr/abs-2312-10355} and instruction tuning \citep{DBLP:journals/corr/abs-2311-08788}.

\section{Conclusions}

In this work, we conduct fine-grained pertubation attack tests guided by the classification system and human judgments on LLMs to reveal their actual performance in NLG evaluation. Our findings can be concluded as follows: \textbf{1)} The performance of LLMs in our perturbation tests deviates significantly from expectations, with both unawareness and oversensitivity in some aspects. \textbf{2)} The different levels of detail in criteria almost do not change the evaluation behaviors of LLMs, except for criteria with uncommon terms like non-hallucination. \textbf{3)} The oversensitivity may be inherent in LLMs and not caused by the problems within criterion descriptions, due to its still existing in evaluations with empty criteria. \textbf{4)} The confusion issues are so stubborn that even the explicit instructions to hint LLMs to consider or not consider the specific problems cannot have obvious effects.
These results show that LLM-based evaluation is not that reliable across different evaluation aspects. Therefore, in-depth analysis of the aforementioned problems in LLMs and effective methods for improving the evaluation capabilities of LLMs are necessary and worth exploring in future research.

\section*{Limitations}
Our summarized classification system and designed perturbation attacks are mainly applicable to the commonly used aspects in closed-end text generation tasks. So our work does not include aspects with strong subjectivity, such as interestingness, which can be further explored in future work.

Due to limited resources, the domains and task types covered in our experiments are limited. The lengths of source documents and texts to be evaluated are generally a few hundred words, and the data we use is in English. Therefore, we cannot guarantee the same conclusions for long texts, other languages, or data from special domains.

We make extensive use of APIs from GPT-3.5 and GPT-4 for constructing data and testing, which incurs significant costs. This may discourage others from replicating these experiments, but we have released all the resources and data to facilitate related research.

\section*{Acknowledgements}
This work was supported by National Key R\&D Program of China (2021YFF0901502), Beijing Science and Technology Program (Z231100007423011), Ant Group Research Fund and Key Laboratory of Science, Technology and Standard in Press Industry (Key Laboratory of Intelligent Press Media Technology). We appreciate the anonymous reviewers for their helpful comments, and everyone who has provided assistance in this work. Xiaojun Wan is the corresponding author.

\bibliography{custom}

\appendix

\section{Details for Preliminary Study}
\label{sec:preliminary}

\begin{table*}[!htp]
\centering
\begin{tabular}{lccccc}
\toprule
Evaluation Form & Fluency & Coherence & Relevance & Consistency & Average \\ 
\midrule
Score only & 0.362 & 0.437 & 0.450 & 0.352 & 0.400 \\
Score only (T=0) & 0.344 & 0.353 & 0.394 & 0.321 & 0.353 \\
Score only (1-shot) & 0.342 & 0.323 & \textbf{0.502} & 0.401 & 0.392 \\
Score only (5-shot) & 0.415 & 0.399 & 0.471 & 0.538 & 0.456 \\
Rate-explain & 0.371 & 0.532 & 0.475 & 0.439 & 0.454 \\
Rate-explain (T=0) & 0.343 & 0.479 & 0.438 & 0.415 & 0.428 \\
Analyze-rate & 0.406 & \textbf{0.581} & 0.501 & \textbf{0.573} & \textbf{0.515} \\
Analyze-rate (T=0) & 0.367 & 0.525 & 0.362 & 0.468 & 0.431 \\
Analyze-rate (1-shot) & 0.311 & 0.423 & 0.334 & 0.420 & 0.372 \\
Analyze-rate (5-shot) & \textbf{0.474} & 0.505 & 0.443 & 0.526 & 0.487 \\
\midrule
Analyze-rate (GPT-4) & 0.617 & 0.572 & 0.588 & 0.752 & 0.632 \\
Analyze-rate (Prometheus) & 0.298 & 0.352 & 0.376 & 0.343 & 0.342 \\
\bottomrule
\end{tabular}
\caption{Pearson correlation coefficients between scores generated by different LLMs with different settings and forms of evaluation and human judgments on SummEval.}
\label{tab:6}
\end{table*}

\begin{figure}[h]
\centering
\includegraphics[width=0.48\textwidth]{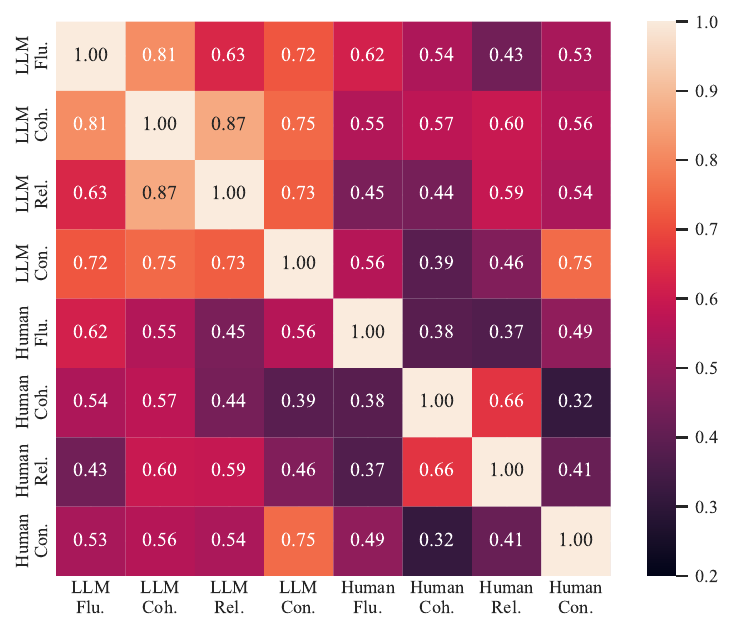}
\caption{Pearson correlation coefficients between scores generated by GPT-4 or human annotators on four criteria in SummEval.}
\label{fig:3}
\end{figure}

\begin{figure}[h]
\centering
\includegraphics[width=0.48\textwidth]{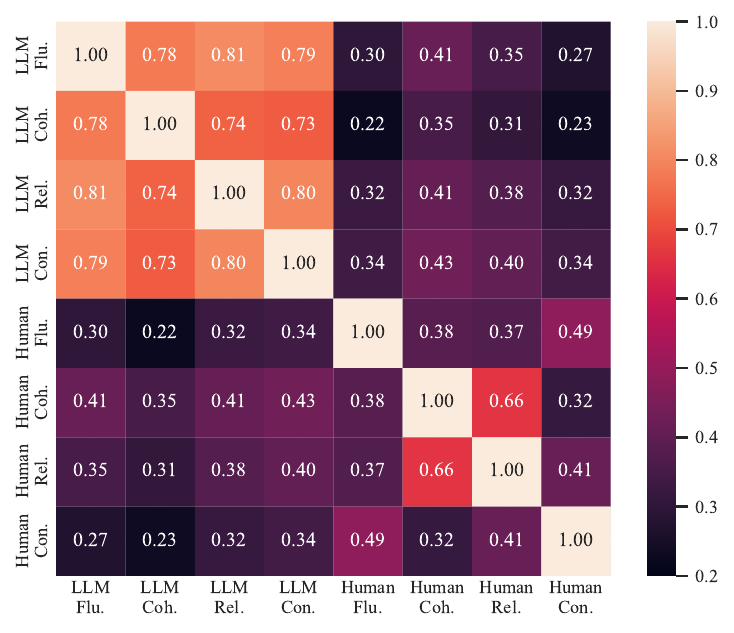}
\caption{Pearson correlation coefficients between scores generated by Prometheus or human annotators on four criteria in SummEval.}
\label{fig:4}
\end{figure}

We follow the evaluation forms proposed by~\citet{DBLP:conf/emnlp/ChiangL23}, including scoring modes, temperatures, and sampling settings. For more information, please refer to their paper and repository~\citep{liu2023geval}. As for the prompts and instructions used for evaluation, we employ those from~\citet{DBLP:conf/emnlp/ChiangL23} for GPT-3.5 and GPT-4, while those provided by~\citet{DBLP:journals/corr/abs-2310-08491} for Prometheus. The complete results are included in Table~\ref{tab:6} with the default settings where the sampling number is 20, and the temperature is set to 1 with zero-shot evaluations. Multiple results are post-processed and averaged to be the final scores. During few-shot evaluations, the selected demonstrations possess human labels of a uniform distribution, and analyses are correspondingly generated using GPT-3.5 if required. Moreover, the correlation matrices for GPT-4 and Prometheus are shown in Figure~\ref{fig:3} and Figure~\ref{fig:4}, respectively. Although the performance of GPT-4 is significantly better than that of GPT-3.5, its confusion issues seem to be more severe than GPT-3.5; meanwhile, Prometheus not only performs the worst, but its confusion is also quite serious.

\section{Details for Perturbation Attacks}
\label{sec:perturbation}

We construct four relatively simple types of perturbations—sentence exchange, word exchange, spelling mistake, and sentence deletion—based on the corresponding rules, while the remaining 14 types are generated by GPT-4 in 10-shot settings. We manually write these 140 demonstrations and carefully check them to ensure they meet the requirements of the corresponding perturbation attacks. Then, we prompt GPT-4 with these demonstrations as well as the detailed instructions to enable GPT-4 to generate the desired perturbed texts as closely as possible. And the corresponding instructions for GPT-4 and codes for rule-based constructions can be found in our released resources. In addition, we show different criteria for each aspect as described in Section 3.3 in Table~\ref{tab:20}-\ref{tab:18}.

\begin{table*}[!htp]
  \centering\small
  \begin{tabular}{lp{13cm}}
  \toprule
  \textbf{Type} & \textbf{Criterion (Term and Definition)}\\
  \midrule
  Default & Overall quality: It measures not only the quality of the target text itself, including writing and logic, but also how well the target text matches the required information of the source content according to the corresponding task.\\ \midrule
  Simplified & Overall quality: It measures whether the target text is well-written and logical, and matches the required points of the source content.\\ \midrule
  Term & Overall quality: It measures the overall quality of the target text.\\ \midrule
  Detailed & Overall quality: It measures not only the quality of the target text itself, including writing and logic, but also how well the target text matches the required information of the source content according to the corresponding task. Consider whether the target text is grammatically correct and naturally written, with clear meanings and good context-relatedness. Also consider whether all the information in the target text is supported by the source content and covers all and only the contents needed of the source content.\\ \midrule
  List & Overall quality: It measures not only the quality of the target text itself, including writing and logic, but also how well the target text matches the required information of the source content according to the corresponding task. \\ & Score 5: Good overall quality, with no errors of grammar, expression, content alignment required, and so on. \\ & Score 4: Only with some minor writing and content problems. \\ & Score 3: There are some obvious errors that affect the meaning and understanding of the target text, like unclear expressions and illogical context-relatedness. \\ & Score 2: Containing many major writing and logical errors and unmatched contents, but some good segments. \\ & Score 1: Poor overall quality, full of fragments that contain unrelated or untrue information and cannot be understood.\\ \midrule
  Selection1 & Overall (1-5) What is your overall impression of this target text?\\ & - A score of 1 (very bad). A completely invalid target text. It would be difficult to recover the source content from this. \\ & - A score of 2 (bad). Valid target text, but otherwise poor in quality. \\ & - A score of 3 (neutral) means this target text is neither good nor bad. This target text has no negative qualities, but no positive ones either. \\ & - A score of 4 (good) means this is a good target text, but falls short of being perfect because of a key flaw. \\ & - A score of 5 (very good) means this target text is good and does not have any strong flaws.\\ \midrule
  Selection2 & Overall quality: How good is the target text overall at representing the source content? If it’s hard to find ways to make the target text better, give the target text a high score. If there are lots of different ways the target text can be made better, give the text a low score. \\
  \bottomrule
  \end{tabular}
  \caption{Examples of different criterion descriptions for overall quality.}
  \label{tab:20}
\end{table*}

\begin{table*}[!htp]
  \centering\small
  \begin{tabular}{lp{13cm}}
  \toprule
  \textbf{Type} & \textbf{Criterion (Term and Definition)}\\
  \midrule
  Default & Readability: It measures the quality of both inter- and intra-sentences, are they grammatically correct and naturally written, with clear meanings and good context-relatedness and logic.\\ \midrule
  Simplified & Readability: It measures whether the target text is well-written, logical and clear.\\ \midrule
  Term & Readability: It measures whether the target text is readable.\\ \midrule
  Detailed & Readability: It measures the quality of both inter- and intra-sentences, are they grammatically correct and naturally written, with clear meanings and good context-relatedness and logic. Consider whether there are grammar errors, duplication, uncommon phrases and syntactic structures, unreasonable conjunctions, semantic inconsistency, and so on.\\ \midrule
  List & Readability: It measures the quality of both inter- and intra-sentences, are they grammatically correct and naturally written, with clear meanings and good context-relatedness and logic.\\ & Score 5: Entirely well readable, with no grammar errors, uncommon usages, or poor logic.\\ & Score 4: Only containing some minor writing or logical problems that basically do not affect reading.\\ & Score 3: Readable in general, with some obvious errors in grammar, collocations, or consistency.\\ & Score 2: There are major writing and logical problems, but some readable segments.\\ & Score 1: Not readable at all, full of fragments that cannot be understood.\\ \midrule
  Selection1 & Readability takes into account word and grammatical error rate to evaluate how fluent the target text language is.\\ \midrule
  Selection2 & Fluency: The target text sentences should be grammatically correct, easy to read and understand." \\
  \bottomrule
  \end{tabular}
  \caption{Examples of different criterion descriptions for readability.}
  \label{tab:14}
\end{table*}

\begin{table*}[!htp]
  \centering\small
  \begin{tabular}{lp{13cm}}
  \toprule
  \textbf{Type} & \textbf{Criterion (Term and Definition)}\\
  \midrule
  Default & Coherence: It measures the quality of all sentences collectively, do they make sense as a whole, with the context organized and connected logically. \\ \midrule
  Simplified & Coherence: It measures whether all the sentences are organized and connected logically. \\ \midrule
  Term & Coherence: It measures whether the target text is coherent. \\ \midrule
  Detailed & Coherence: It measures the quality of all sentences collectively, do they make sense as a whole, with the context organized and connected logically. Consider whether they have good context-relatedness with reasonable conjunctions, semantic consistency, and inter-sentence causal and temporal dependencies. \\ \midrule
  List & Coherence: It measures the quality of all sentences collectively, do they make sense as a whole, with the context organized and connected logically. \\ & Score 5: Entirely coherent, with good context-relatedness among all the sentences. \\ & Score 4: Only containing some minor illogical parts that basically do not affect overall coherency. \\ & Score 3: Coherent in general, with some obvious conflicting logical or inconsistent problems. \\ & Score 2: There are major unreasonable logic and semantic inconsistencies, but at least the related topic. \\ & Score 1: Not coherent at all, full of self-contradictory or unrelated content. \\ \midrule
  Selection1 & Discourse Coherence: Whether the target text is well organized, with the sentences smoothly connected and flow together logically and aesthetically? \\ \midrule
  Selection2 & Coherence: Description: Collective quality of all sentences. \\ \midrule
  Selection3 & Coherence: The rating measures the quality of all sentences collectively, to fit together and sound natural. Consider the quality of the target text as a whole. \\
  \bottomrule
  \end{tabular}
  \caption{Examples of different criterion descriptions for coherence.}
  \label{tab:11}
\end{table*}

\begin{table*}[!htp]
\centering\small
\begin{tabular}{lp{13cm}}
\toprule
\textbf{Type} & \textbf{Criterion (Term and Definition)}\\
\midrule
Default & Fluency: It measures the quality of individual sentences, are they grammatically correct, non-repetitive, and in accord with common English usage, with clear meanings. \\ \midrule
Simplified & Fluency: It measures whether individual sentences are grammatically correct and well-written. \\ \midrule
Term & Fluency: It measures whether the target text is fluent. \\ \midrule
Detailed & Fluency: It measures the quality of individual sentences, are they grammatically correct, non-repetitive, and in accord with common English usage, with clear meanings. Consider whether there are misspellings, tense errors, missing determiners, or more severe problems, such as duplication, unfamiliar phrases, complex syntactic structures, and missing components.\\ \midrule
List & Fluency: It measures the quality of individual sentences, are they grammatically correct, non-repetitive, and in accord with common English usage, with clear meanings. \\ & Score 5: Entirely fluent, grammatically correct, and well-written. \\ & Score 4: Only containing some minor non-fluent parts or grammatical errors that basically have no effect on fluency. \\ & Score 3: Fluent in general, with some obvious grammatical errors and unfamiliar phrases. \\ & Score 2: There are major grammatical errors, duplication, unfamiliar phrases and syntactic structures, and missing components, but some fluent segments. \\ & Score 1: Not fluent at all, full of meaningless fragments and unclear contents.\\ \midrule
Selection1 & Fluency: Description: Quality of individual sentences.\\ \midrule
Selection2 & Fluency: Whether the generated target text is grammatically correct.\\ \midrule
Selection3 & Fluency: The rating measures the quality of individual sentences, are they well-written and grammatically correct. Consider the quality of individual sentences. \\
\bottomrule
\end{tabular}
\caption{Examples of different criterion descriptions for fluency.}
\label{tab:10}
\end{table*}

\begin{table*}[!htp]
  \centering\small
  \begin{tabular}{lp{13cm}}
  \toprule
  \textbf{Type} & \textbf{Criterion (Term and Definition)}\\
  \midrule
  Default & Grammaticality: It measures whether the target text is grammatically correct without any lexical or syntax errors, regardless of its content and meaning. \\ \midrule
  Simplified & Grammaticality: It measures whether the target text has no grammatical errors. \\ \midrule
  Term & Grammaticality: It measures whether the target text is grammatical. \\ \midrule
  Detailed & Grammaticality: It measures whether the target text is grammatically correct without any lexical or syntax errors, regardless of its content and meaning. Consider whether the target text itself complies with the English standard usage and rules of grammar, such as tense errors, misspellings, incorrect prepositions, collocation misusages, and so on. \\ \midrule
  List & Grammaticality: It measures whether the target text is grammatically correct without any lexical or syntax errors, regardless of its content and meaning. \\ & Score 5: Entirely grammatically correct, following the rules of English grammar. \\ & Score 4: Basically grammatical, with a few minor grammar errors. \\ & Score 3: There are some obvious grammatical errors that affect the sentence’s expression. \\ & Score 2: Containing many severe grammatical errors whose originally intended usages even cannot be judged. \\ & Score 1: Not grammatical at all, full of grammar errors. \\ \midrule
  Selection1 & Grammar – ability to generate grammatically correct and fluent target texts. \\ \midrule
  Selection2 & Grammaticality measures whether the target text contains syntax errors. It refers to the conformity of the target text to the rules defined by the specific grammar of a language. \\ \midrule
  Selection3 & Correctness: Whether there are grammatical errors in the target text. \\
  \bottomrule
  \end{tabular}
  \caption{Examples of different criterion descriptions for grammaticality.}
  \label{tab:12}
\end{table*}

\begin{table*}[!htp]
\centering\small
\begin{tabular}{lp{13cm}}
\toprule
\textbf{Type} & \textbf{Criterion (Term and Definition)}\\
\midrule
Default & Simplicity: It measures whether the target text is sufficiently simple and easy for people who aren't very good at English to get the correct meaning.\\ \midrule
Simplified & Simplicity: It measures whether the target text is simple and easy to get the meaning.\\ \midrule
Term & Simplicity: It measures whether the target text is simple.\\ \midrule
Detailed & Simplicity: It measures whether the target text is sufficiently simple and easy for people who aren't very good at English to get the correct meaning. Consider whether the target text adopts simplified and common usage of phrases and sentences, and avoid any unfamiliar words or complicated syntactic structures.\\ \midrule
List & Simplicity: It measures whether the target text is sufficiently simple and easy for people who aren't very good at English to get the correct meaning.\\ & Score 5: Entirely simple, without any complicated words or syntactic structures.\\ & Score 4: Only containing a small number of unfamiliar words.\\ & Score 3: There are some uncommon phrases and complicated structures, which makes it a little hard to get the target text's meaning.\\ & Score 2: Despite some simple words, the target text has many unfamiliar phrases and complicated sentences.\\ & Score 1: Not simple at all, full of complex expressions that are quite difficult to read.\\ \midrule
Selection1 & The goal is to judge whether the target text is simpler than the source content. \\
\bottomrule
\end{tabular}
\caption{Examples of different criterion descriptions for simplicity.}
\label{tab:13}
\end{table*}

\begin{table*}[!htp]
  \centering\small
  \begin{tabular}{lp{13cm}}
  \toprule
  \textbf{Type} & \textbf{Criterion (Term and Definition)}\\
  \midrule
  Default & Adequacy: It measures whether the entire contents of the target text exactly match the required information of the source content without unnecessary points, according to the corresponding task.\\ \midrule
  Simplified & Adequacy: It measures how well the target text matches the required information of the source content.\\ \midrule
  Term & Adequacy: It measures whether the target text is adequate.\\ \midrule
  Detailed & Adequacy: It measures whether the entire contents of the target text exactly match the required information of the source content without unnecessary points, according to the corresponding task. Consider whether all the information contained in the target text is factually supported by the source content and covers all and only the contents that the task needs in the source content.\\ \midrule
  List & Adequacy: It measures whether the entire contents of the target text exactly match the required information of the source content without unnecessary points, according to the corresponding task. \\ & Score 5: Entirely adequate, the whole target text matches the required information of the source content. \\ & Score 4: Just containing some minor unmatched or unnecessary information. \\ & Score 3: There is some key information that cannot be supported by the source content or is not needed by the task. \\ & Score 2: Only a small part of the information in the target text matches the required information of the source content, with many incorrect points. \\ & Score 1: Not adequate at all, the target text is irrelevant and does not cover any content in the source content.\\ \midrule
  Selection1 & Adequacy is defined as how much information is preserved in the target text. A score of 1 would mean that the target text is meaningless and has no correlation with the source content. A score of 5 would mean the target text retains all of the information. \\ \midrule
  Selection2 & Adequacy: Description: How correct is the target text from the given source content. \\
  \bottomrule
  \end{tabular}
  \caption{Examples of different criterion descriptions for adequacy.}
  \label{tab:19}
\end{table*}

\begin{table*}[!htp]
\centering\small
\begin{tabular}{lp{13cm}}
\toprule
\textbf{Type} & \textbf{Criterion (Term and Definition)}\\
\midrule
Default & Faithfulness: It measures whether all the information contained in the target text is consistent with and factually supported by the source content.\\ \midrule
Simplified & Faithfulness: It measures whether the target text can be supported by the source content.\\ \midrule
Term & Faithfulness: It measures whether the target text is faithful.\\ \midrule
Detailed & Faithfulness: It measures whether all the information contained in the target text is consistent with and factually supported by the source content. Consider whether there are fabricated contents that cannot be inferred from the source content, including those contradicting the facts in the source content, and additional information that is not mentioned and cannot be verified by the source content.\\ \midrule
List & Faithfulness: It measures whether all the information contained in the target text is consistent with and factually supported by the source content.\\ & Score 5: Entirely faithful, all the facts in the target text can be inferred from the source content.\\ & Score  4: The target text is almost factually aligned with the source content but contains unsupported minor information.\\ & Score  3: There are some main but unverifiable or contradictory contents in the target text, according to the source content.\\ & Score  2: Only a small part of the information in the target text can be inferred from the source content.\\ & Score  1: Not faithful at all, the target text has nothing to do with the source content.\\ \midrule
Selection1 & Faithfulness: Whether the target text accords with the facts expressed in the source content.\\ \midrule
Selection2 & Consistency: The rating measures whether the facts in the target text are consistent with the facts in the source content. Consider whether the target text does reproduce all facts accurately and does not make up untrue information.\\ \midrule
Selection3 & Faithful or factually consistent: A target text is factually consistent to the source content if all the information in the target text can be supported by the source content. Common errors in model-generated target texts include information that is not mentioned or incorrect according to the input source content. Sometimes, the target text can be misleading because a crucial piece of information is absent. \\
\bottomrule
\end{tabular}
\caption{Examples of different criterion descriptions for faithfulness.}
\label{tab:15}
\end{table*}

\begin{table*}[!htp]
  \centering\small
  \begin{tabular}{lp{13cm}}
  \toprule
  \textbf{Type} & \textbf{Criterion (Term and Definition)}\\
  \midrule
  Default & Non-hallucination: It measures whether the target text contains no additional information that is not exactly mentioned and cannot be verified by the source content.\\ \midrule
  Simplified & Non-hallucination: It measures whether the target text is verifiable according to the source content.\\ \midrule
  Term & Non-hallucination: It measures whether the target text has no hallucinations.\\ \midrule
  Detailed & Non-hallucination: It measures whether the target text contains no additional information that is not exactly mentioned and cannot be verified by the source content. Consider whether there are contents other than the source content that cannot be proven correct or incorrect based on the source content, or even are unrelated to the source content.\\ \midrule
  List & Non-hallucination: It measures whether the target text contains no additional information that is not exactly mentioned and cannot be verified by the source content. \\ & Score 5: No hallucinations, all the facts in the target text can be proven correct or incorrect based on the source content. \\ & Score 4: Just containing a few unverifiable but unimportant contents. \\ & Score 3: There are some non-negligible contents other than the source content, leading to distorted meanings. \\ & Score 2: Almost all the contents of the target text are unverifiable based on the source content, except for several facts. \\ & Score 1: Full of hallucinations, the target text cannot be verified by the source content at all.\\ \midrule
  Selection1 & Hallucination error: Fabricated content that does not occur in the source content.\\ \midrule
  Selection2 & Not enough info: The target text information is not relevant or not sufficient to support/refute the source content. \\
  \bottomrule
  \end{tabular}
  \caption{Examples of different criterion descriptions for non-hallucination.}
  \label{tab:17}
  \end{table*}

\begin{table*}[!htp]
\centering\small
\begin{tabular}{lp{13cm}}
\toprule
\textbf{Type} & \textbf{Criterion (Term and Definition)}\\
\midrule
Default & Non-contradiction: It measures whether the target text contains no information that definitely contradicts certain contents of the source content.\\ \midrule
Simplified & Non-contradiction: It measures whether the target text does not contradict the source content.\\ \midrule
Term & Non-contradiction: It measures whether the target text has no contradictions.\\ \midrule
Detailed & Non-contradiction: It measures whether the target text contains no information that definitely contradicts certain contents of the source content. Consider whether there are contradictory contents such as incorrect entities, different expressions that distort the original meaning, false concatenation of crucial information from different places of the source content, and so on.\\ \midrule
List & Non-contradiction: It measures whether the target text contains no information that definitely contradicts certain contents of the source content.\\ & Score  5: No contradictions, all the facts in the target text do not conflict with the source content.\\ & Score  4: Just containing a few contradictory but unimportant contents.\\ & Score  3: There is some main information, like key entities, contradicting the source content, leading to distorted meanings.\\ & Score  2: Almost all the contents of the target text conflict with the source content, except for several facts.\\ & Score  1: Entirely contradictory, all the facts in the target text do contradict the source content.\\ \midrule
Selection1 & Contradiction, whether the target text contains any pieces of information that are contradicting the given source content or not. \\
\bottomrule
\end{tabular}
\caption{Examples of different criterion descriptions for non-contradiction.}
\label{tab:16}
\end{table*}

\begin{table*}[!htp]
\centering\small
\begin{tabular}{lp{13cm}}
\toprule
\textbf{Type} & \textbf{Criterion (Term and Definition)}\\
\midrule
Default & Informativeness: It measures how much required information of the source content is contained in the target text, according to the corresponding task. \\ \midrule
Simplified & Informativeness: It measures how well the target text covers required contents of the source content. \\ \midrule
Term & Informativeness: It measures whether the target text is informative. \\ \midrule
Detailed & Informativeness: It measures how much required information of the source content is contained in the target text, according to the corresponding task. Consider how well the target text correctly covers the contents that the task needs in the source content, which may be necessary information and key points or the entire content. \\ \midrule
List & Informativeness: It measures how much required information of the source content is contained in the target text, according to the corresponding task. \\ & Score 5: Entirely informative, the target text covers all the required information of the source content. \\ & Score 4: The target text captures the main points and only misses minor required information of the source content. \\ & Score 3: There is some important information needed but not contained in the target text, which disturbs the source content’s meaning. \\ & Score 2: Only a few contents that the task needs in the source content can be found in the target text. \\ & Score 1: Not informative at all, the target text does not involve any contents of the source content. \\ \midrule
Selection1 & Informativeness: Is important information captured? \\ \midrule
Selection2 & Informativeness: Whether the target text provides enough and necessary content coverage from the input source content. \\ \midrule
Selection3 & Coverage, i.e., whether the target text covers the whole source content or only part of the source content. \\
\bottomrule
\end{tabular}
\caption{Examples of different criterion descriptions for informativeness.}
\label{tab:18}
\end{table*}

\section{Details for Data and Test Settings}
\label{sec:data}

\subsection{Datasets}
\label{sec:appendix_dataset}

\begin{table*}[!htp]
  \centering\small
  \begin{tabular}{p{15cm}}
  \toprule
  \textbf{Prompts and Instructions} \\
  \midrule
  
  News Summarization \\
  \midrule
  
  Please summarize the following news article in three to four sentences. \\
  Note that you should use simple and short sentences, avoiding uncommon words and complex sentences. \\
  \\
  News Article: \\
  \{article\} \\
  Summary:\\ 
  \midrule
  
  Paraphrase Generation \\ 
  \midrule
  Please rephrase the following original text, maintaining exactly the same meanings.
  Note that you should use simple and short sentences, avoiding uncommon words and complex sentences. \\
  Note that you must not add any additional information and not delete or lose any information of the original text. \\
  \\
  Original Text: \\
  \{source\} \\
  Rephrasing: \\ 
  \midrule
  
  Table-to-text Generation \\
  \midrule
  
  Please modify the original description to contain exactly the same meanings as the table, and make the new description fluent and coherent. \\
  Note that you should use simple and short sentences, avoiding unnatural passive voices or intransitive verbs, uncommon words, and complex sentences. \\ 
  Note that you must not add any additional information and not delete or lose any information of the table. \\
  \\
  Table: \\
  \{table\} \\ 
  Original Description: \\
  \{ref\} \\
  New Description: \\
  \bottomrule
  \end{tabular}
  \caption{Prompts and instructions used for improving references by GPT-4.}
  \label{tab:8}
  \end{table*}

We select 200, 200, 300, and 300 pieces of data from CNN/Dailymail~\citep{DBLP:conf/nips/HermannKGEKSB15}, SAMSum~\citep{gliwa-etal-2019-samsum}, News Commentary~\footnote{\url{http://data.statmt.org/news-commentary/v18.1}}, and WebNLG~\citep{DBLP:conf/acl/GardentSNP17} respectively for tasks of news summarization, dialogue summarization, paraphrase generation, and table-to-text generation. However, many times the original references in common datasets for these tasks are not written by humans or are even missing. For instance, references for news summarization often employ the assemblage of highlights to build large-scale datasets but tend to be incoherent or contain information not present in the source news. 

To ensure the quality of references to better serve as the original texts in perturbation attack tests, we take advantage of the powerful GPT-4 to improve them, avoiding expert annotations that are hard to obtain. Specifically, depending on the condition of the original references in different tasks, we prompt GPT-4 to generate new references for news summarization and paraphrase generation, while the original references are modified and improved by GPT-4 in table-to-text generation. And we directly use the original references from SAMSum in dialogue summarization since they are human-written. 

As shown in the evaluation results of GPT-3.5, the original texts for perturbations, namely the references, are generally scored around 5 in all aspects, showing their high qualities. The prompts we use here are shown in Table~\ref{tab:8}. For each reference, we construct 18 different perturbed texts in various directions, leading to 19000 samples to be evaluated. Moreover, taking into account eleven different aspects and the different types of definitions involved with each, there are a total of 80 distinct evaluation criteria. Combined together, they constitute our data for experiments with the scale of 80*19000 = 1.52M.

Moreover, unlike traditional task-oriented NLG evaluation research, we focus on the general reliability of LLMs in NLG evaluation, so our considered perturbations and aspects are task-agnostic (not task-specific), which can be applied for many NLG tasks. We select four common NLG tasks in our experiments and find that LLMs show consistent evaluation issues across all tasks. Therefore, for the sake of clarity in our paper, we merged the data from different tasks to present our experiments and discussions. More details and results of specific tasks can be found in our released resources.

\subsection{LLMs}
\label{sec:appendix_llms}    
  \newcolumntype{L}[1]{>{\raggedright\arraybackslash}p{#1}}
  \begin{table*}[t]
    \centering\small
    \begin{tabular}{L{7cm}|L{7.5cm}}
    \toprule
    \multicolumn{2}{l}{\textbf{Prompts and Instructions}} \\
    \midrule
    GPT-3.5 and GPT-4 &  Prometheus\\
    \midrule
    You will be given an example of the source content and target text. The target text is generated from the source content according to the corresponding task type. \newline
    Your task is to rate the target text according to the evaluation criterion on a Likert scale from 1 to 5. Please make sure you read and understand these instructions carefully.\newline
    \newline
    Task Type Description:\newline
    \{task description\} \newline
    \newline
    Evaluation Criterion: \newline
    \{aspect description\} \newline
    \newline
    Example:\newline
    \newline
    Source Content: \newline
    \{source\} \newline
    \newline
    Target Text: \newline
    \{target\}\newline
    \newline
    Evaluation Form: \newline
    Answer by starting with "Analysis:" to analyze the given example regarding the evaluation criterion as concisely as possible, and then give the numeric rating on the next line by "Rating:". \newline
    \newline
    Your Answer:\newline
    & 
    \#\#\#Task Description: \newline
    An instruction (might include an Input inside it), a response to evaluate, a reference answer that gets a score of 5, and a score rubric representing a evaluation criteria are given.\newline
    1. Write a detailed feedback that assess the quality of the response strictly based on the given score rubric, not evaluating in general.\newline
    2. After writing a feedback, write a score that is an integer between 1 and 5. You should refer to the score rubric.\newline
    3. The output format should look as follows: "Feedback: (write a feedback for criteria) [RESULT] (an integer number between 1 and 5)"\newline
    4. Please do not generate any other opening, closing, and explanations.\newline
    \quad \newline
    \#\#\#The instruction to evaluate:\newline
    \{task description\}\newline
    \{source\}\newline
    \quad \newline
    \#\#\#Response to evaluate:\newline
    \{target\}\newline
    \quad \newline
    \#\#\#Reference Answer (Score 5):\newline
    \{reference\}\newline
    \quad \newline
    \#\#\#Score Rubrics:\newline
    \{[aspect description]\}\newline
    
    \#\#\#Feedback: \\
    \bottomrule
    \end{tabular}
    \caption{Prompts and instructions used for evaluation of GPT-3.5, GPT-4 and Prometheus.}
    \label{tab:9}
    \end{table*}

\begin{figure*}[!htp]
  \centering
  \includegraphics[width=\textwidth]{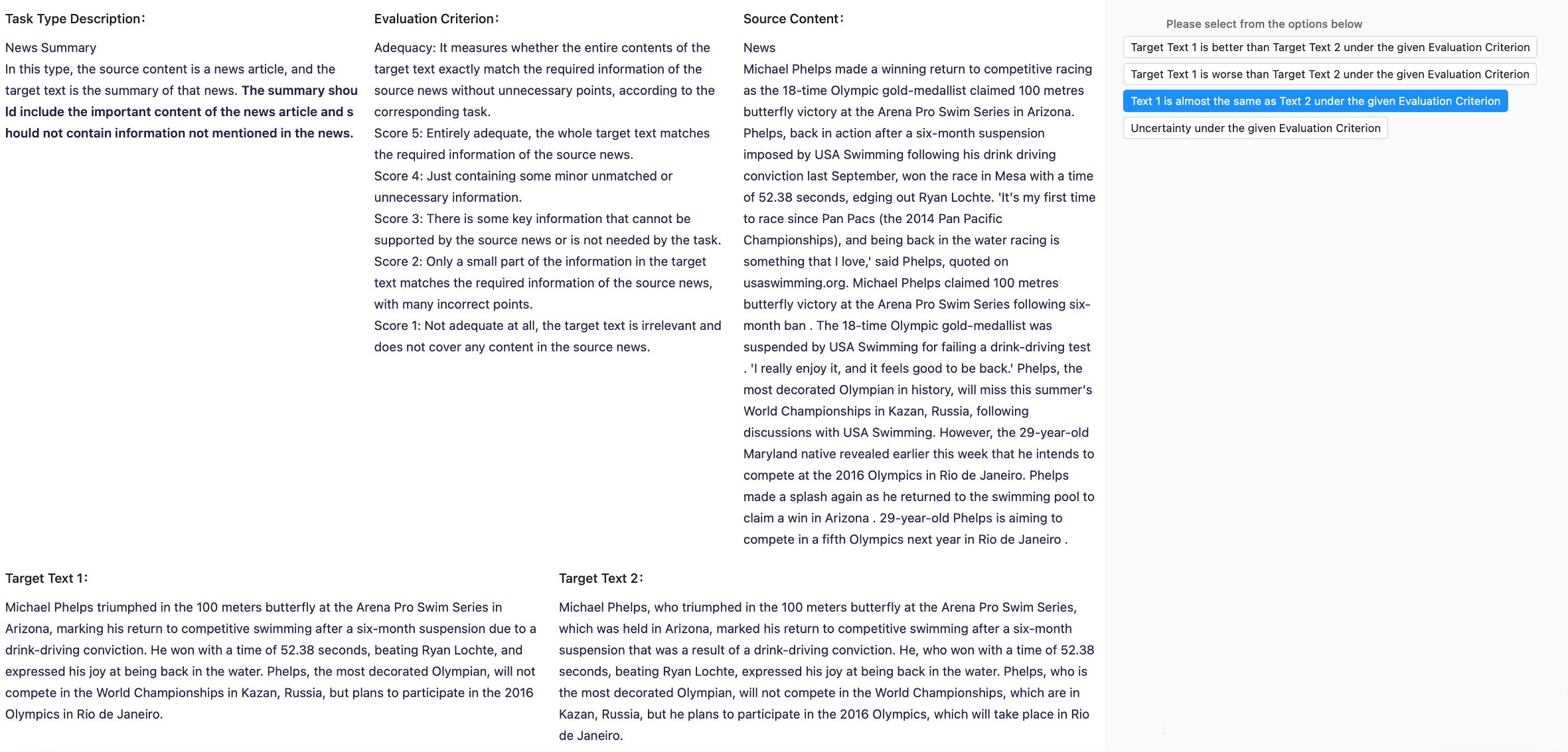}
  \caption{A screenshot of the annotation interface used in human evaluation.}
  \label{fig:annotation_interface}
  \end{figure*}

We test GPT-3.5 and GPT-4 with the API provided by OpenAI, whose versions are GPT-3.5 Turbo (1106) and GPT-4 Turbo (1106), respectively. On the other hand, Prometheus~\citep{DBLP:journals/corr/abs-2310-08491} has been proposed aiming to achieve performance close to that of proprietary LLMs like GPT-4 in NLG evaluation. They elaborately constructed 100K evaluations and feedbacks through GPT-4 and fine-tuned Llama-2-Chat-13B~\citep{DBLP:journals/corr/abs-2307-09288} on them, endowing the model with the capacity of evaluation across diverse and customized criteria. And we directly use the prompts provided by themselves~\citep{DBLP:journals/corr/abs-2310-08491} for the evaluation of Prometheus. For all three LLMs in our test, we follow the setting of~\citet{DBLP:conf/emnlp/ChiangL23} to conduct analysis before rating scores of 1–5 and set temperature and sampling number to 1.0 and 10, respectively in zeroshot, with prompts shown in Table~\ref{tab:9}.

\definecolor{myred}{HTML}{FFC8C8}
\definecolor{myblue}{HTML}{C8E7FF}
\begin{table*}[!htp]
\centering
\small
\begin{tabular}{llccccccccccc}
\toprule
\multicolumn{2}{c}{\textbf{Perturbation Attack}} & Flu. & Coh. & Gram. & Sim. & Read. & Fai. & Cont. & Hal. & Inf. & Ade. & All. \\
\midrule

\multirow{3}{*}{Flu.} & Repetition & \ding{55}\ \ding{51} & \cellcolor{myblue} \ding{55}\ \ding{55} & \cellcolor{myblue} \ding{55}\ \ding{55} & \cellcolor{myblue} \ding{55}\ \ding{55} & \ding{55}\ \ding{51} & \cellcolor{myblue} \ding{55}\ \ding{55} & \cellcolor{myblue} \ding{55}\ \ding{55} & \cellcolor{myblue} \ding{55}\ \ding{55} & \cellcolor{myblue} \ding{55}\ \ding{55} & \cellcolor{myblue} \ding{55}\ \ding{55} & \ding{55}\ \ding{51} \\
 & Passive Voice & \cellcolor{myred} \ding{51}\ \ding{51} & \cellcolor{myblue} \ding{55}\ \ding{55} & \cellcolor{myblue} \ding{55}\ \ding{55} & \cellcolor{myblue} \ding{55}\ \ding{55} & \cellcolor{myred} \ding{51}\ \ding{51} & \cellcolor{myblue} \ding{55}\ \ding{55} & \cellcolor{myblue} \ding{55}\ \ding{55} & \cellcolor{myblue} \ding{55}\ \ding{55} & \cellcolor{myblue} \ding{55}\ \ding{55} & \cellcolor{myblue} \ding{55}\ \ding{55} & \ding{55}\ \ding{51} \\
 & Inversion & \cellcolor{myred} \ding{51}\ \ding{51} & \cellcolor{myblue} \ding{55}\ \ding{55} & \cellcolor{myblue} \ding{55}\ \ding{55} & \cellcolor{myblue} \ding{55}\ \ding{55} & \cellcolor{myred} \ding{51}\ \ding{51} & \cellcolor{myblue} \ding{55}\ \ding{55} & \cellcolor{myblue} \ding{55}\ \ding{55} & \cellcolor{myblue} \ding{55}\ \ding{55} & \cellcolor{myblue} \ding{55}\ \ding{55} & \cellcolor{myblue} \ding{55}\ \ding{55} & \cellcolor{myred} \ding{51}\ \ding{51} \\
\midrule

\multirow{2}{*}{Coh.} & Improper Connective & \cellcolor{myblue} \ding{55}\ \ding{55} & \cellcolor{myred} \ding{51}\ \ding{51} & \cellcolor{myblue} \ding{55}\ \ding{55} & \cellcolor{myblue} \ding{55}\ \ding{55} & \cellcolor{myred} \ding{51}\ \ding{51} & \cellcolor{myblue} \ding{55}\ \ding{55} & \cellcolor{myblue}  \ding{55}\ \ding{55}& \cellcolor{myblue} \ding{55}\ \ding{55} & \cellcolor{myblue} \ding{55}\ \ding{55} & \cellcolor{myblue} \ding{55}\ \ding{55} & \cellcolor{myred} \ding{51}\ \ding{51} \\
 & Sentence Exchange & \cellcolor{myblue} \ding{55}\ \ding{55} & \cellcolor{myred} \ding{51}\ \ding{51} & \cellcolor{myblue} \ding{55}\ \ding{55} & \cellcolor{myblue} \ding{55}\ \ding{55} & \cellcolor{myred} \ding{51}\ \ding{51} & \cellcolor{myblue} \ding{55}\ \ding{55} & \cellcolor{myblue} \ding{55}\ \ding{55} & \cellcolor{myblue} \ding{55}\ \ding{55} & \cellcolor{myblue} \ding{55}\ \ding{55} & \cellcolor{myblue} \ding{55}\ \ding{55} & \cellcolor{myred} \ding{51}\ \ding{51} \\
\midrule

\multirow{3}{*}{Gram.} & Incorrect Verb Form & \cellcolor{myred} \ding{51}\ \ding{51} & \cellcolor{myblue} \ding{55}\ \ding{55} & \cellcolor{myred} \ding{51}\ \ding{51} & \cellcolor{myblue} \ding{55}\ \ding{55} & \cellcolor{myred} \ding{51}\ \ding{51} & \cellcolor{myblue} \ding{55}\ \ding{55} & \cellcolor{myblue} \ding{55}\ \ding{55} & \cellcolor{myblue} \ding{55}\ \ding{55} & \cellcolor{myblue} \ding{55}\ \ding{55} & \cellcolor{myblue} \ding{55}\ \ding{55} & \cellcolor{myred} \ding{51}\ \ding{51} \\
 & Word Exchange & \cellcolor{myred} \ding{51}\ \ding{51}& \cellcolor{myblue} \ding{55}\ \ding{55} & \cellcolor{myred} \ding{51}\ \ding{51} & \cellcolor{myblue} \ding{55}\ \ding{55} & \cellcolor{myred} \ding{51}\ \ding{51} & \cellcolor{myblue} \ding{55}\ \ding{55} & \cellcolor{myblue} \ding{55}\ \ding{55} & \cellcolor{myblue} \ding{55}\ \ding{55} & \cellcolor{myblue} \ding{55}\ \ding{55} & \cellcolor{myblue} \ding{55}\ \ding{55} & \cellcolor{myred} \ding{51}\ \ding{51} \\
 & Spelling Mistake & \cellcolor{myred} \ding{51}\ \ding{51} & \cellcolor{myblue} \ding{55}\ \ding{55} & \cellcolor{myred} \ding{51}\ \ding{51} & \cellcolor{myblue} \ding{55}\ \ding{55} & \ding{55}\ \ding{51} & \cellcolor{myblue} \ding{55}\ \ding{55} & \cellcolor{myblue} \ding{55}\ \ding{55} & \cellcolor{myblue} \ding{55}\ \ding{55} & \cellcolor{myblue} \ding{55}\ \ding{55} & \cellcolor{myblue} \ding{55}\ \ding{55} & \cellcolor{myred} \ding{51}\ \ding{51} \\
\midrule

\multirow{2}{*}{Sim.} & Uncommon Phrase & \cellcolor{myblue} \ding{55}\ \ding{55} & \cellcolor{myblue} \ding{55}\ \ding{55} & \cellcolor{myblue} \ding{55}\ \ding{55} & \cellcolor{myred} \ding{51}\ \ding{51} & \cellcolor{myred} \ding{51}\ \ding{51} & \cellcolor{myblue} \ding{55}\ \ding{55} & \cellcolor{myblue} \ding{55}\ \ding{55} & \cellcolor{myblue} \ding{55}\ \ding{55} & \cellcolor{myblue} \ding{55}\ \ding{55} & \cellcolor{myblue} \ding{55}\ \ding{55} & \ding{55}\ \ding{51} \\
 & Complex Sentence & \cellcolor{myblue} \ding{55}\ \ding{55} & \cellcolor{myblue} \ding{55}\ \ding{55} & \cellcolor{myblue} \ding{55}\ \ding{55} & \cellcolor{myred} \ding{51}\ \ding{51} & \ding{55}\ \ding{51} & \cellcolor{myblue} \ding{55}\ \ding{55} & \cellcolor{myblue} \ding{55}\ \ding{55} & \cellcolor{myblue} \ding{55}\ \ding{55} & \cellcolor{myblue} \ding{55}\ \ding{55} & \cellcolor{myblue} \ding{55}\ \ding{55} & \ding{55}\ \ding{51} \\
\midrule
 
\multirow{3}{*}{Inf.} & Abbreviation & \cellcolor{myblue} \ding{55}\ \ding{55} & \cellcolor{myblue} \ding{55}\ \ding{55} & \cellcolor{myblue} \ding{55}\ \ding{55} & \cellcolor{myblue} \ding{55}\ \ding{55} & \cellcolor{myblue} \ding{55}\ \ding{55} & \cellcolor{myblue} \ding{55}\ \ding{55} & \cellcolor{myblue} \ding{55}\ \ding{55} & \cellcolor{myblue} \ding{55}\ \ding{55} & \cellcolor{myred} \ding{51}\ \ding{51} & \cellcolor{myred} \ding{51}\ \ding{51} & \cellcolor{myred} \ding{51}\ \ding{51} \\
 & Hypernym & \cellcolor{myblue} \ding{55}\ \ding{55} & \cellcolor{myblue} \ding{55}\ \ding{55} & \cellcolor{myblue} \ding{55}\ \ding{55} & \cellcolor{myblue} \ding{55}\ \ding{55} & \cellcolor{myblue} \ding{55}\ \ding{55} & \cellcolor{myblue} \ding{55}\ \ding{55} & \cellcolor{myblue} \ding{55}\ \ding{55} & \cellcolor{myblue} \ding{55}\ \ding{55} & \cellcolor{myred} \ding{51}\ \ding{51} & \cellcolor{myred} \ding{51}\ \ding{51} & \cellcolor{myred} \ding{51}\ \ding{51} \\
 & Sentence Deletion & \cellcolor{myblue} \ding{55}\ \ding{55} & \cellcolor{myblue} \ding{55}\ \ding{55} & \cellcolor{myblue} \ding{55}\ \ding{55} & \cellcolor{myblue} \ding{55}\ \ding{55} & \cellcolor{myblue} \ding{55}\ \ding{55} & \cellcolor{myblue} \ding{55}\ \ding{55} & \cellcolor{myblue} \ding{55}\ \ding{55} & \cellcolor{myblue} \ding{55}\ \ding{55} & \cellcolor{myred} \ding{51}\ \ding{51} & \cellcolor{myred} \ding{51}\ \ding{51} & \cellcolor{myred} \ding{51}\ \ding{51} \\
\midrule
 
\multirow{2}{*}{Hal.} & Complement & \cellcolor{myblue} \ding{55}\ \ding{55} & \cellcolor{myblue} \ding{55}\ \ding{55} & \cellcolor{myblue} \ding{55}\ \ding{55} & \cellcolor{myblue} \ding{55}\ \ding{55} & \cellcolor{myblue} \ding{55}\ \ding{55} & \cellcolor{myred} \ding{51}\ \ding{51} & \cellcolor{myblue} \ding{55}\ \ding{55} & \cellcolor{myred} \ding{51}\ \ding{51} & \cellcolor{myblue} \ding{55}\ \ding{55} & \cellcolor{myred} \ding{51}\ \ding{51} & \cellcolor{myred} \ding{51}\ \ding{51} \\
 & Continuation & \cellcolor{myblue} \ding{55}\ \ding{55} & \cellcolor{myblue} \ding{55}\ \ding{55} & \cellcolor{myblue} \ding{55}\ \ding{55} & \cellcolor{myblue} \ding{55}\ \ding{55} & \cellcolor{myblue} \ding{55}\ \ding{55} & \cellcolor{myred} \ding{51}\ \ding{51} & \cellcolor{myblue} \ding{55}\ \ding{55} & \cellcolor{myred} \ding{51}\ \ding{51} & \cellcolor{myblue} \ding{55}\ \ding{55} & \cellcolor{myred} \ding{51}\ \ding{51} & \cellcolor{myred} \ding{51}\ \ding{51} \\
\midrule
 
\multirow{3}{*}{Cont.} & Different Entity & \cellcolor{myblue} \ding{55}\ \ding{55} & \cellcolor{myblue} \ding{55}\ \ding{55} & \cellcolor{myblue} \ding{55}\ \ding{55} & \cellcolor{myblue} \ding{55}\ \ding{55} & \cellcolor{myblue} \ding{55}\ \ding{55} & \cellcolor{myred} \ding{51}\ \ding{51} & \cellcolor{myred} \ding{51}\ \ding{51} & \ding{51}\ \ding{55} & \cellcolor{myred} \ding{51}\ \ding{51} & \cellcolor{myred} \ding{51}\ \ding{51} & \cellcolor{myred} \ding{51}\ \ding{51} \\
 & Conflicting Fact & \cellcolor{myblue} \ding{55}\ \ding{55} & \cellcolor{myblue} \ding{55}\ \ding{55} & \cellcolor{myblue} \ding{55}\ \ding{55} & \cellcolor{myblue} \ding{55}\ \ding{55} & \cellcolor{myblue} \ding{55}\ \ding{55} & \cellcolor{myred} \ding{51}\ \ding{51} & \cellcolor{myred} \ding{51}\ \ding{51} & \ding{51}\ \ding{55} & \cellcolor{myred} \ding{51}\ \ding{51} & \cellcolor{myred} \ding{51}\ \ding{51} & \cellcolor{myred} \ding{51}\ \ding{51} \\
 & Negation & \cellcolor{myblue} \ding{55}\ \ding{55} & \cellcolor{myblue} \ding{55}\ \ding{55} & \cellcolor{myblue} \ding{55}\ \ding{55} & \cellcolor{myblue} \ding{55}\ \ding{55} & \cellcolor{myblue} \ding{55}\ \ding{55} & \cellcolor{myred} \ding{51}\ \ding{51} & \cellcolor{myred} \ding{51}\ \ding{51} & \ding{51}\ \ding{55} & \cellcolor{myred} \ding{51}\ \ding{51} & \cellcolor{myred} \ding{51}\ \ding{51} & \cellcolor{myred} \ding{51}\ \ding{51} \\

\bottomrule
\end{tabular}
\caption{Human judgments (left) and our expectations based on the classification system (right) for each pair of perturbation attacks and aspects with detailed definitions. \ding{51} presents that the item should be affected, while \ding{55} presents that the item should not be affected. And we identify two sets of \sethlcolor{myred}\hl{$S_T$} and \sethlcolor{myblue}\hl{$S_F$} on those items that have the consistent results.}
\label{tab:7}
\end{table*}

\subsection{Human annotation}
\label{sec:appendix_human}

\paragraph{Settings.}
To facilitate human annotators in comparing texts before and after perturbations, we use a comparative form in human evaluations. This involves displaying two texts simultaneously on the annotation interface, allowing them to judge their quality relationship based on the given description of the quality criterion, as shown in Figure \ref{fig:annotation_interface}. Specifically, considering we design different types of descriptions and select some quality criteria with ambiguous expressions from existing papers, to better record the uncertainty of human annotators facing quality criteria of varying detail, the available quality relationships they can choose include "better than" (A), "worse than" (B), "as well as" (C), and "uncertain" (D). All 40 annotators come from the company's professional data annotation department, have certificates of English proficiency, and are paid more than the local minimum wage. Due to limited resources, we sample one example from each of the four datasets (CNN/Dailymail, SAMSum, News Commentary, and WebNLG) for human annotation. Each sample was subjected to 18 types of perturbation attacks, resulting in 18 pairs of test samples with and without perturbations. We had 11 quality criteria in total, and for each criterion, besides 5 descriptions of varying detail we design, we also select 1-3 descriptions from existing papers, making up a total of 80 descriptions. To prevent interference from other descriptions, for a quality criterion, an annotator is exposed to at most one description. Specifically, we divide the 40 annotators into four groups of ten, with each group annotating all the data once, meaning each test sample is annotated by four annotators. For each group of annotators, an annotator needed to annotate all test samples under the 8 descriptions of different quality criteria, with the types of descriptions distributed as evenly as possible (e.g. an annotator would not annotate all descriptions of the "Term" type). The total volume of annotations was $4 \times 18 \times 80 \times 4 = 23040$. The entire annotation process takes about 20 days.

\paragraph{Results.}
We define the annotation consistency per sample as the proportion of options with the most annotations except for the "uncertain" (D) option. For example, if the options given by four annotators on a sample are $\{A, A, C, D\}$, and the annotation consistency is 0.5. The final annotation consistency is the average across all samples. We calculate the match rate (i.e. the proportion of human judgments about perturbations that match our expectations) in two ways. The result is shown in Table \ref{tab:human_designed_result}.

\section{Details for Experiments}
\label{sec:experiments}

\subsection{Comparison with non-LLM Evaluation Metrics}

It may also be interesting to bring back non-LLM automatic evaluation metrics to compare the performance and reliability across perturbation schemes like \citet{DBLP:conf/emnlp/SaiDSMK21}. However, we need to point out that our work focuses on the understanding and execution capabilities of LLM-based evaluators regarding different aspects and criteria. Only when an evaluation metric can support evaluating with customized aspect definitions is it possible to study whether it confuses different aspects. However, most non-LLM evaluation metrics are designed for overall evaluation, such as ROUGE and BERTScore; while a few are designed for specific aspects and cannot accept criteria or other aspects as inputs, like FactCC, which targets summary faithfulness.

Table \ref{tab:non-LLM results} shows the performance of six commonly used non-LLM evaluation metrics as baselines, including BLEU \citep{DBLP:conf/acl/PapineniRWZ02}, CHRF++ \citep{DBLP:conf/wmt/Popovic17}, ROUGE \citep{lin-2004-rouge}, COMET-QE \citep{DBLP:conf/emnlp/ReiSFL20}, BERTScore \citep{DBLP:conf/iclr/ZhangKWWA20}, and BLEURT \citep{DBLP:conf/acl/SellamDP20}. Due to the factors mentioned above, we can only conduct the directional expectation test (i.e., evaluation scores should decline) and compare with GPT-4 on the evaluation aspect of overall quality. The results show that rule-based metrics, especially BLEU, are generally more sensitive to these perturbations compared with model-based metrics.

\subsection{Other Results}

\begin{figure}[!htp]
\centering
\includegraphics[width=0.42\textwidth]{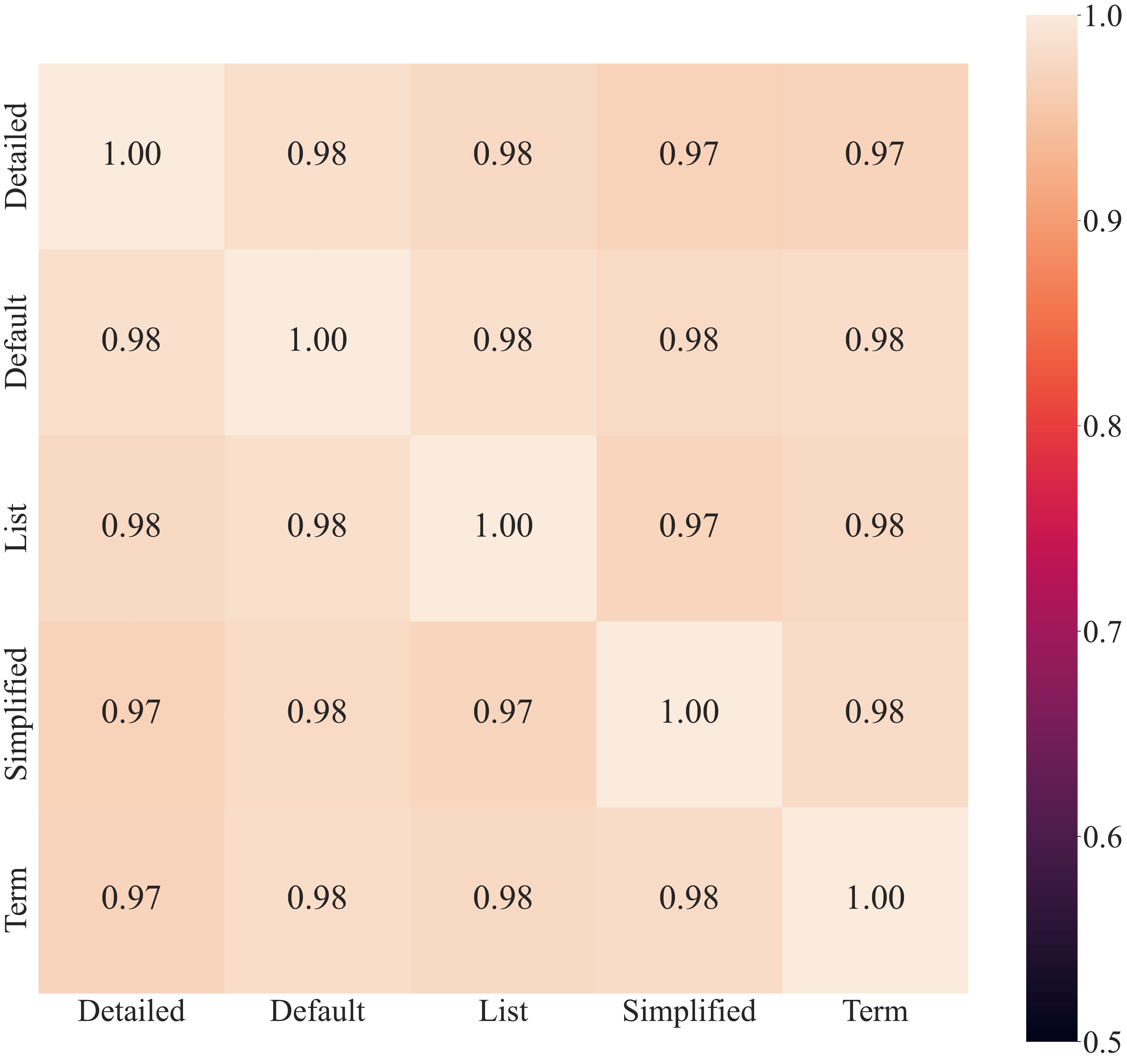}
\caption{Correlations between evaluation scores from GPT-3.5 across different levels of detail in aspect definitions.}
\label{fig:9}
\end{figure}

\begin{figure}[!htp]
\centering
\includegraphics[page=1,width=0.42\textwidth]{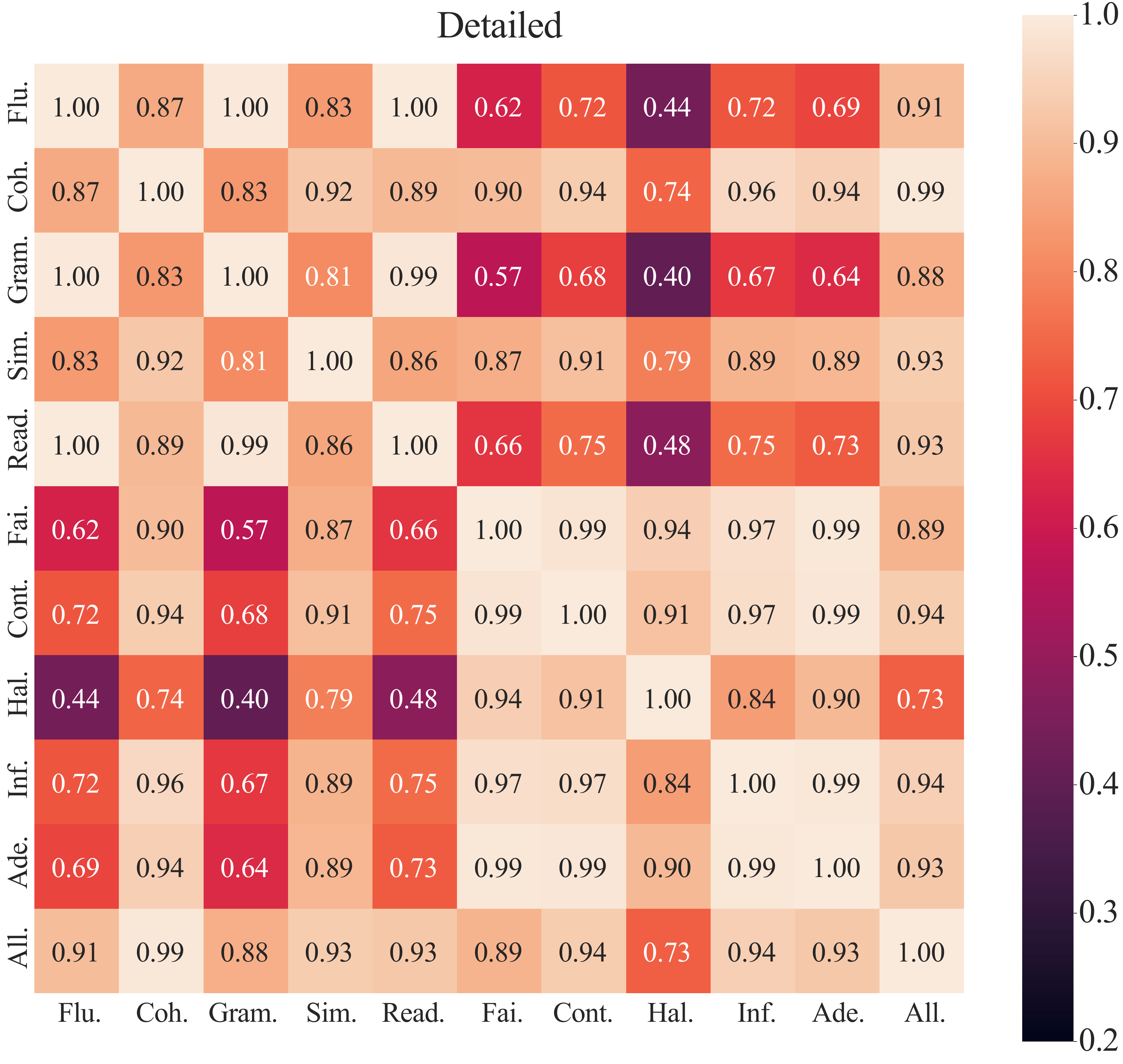}
\caption{Correlations between evaluation scores from GPT-3.5 across different aspects with the description of detailed type.}
\label{fig:58}
\end{figure}

\begin{figure}[!htp]
\centering
\includegraphics[page=2,width=0.42\textwidth]{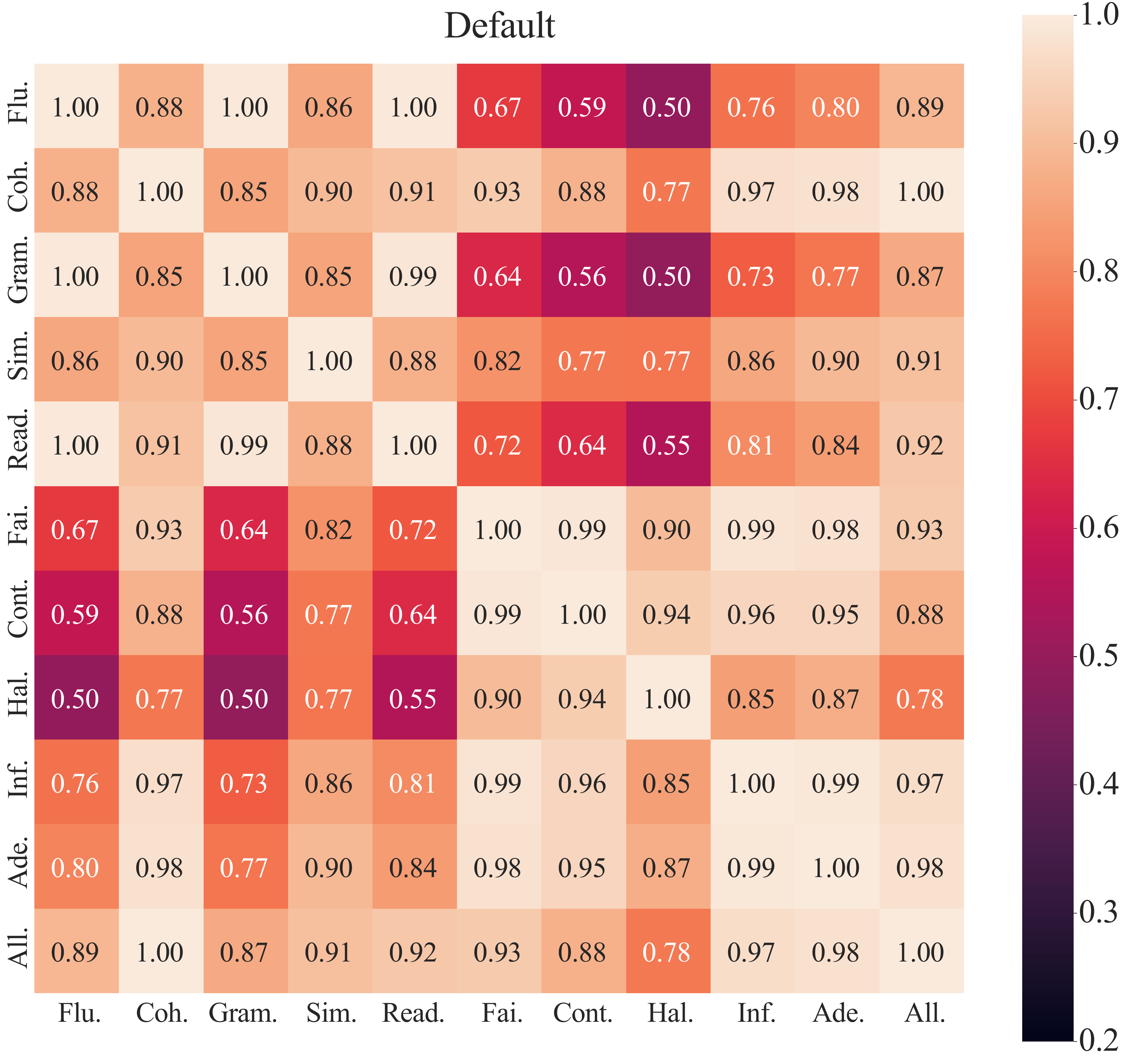}
\caption{Correlations between evaluation scores from GPT-3.5 across different aspects with the description of default type.}
\label{fig:59}
\end{figure}

\begin{figure}[!htp]
\centering
\includegraphics[page=3,width=0.42\textwidth]{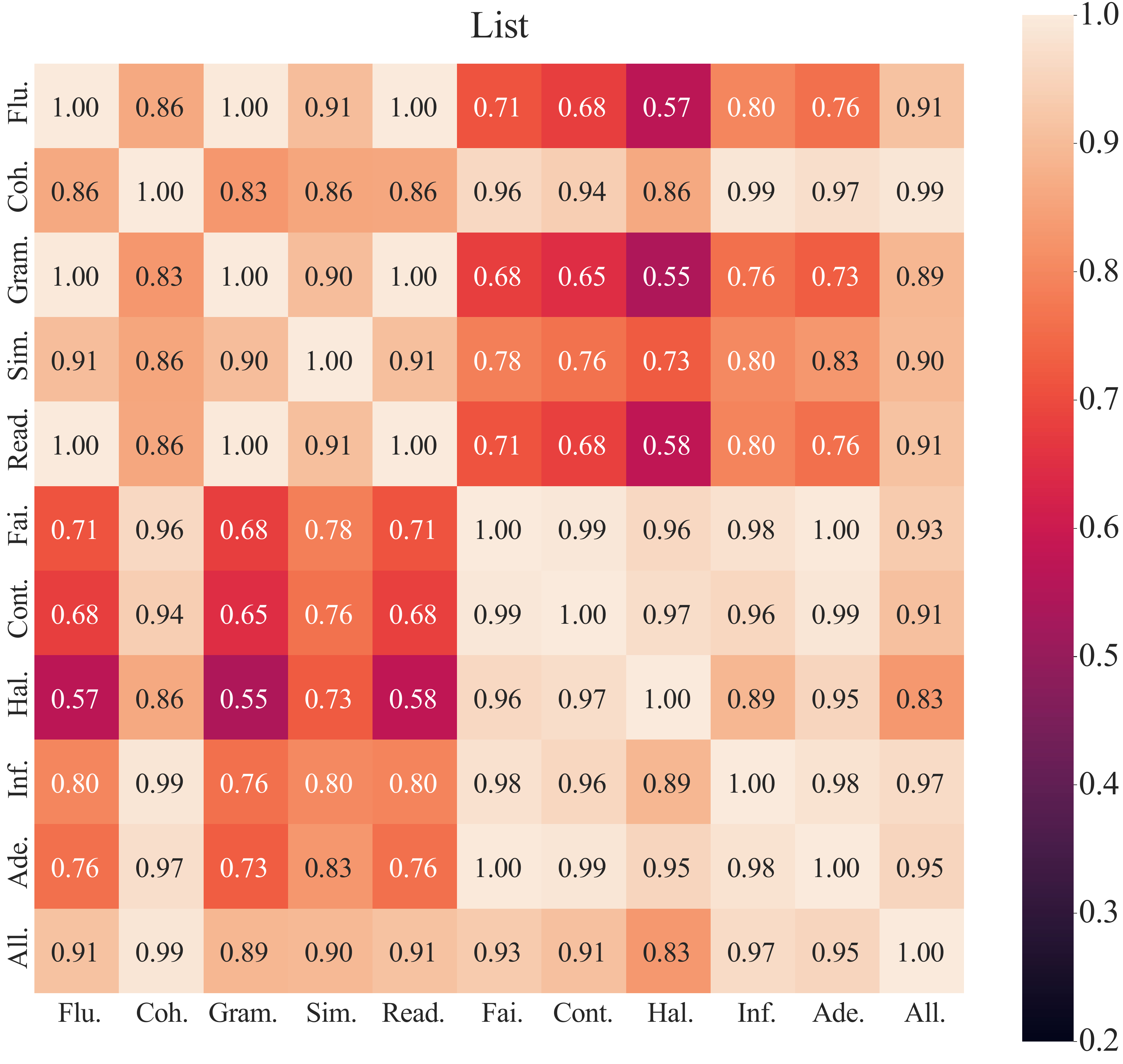}
\caption{Correlations between evaluation scores from GPT-3.5 across different aspects with the description of list type.}
\label{fig:60}
\end{figure}

\begin{figure}[!htp]
\centering
\includegraphics[page=4,width=0.42\textwidth]{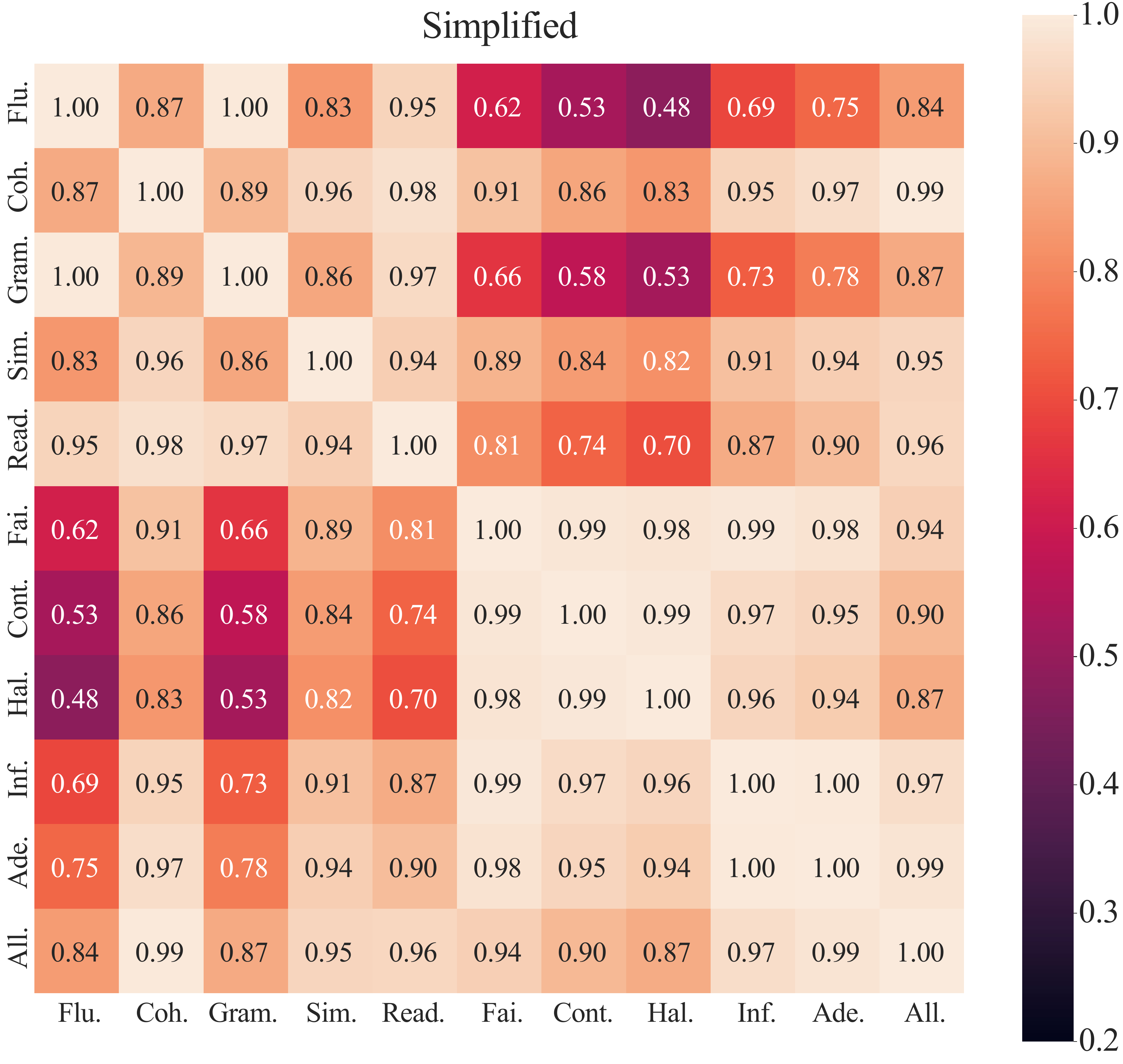}
\caption{Correlations between evaluation scores from GPT-3.5 across different aspects with the description of simplified type.}
\label{fig:61}
\end{figure}

\begin{figure}[!htp]
\centering
\includegraphics[page=5,width=0.42\textwidth]{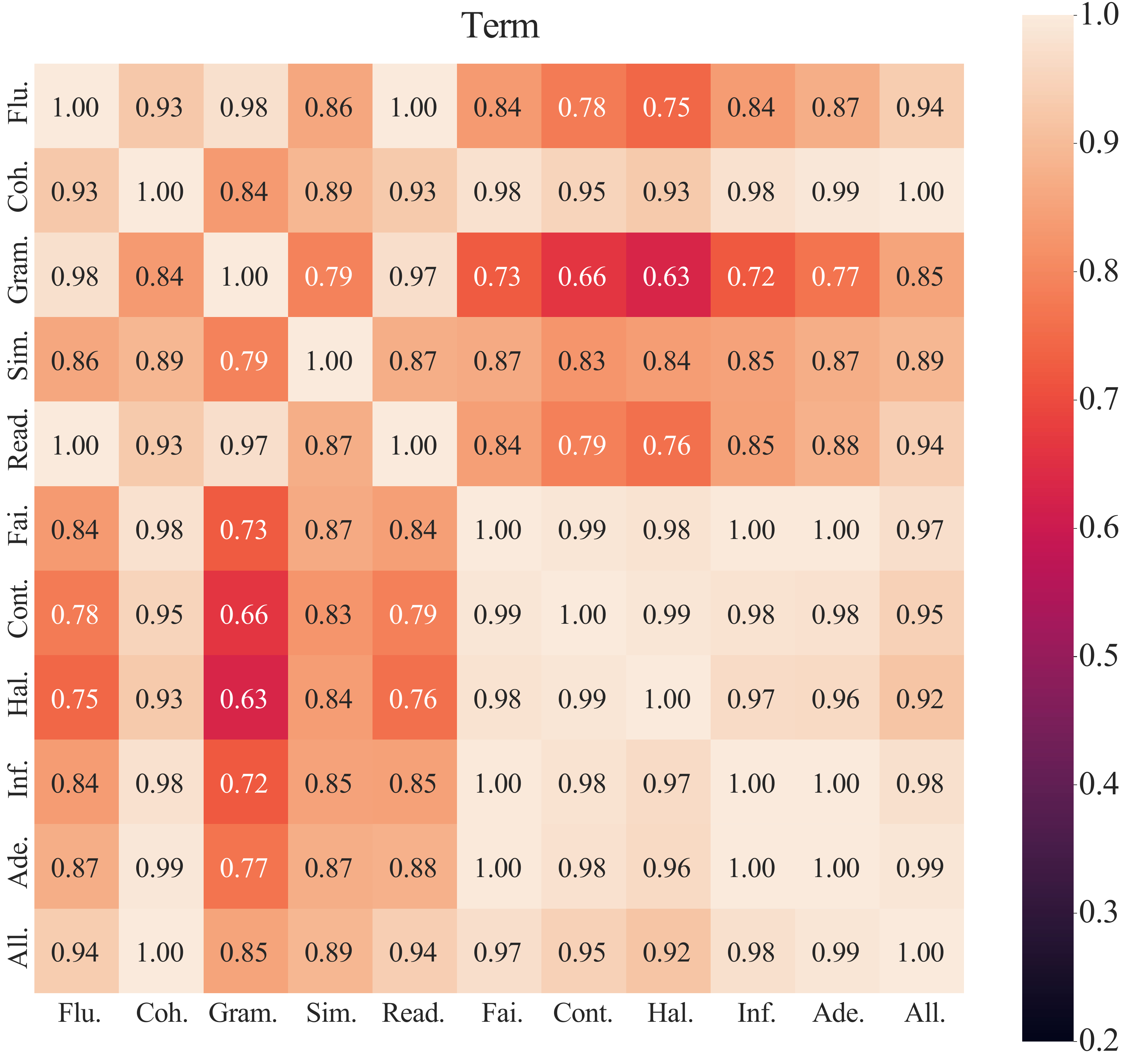}
\caption{Correlations between evaluation scores from GPT-3.5 across different aspects with the description of term type.}
\label{fig:62}
\end{figure}

\paragraph{Correlation matrices.} Figure \ref{fig:9} shows the correlations between evaluation scores from GPT-3.5 across five types of aspect definitions. Figure \ref{fig:58}-\ref{fig:62} show the correlations between evaluation scores from GPT-3.5 across different aspects given a description type.

\begin{figure*}[!htp]
\centering
\includegraphics[width=\textwidth]{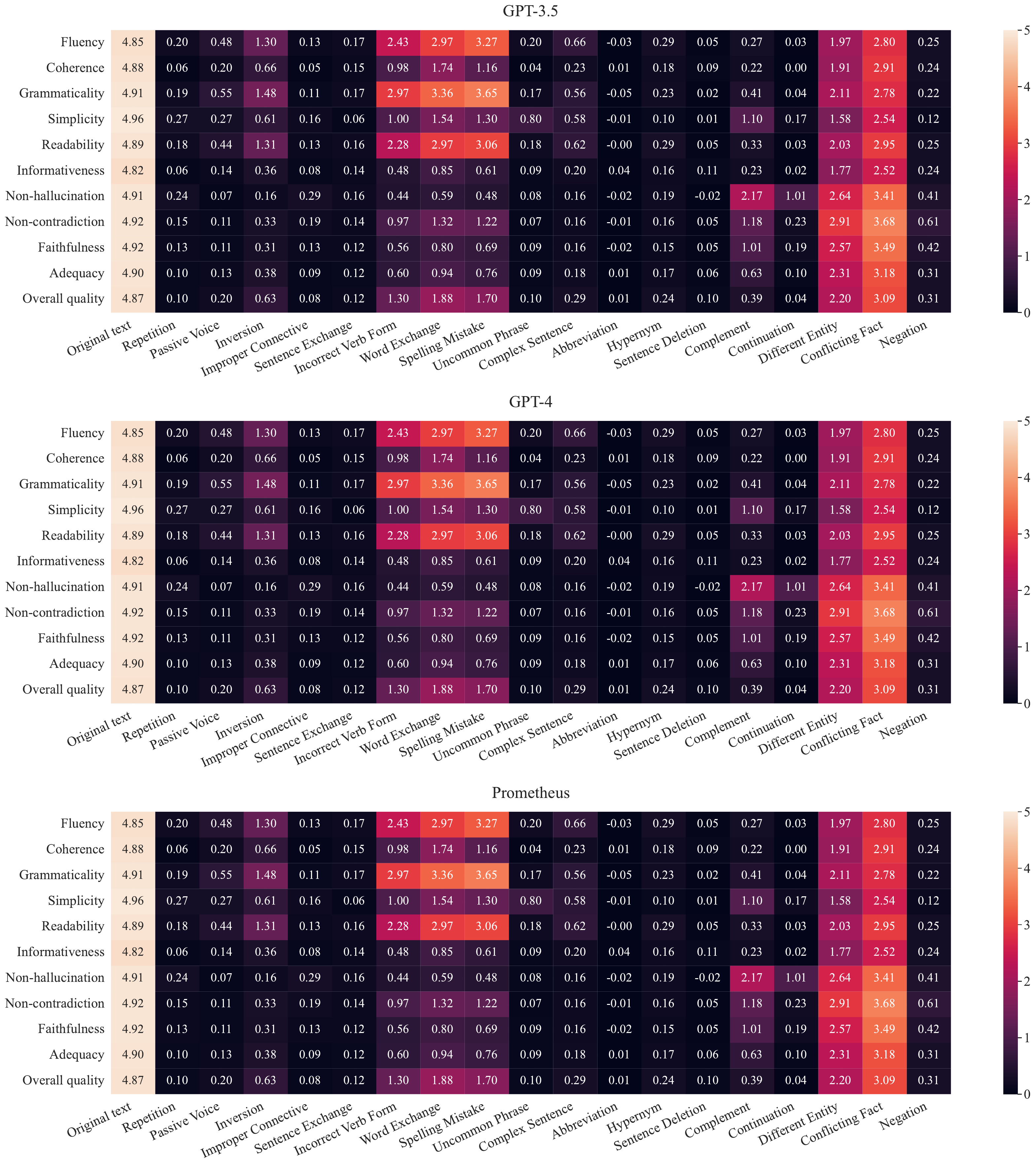}
\caption{The variances of evaluation scores from three LLMs between original texts and different perturbed texts with the description of detailed type.}
\label{fig:10}
\end{figure*}

\begin{figure*}[!htp]
\centering
\includegraphics[width=\textwidth]{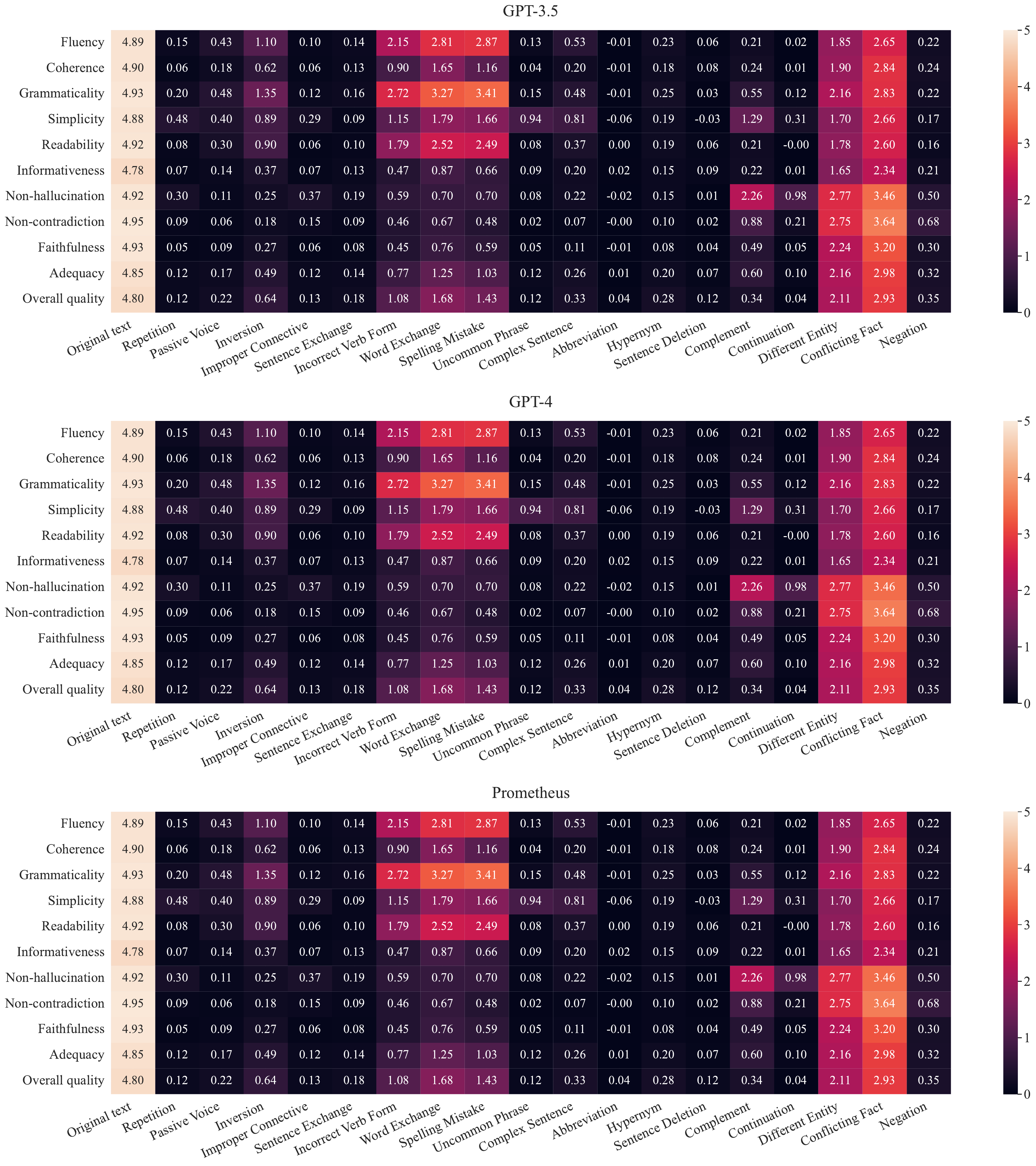}
\caption{The variances of evaluation scores from three LLMs between original texts and different perturbed texts with the description of default type.}
\label{fig:11}
\end{figure*}

\begin{figure*}[!htp]
\centering
\includegraphics[width=\textwidth]{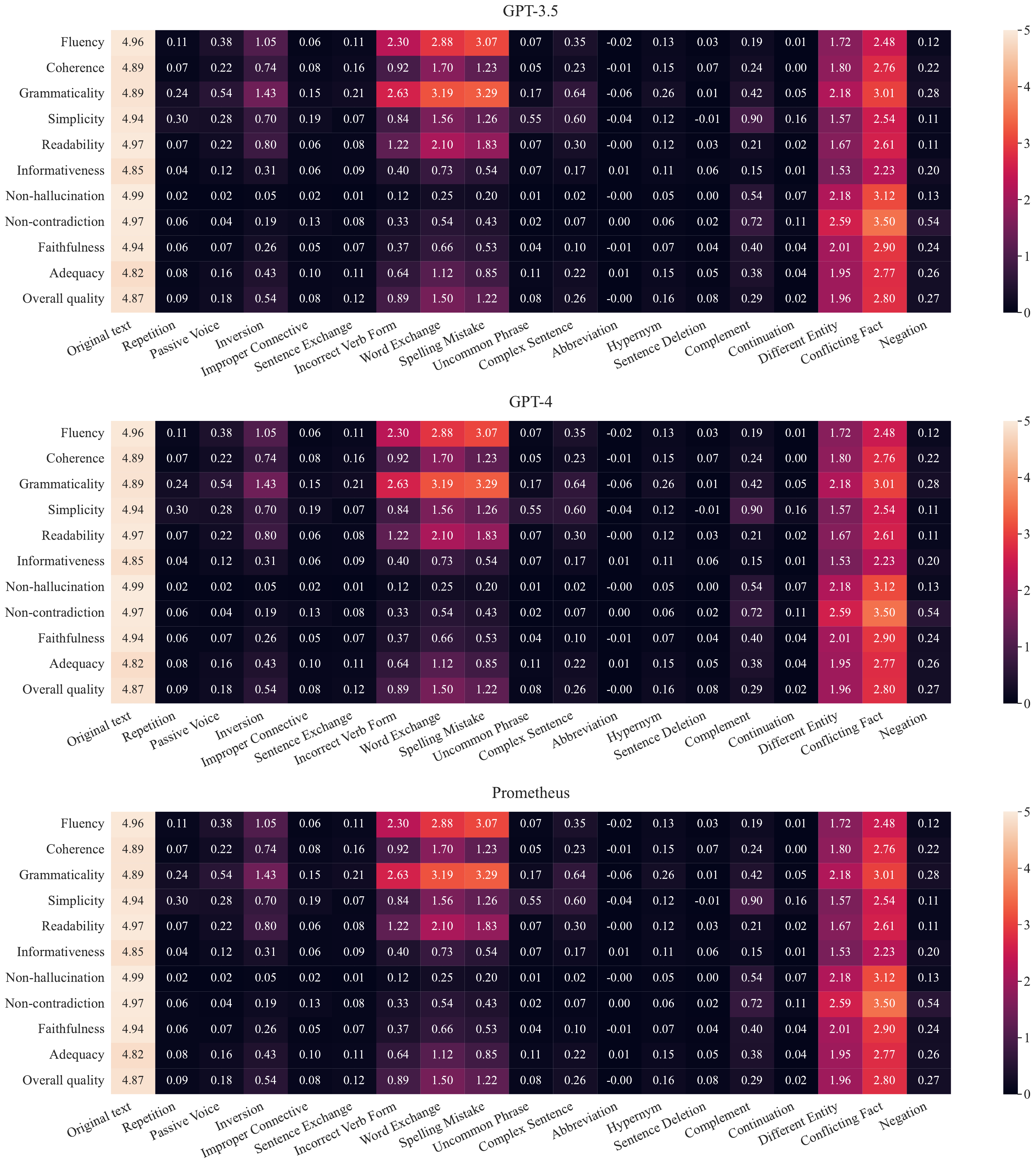}
\caption{The variances of evaluation scores from three LLMs between original texts and different perturbed texts with the description of simplified type.}
\label{fig:12}
\end{figure*}

\begin{figure*}[!htp]
\centering
\includegraphics[width=\textwidth]{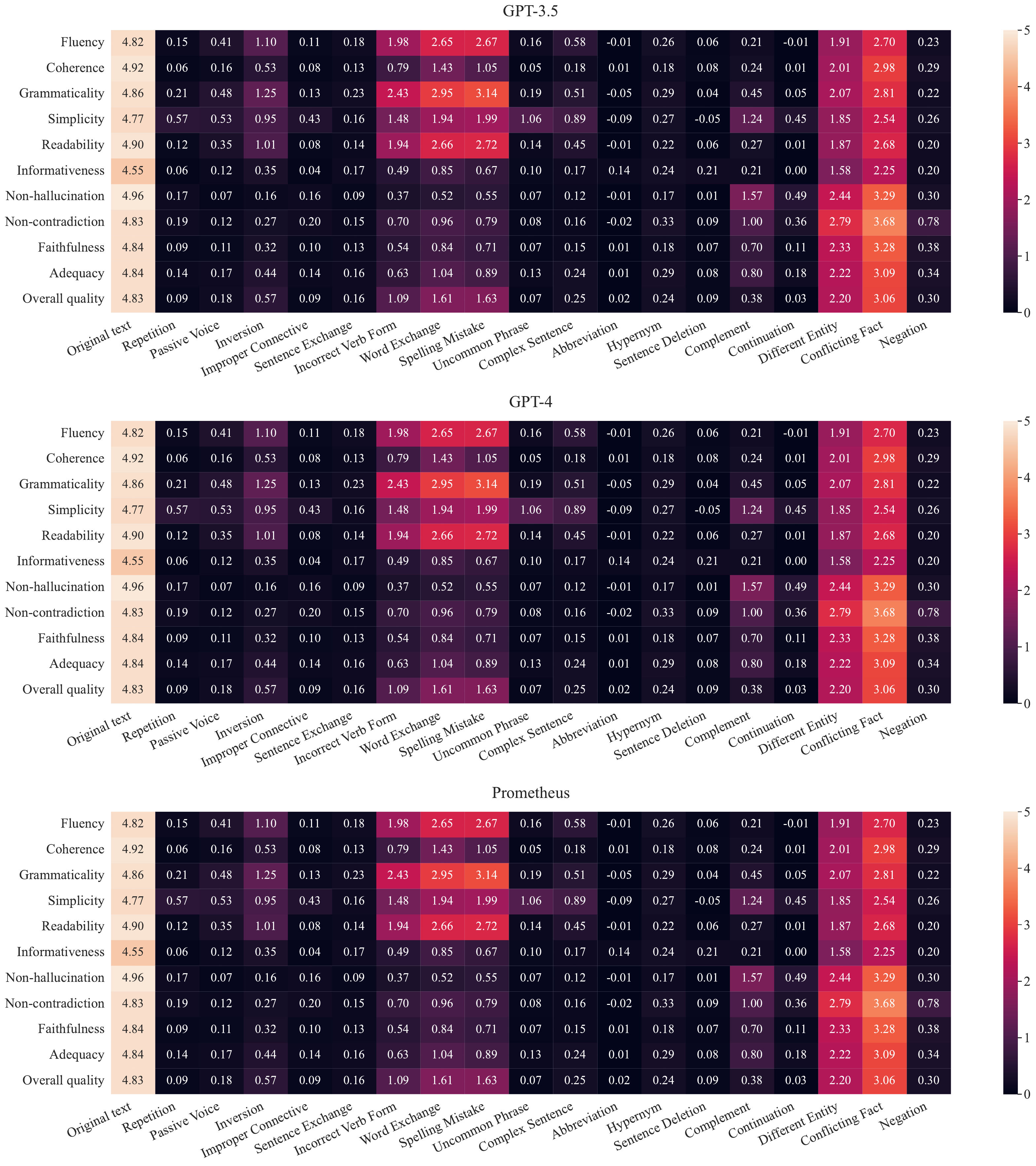}
\caption{The variances of evaluation scores from three LLMs between original texts and different perturbed texts with the description of list type.}
\label{fig:13}
\end{figure*}

\begin{figure*}[!htp]
\centering
\includegraphics[width=\textwidth]{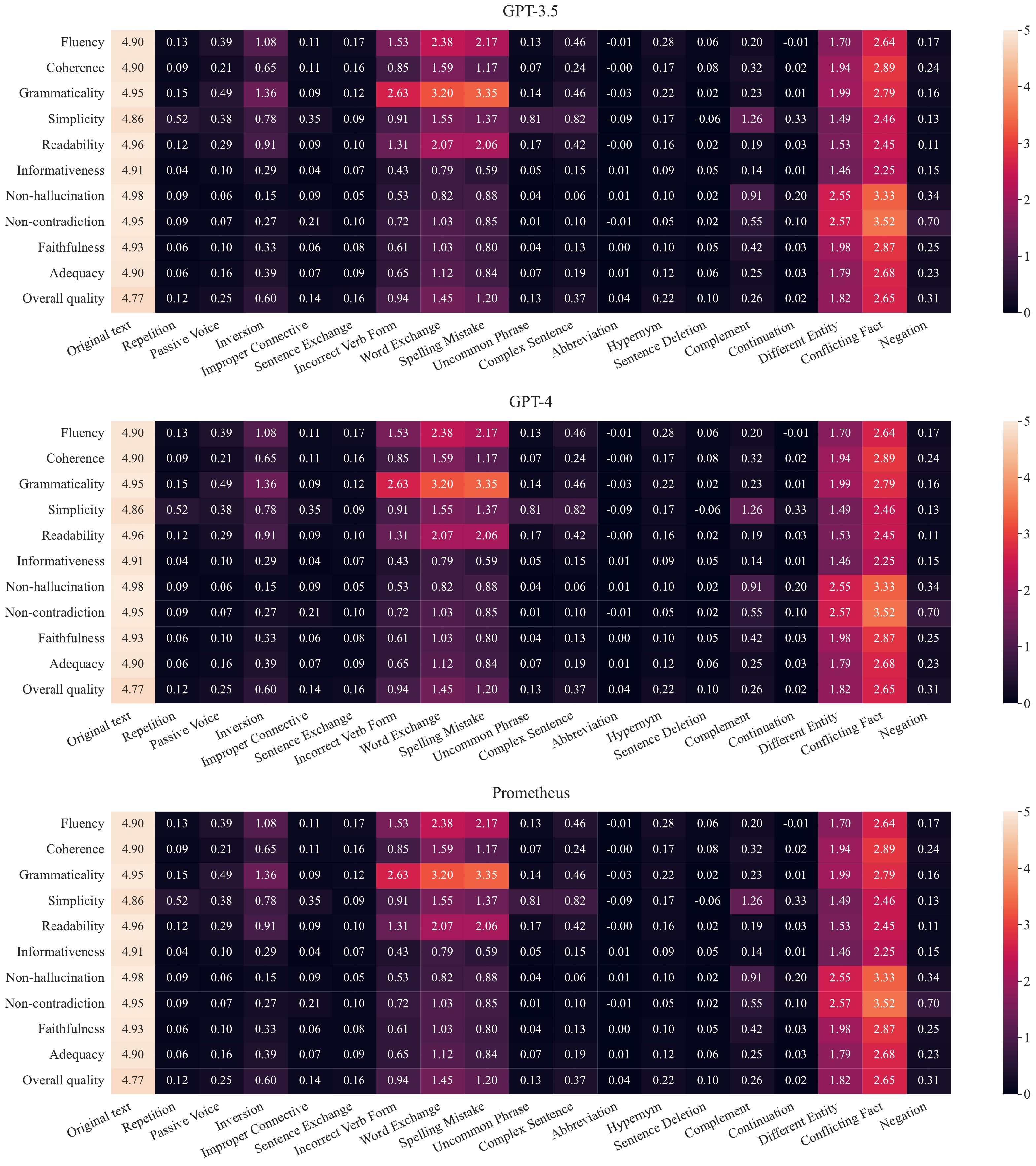}
\caption{The variances of evaluation scores from three LLMs between original texts and different perturbed texts with the description of term type.}
\label{fig:14}
\end{figure*}

\begin{figure*}[!htp]
  \centering
  \includegraphics[width=\textwidth]{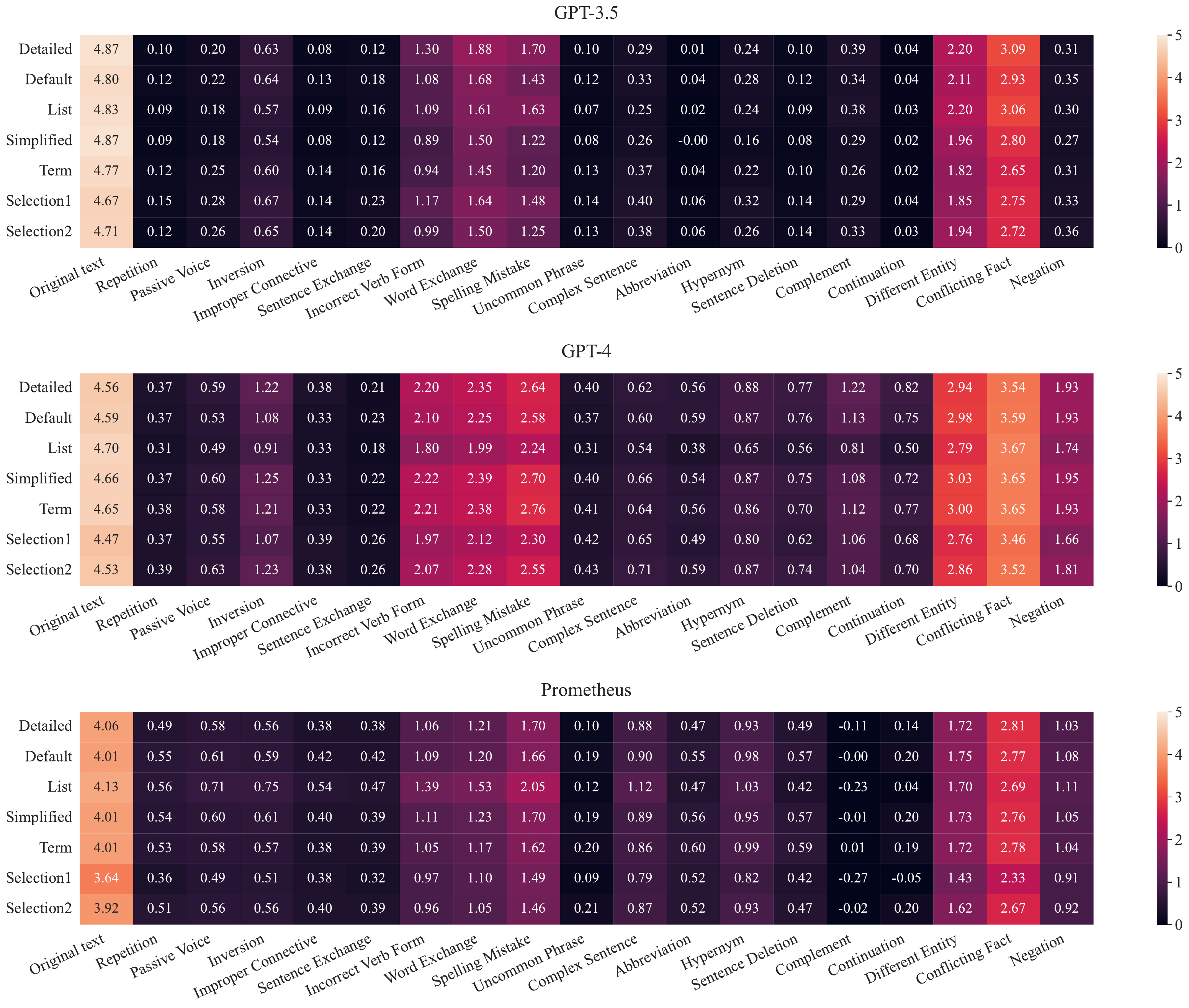}
  \caption{The variances of evaluation scores from three LLMs between original texts and different perturbed texts on Overall quality.}
  \label{fig:25}
  \end{figure*}

\begin{figure*}[!htp]
  \centering
  \includegraphics[width=\textwidth]{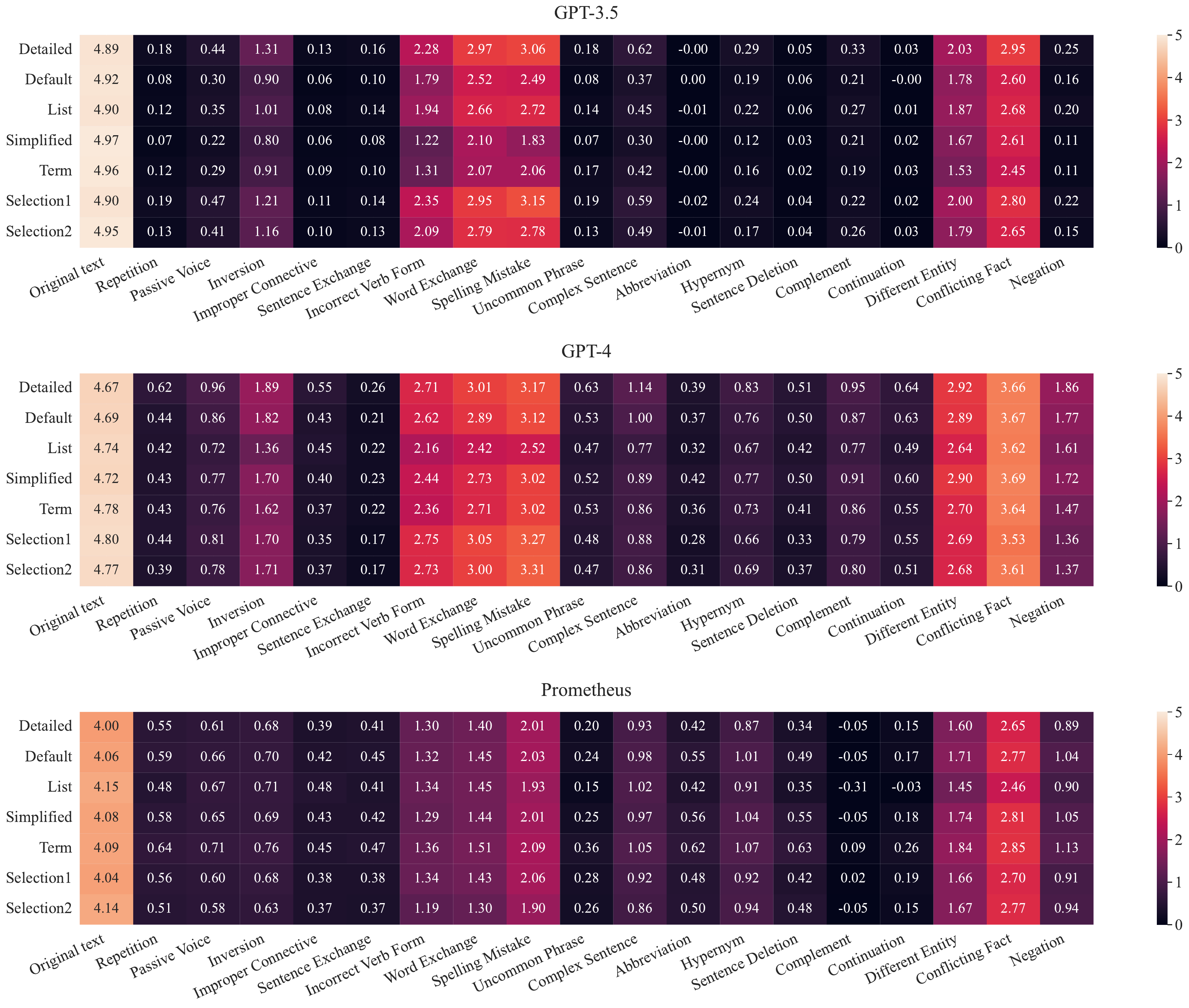}
  \caption{The variances of evaluation scores from three LLMs between original texts and different perturbed texts on Readability.}
  \label{fig:19}
\end{figure*}

\begin{figure*}[!htp]
\centering
\includegraphics[width=\textwidth]{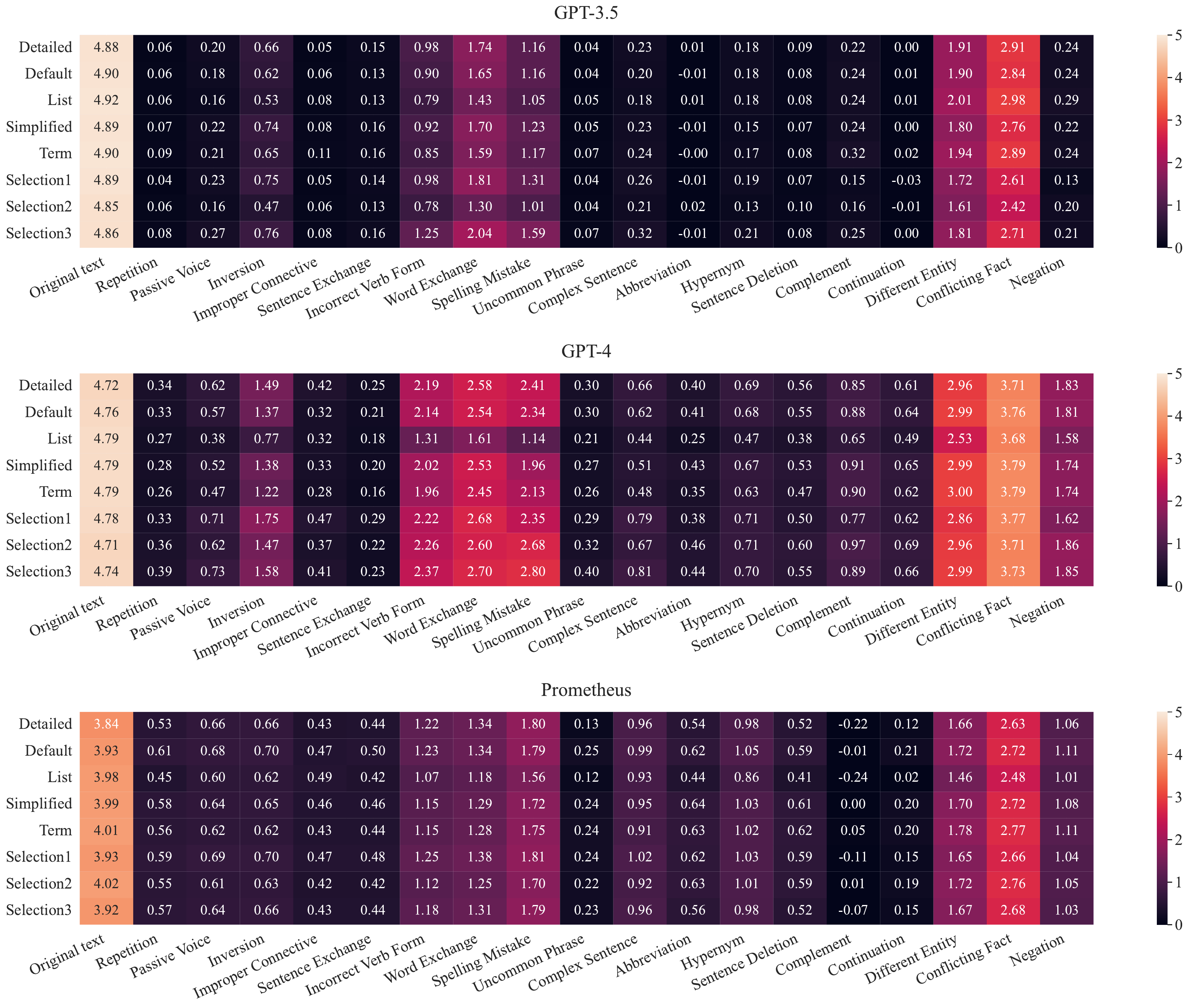}
\caption{The variances of evaluation scores from three LLMs between original texts and different perturbed texts on Coherence.}
\label{fig:16}
\end{figure*}

\begin{figure*}[!htp]
  \centering
  \includegraphics[width=\textwidth]{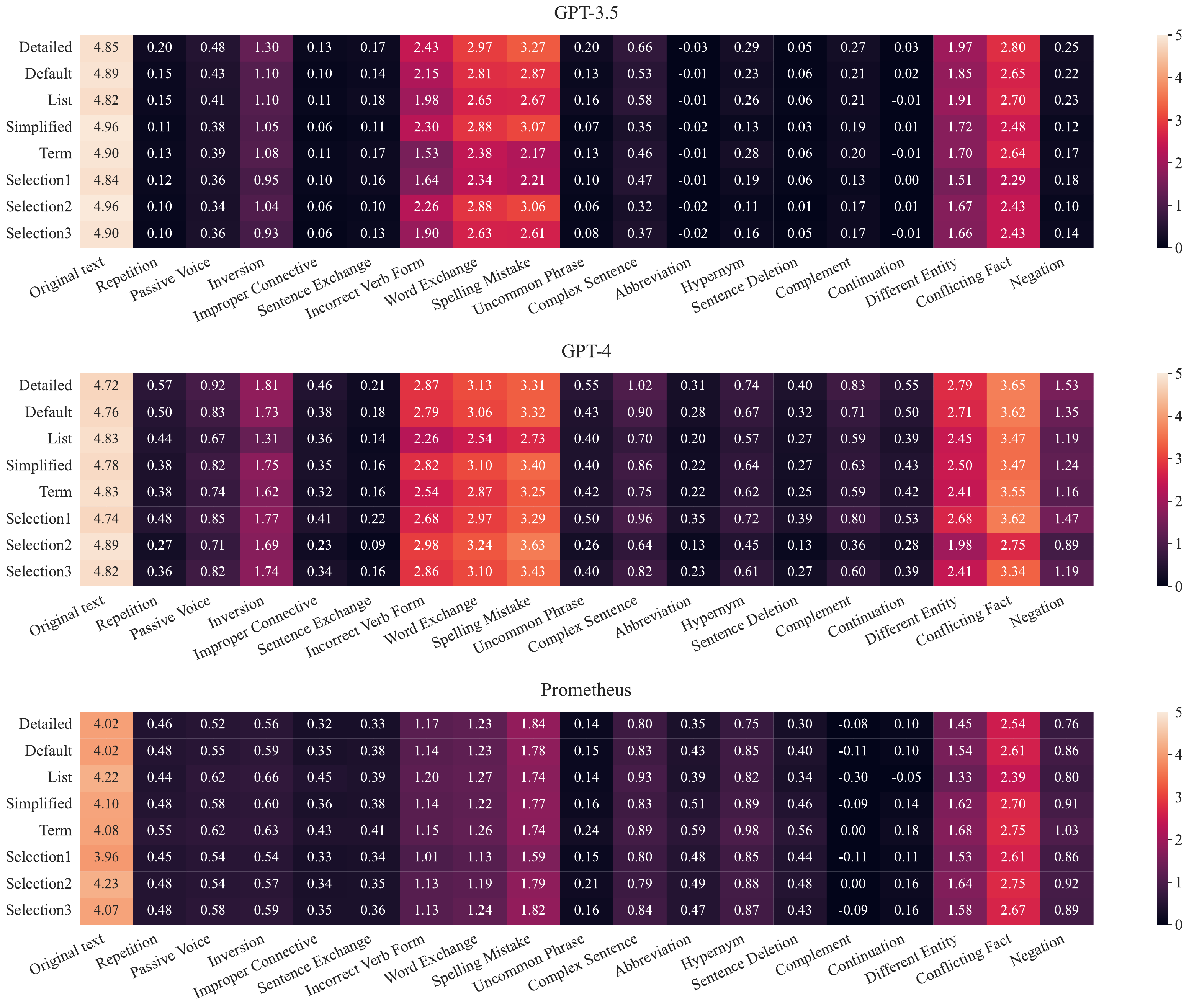}
  \caption{The variances of evaluation scores from three LLMs between original texts and different perturbed texts on Fluency.}
  \label{fig:15}
\end{figure*}

\begin{figure*}[!htp]
\centering
\includegraphics[width=\textwidth]{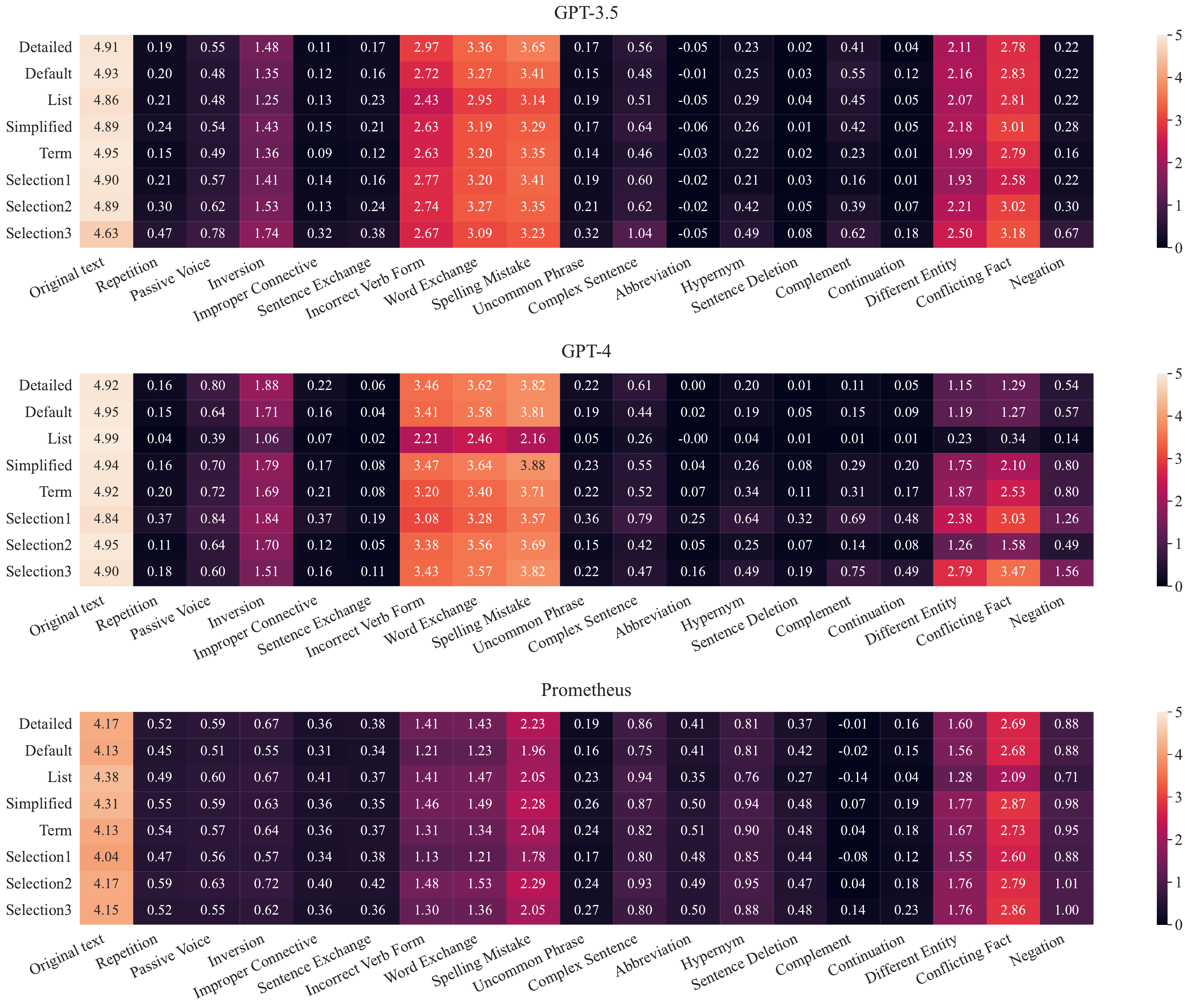}
\caption{The variances of evaluation scores from three LLMs between original texts and different perturbed texts on Grammaticality.}
\label{fig:17}
\end{figure*}

\begin{figure*}[!htp]
\centering
\includegraphics[width=\textwidth]{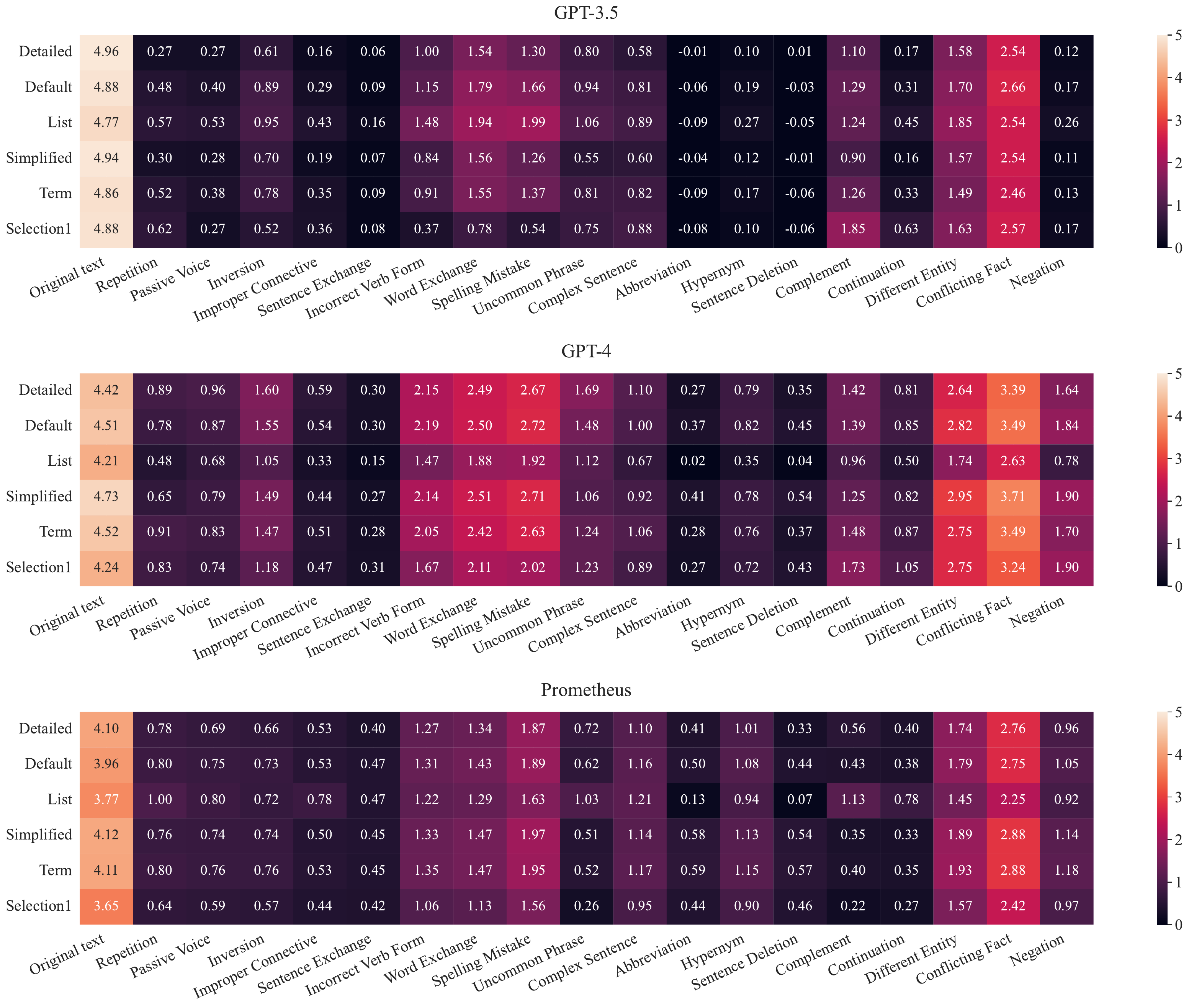}
\caption{The variances of evaluation scores from three LLMs between original texts and different perturbed texts on Simplicity.}
\label{fig:18}
\end{figure*}

\begin{figure*}[!htp]
  \centering
  \includegraphics[width=\textwidth]{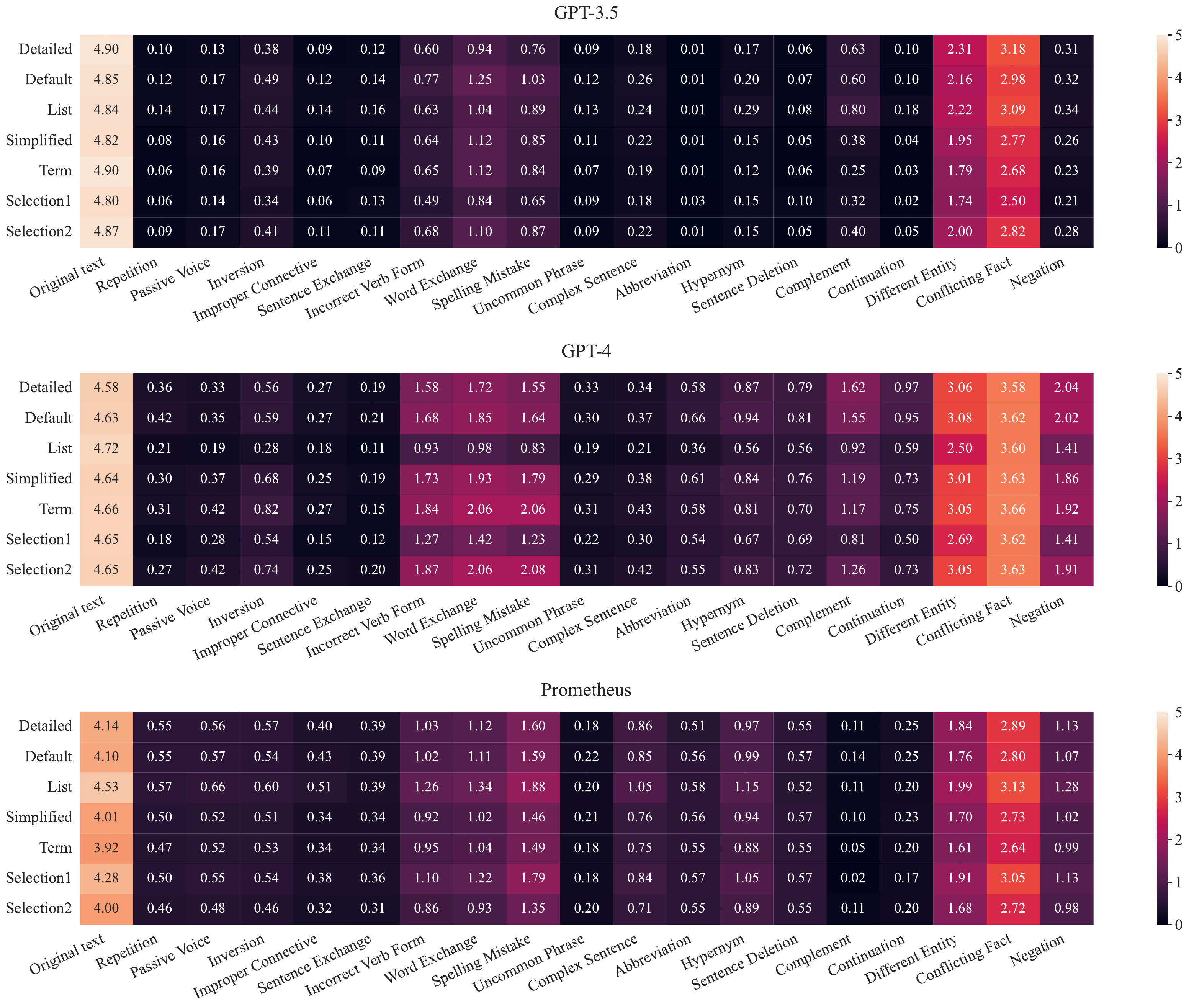}
  \caption{The variances of evaluation scores from three LLMs between original texts and different perturbed texts on Adequacy.}
  \label{fig:24}
  \end{figure*}

\begin{figure*}[!htp]
  \centering
  \includegraphics[width=\textwidth]{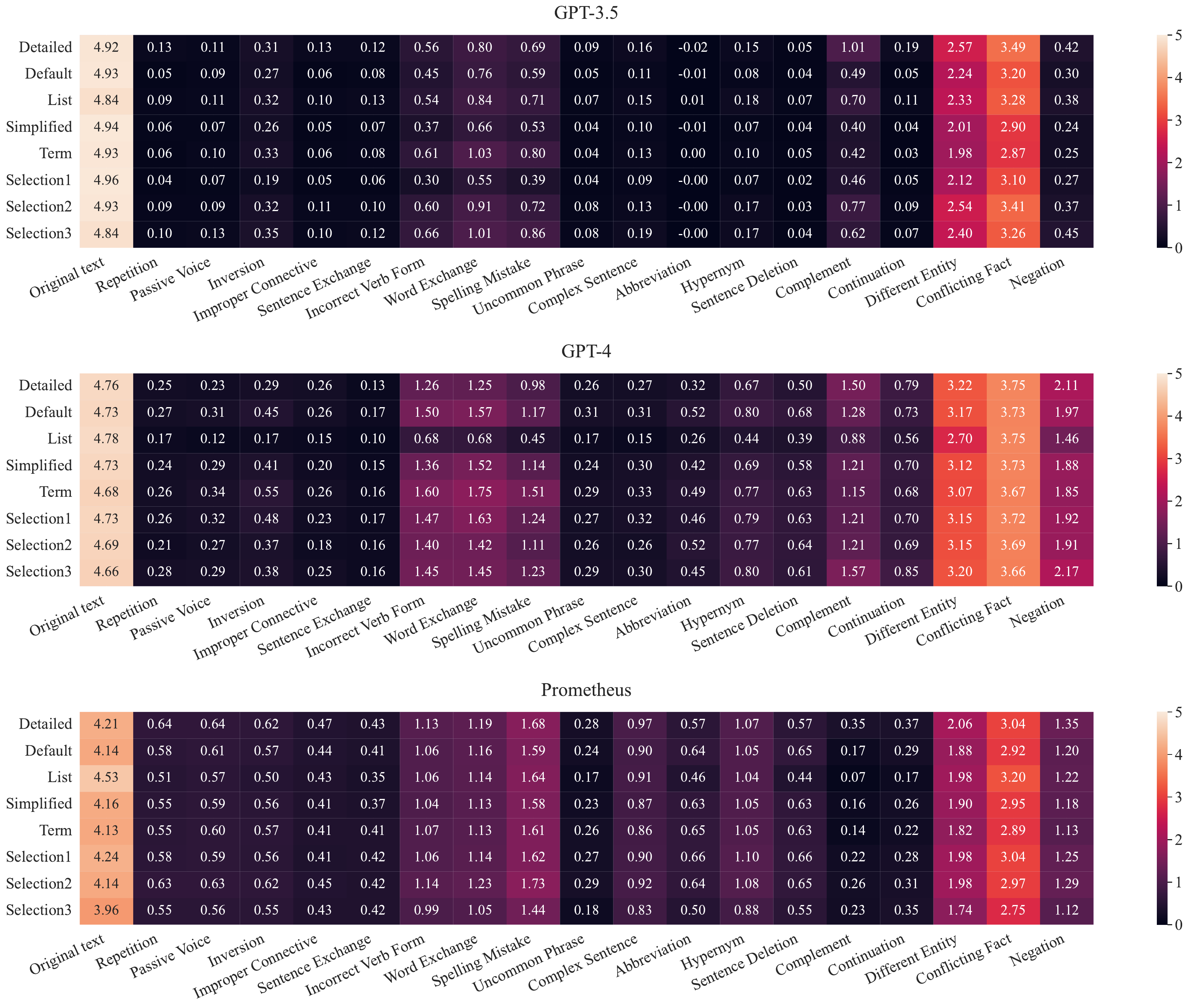}
  \caption{The variances of evaluation scores from three LLMs between original texts and different perturbed texts on Faithfulness.}
  \label{fig:23}
  \end{figure*}

\begin{figure*}[!htp]
  \centering
  \includegraphics[width=\textwidth]{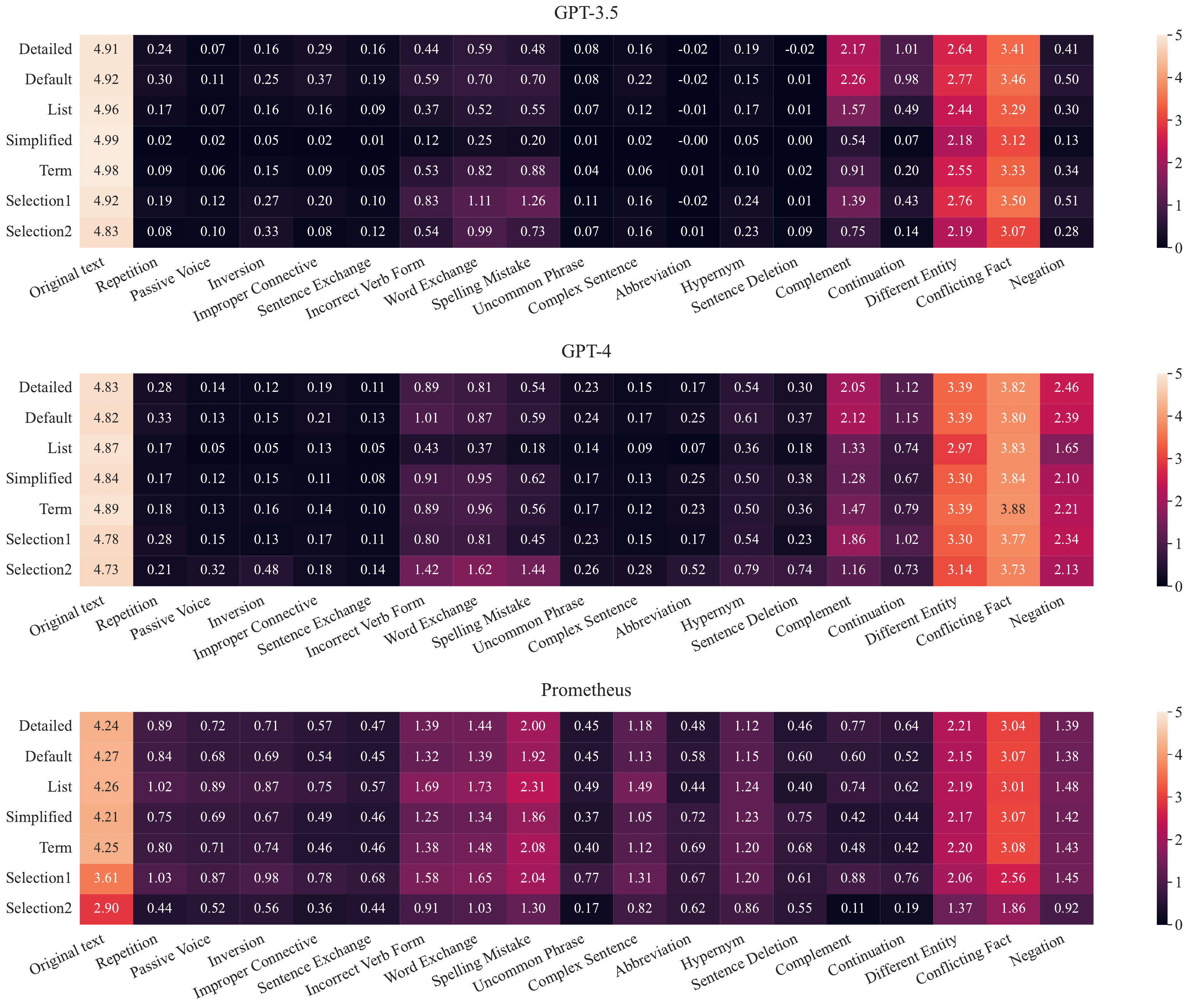}
  \caption{The variances of evaluation scores from three LLMs between original texts and different perturbed texts on Non-hallucination.}
  \label{fig:22}
  \end{figure*}

\begin{figure*}[!htp]
\centering
\includegraphics[width=\textwidth]{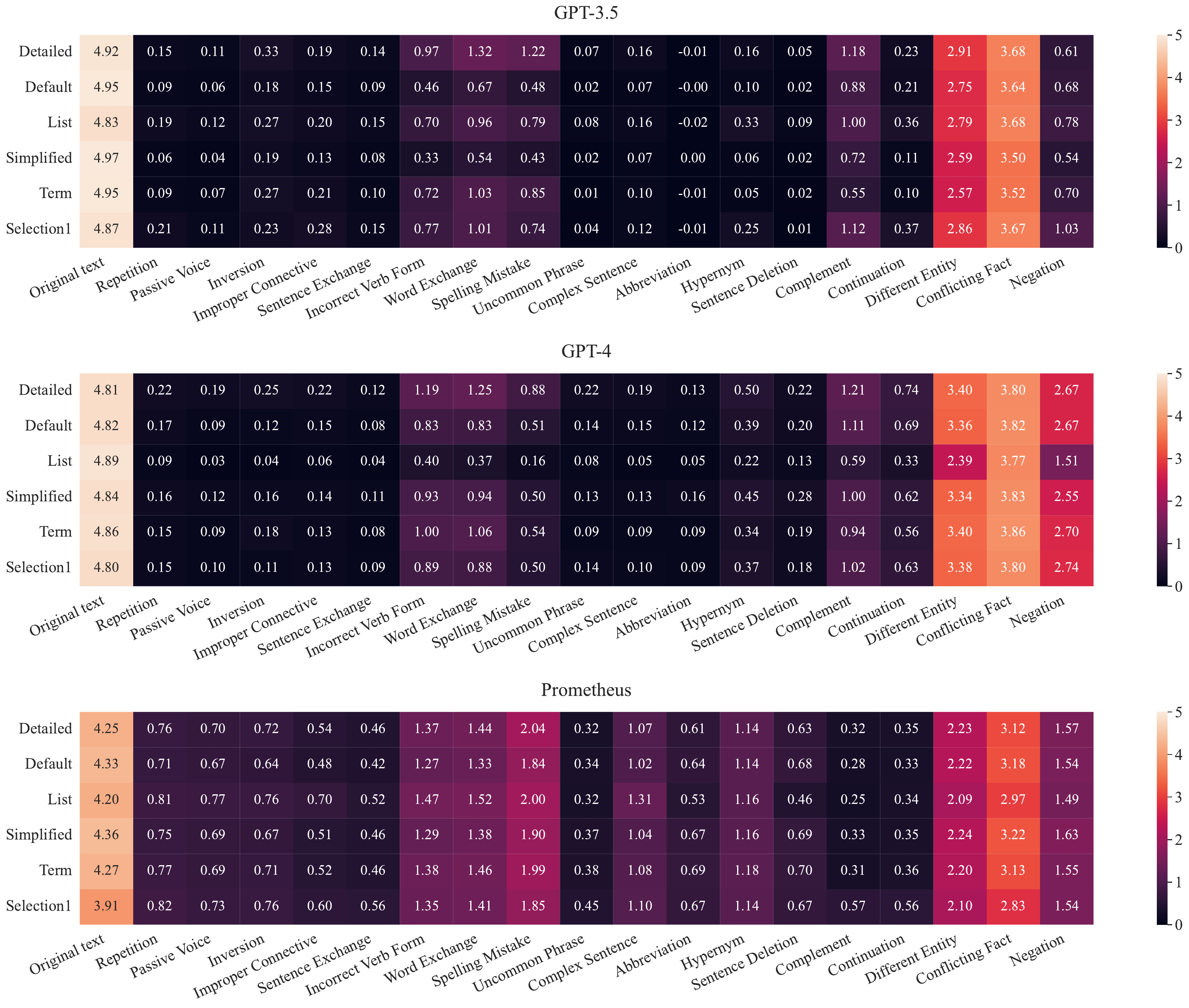}
\caption{The variances of evaluation scores from three LLMs between original texts and different perturbed texts on Non-contradiction.}
\label{fig:21}
\end{figure*}

\begin{figure*}[!htp]
  \centering
  \includegraphics[width=\textwidth]{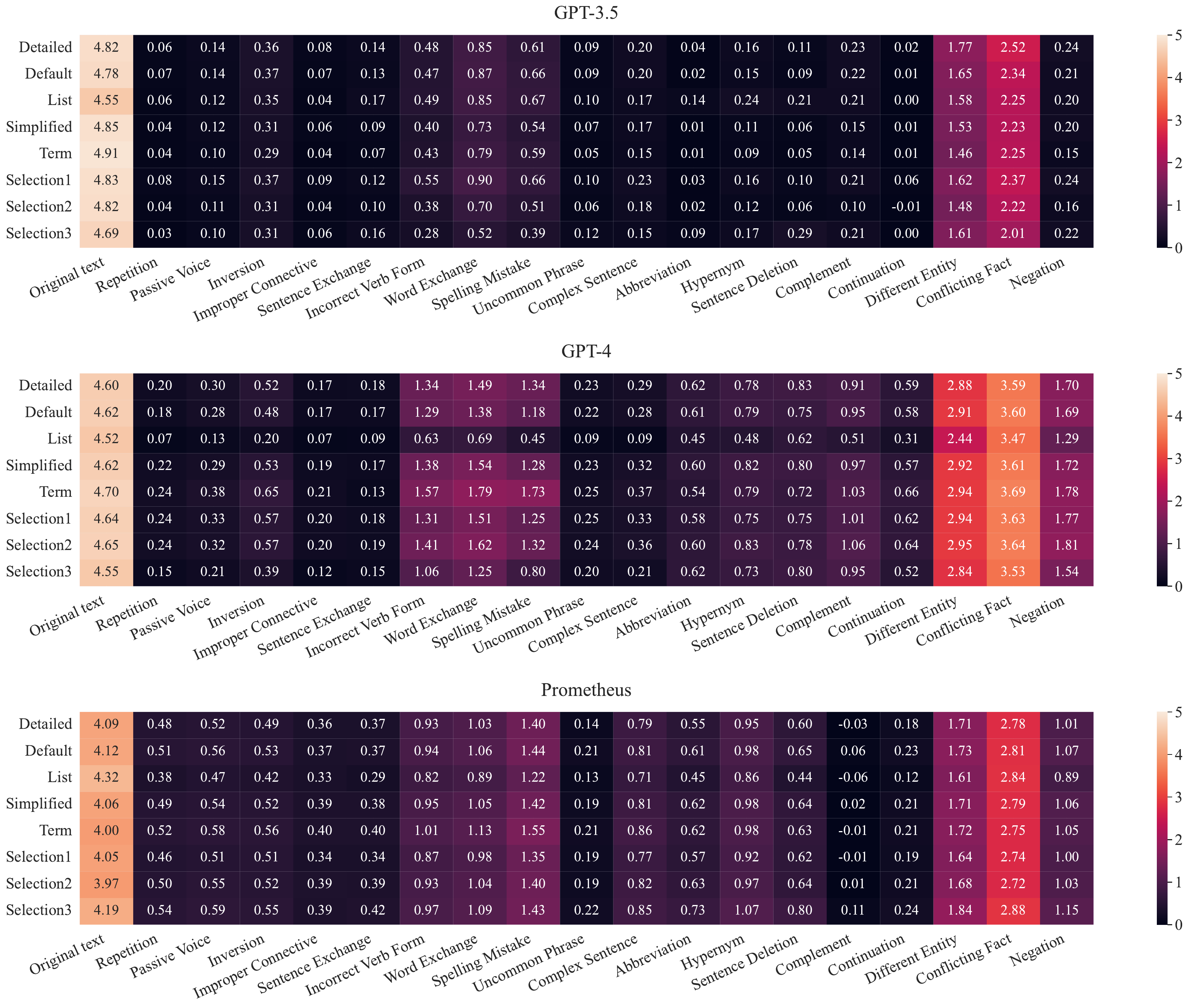}
  \caption{The variances of evaluation scores from three LLMs between original texts and different perturbed texts on Informativeness.}
  \label{fig:20}
  \end{figure*}

\paragraph{Complete perturbation results.} We display all the results of perturbation attacks on GPT-3.5, GPT-4, and Prometheus here. The results are visualized for fixed evaluation aspects or description types separately. Figure \ref{fig:10}-\ref{fig:14} show the perturbation results of the three LLMs given a description type, which makes it convenient to compare different evaluation aspects. On the other hand, Figure \ref{fig:25}-\ref{fig:20} show the perturbation results of the three LLMs given an evaluation aspect, which allows for easier comparison of different description types.

\begin{figure*}[!htp]
  \centering
  \begin{minipage}{\textwidth}
    \includegraphics[width=\textwidth]{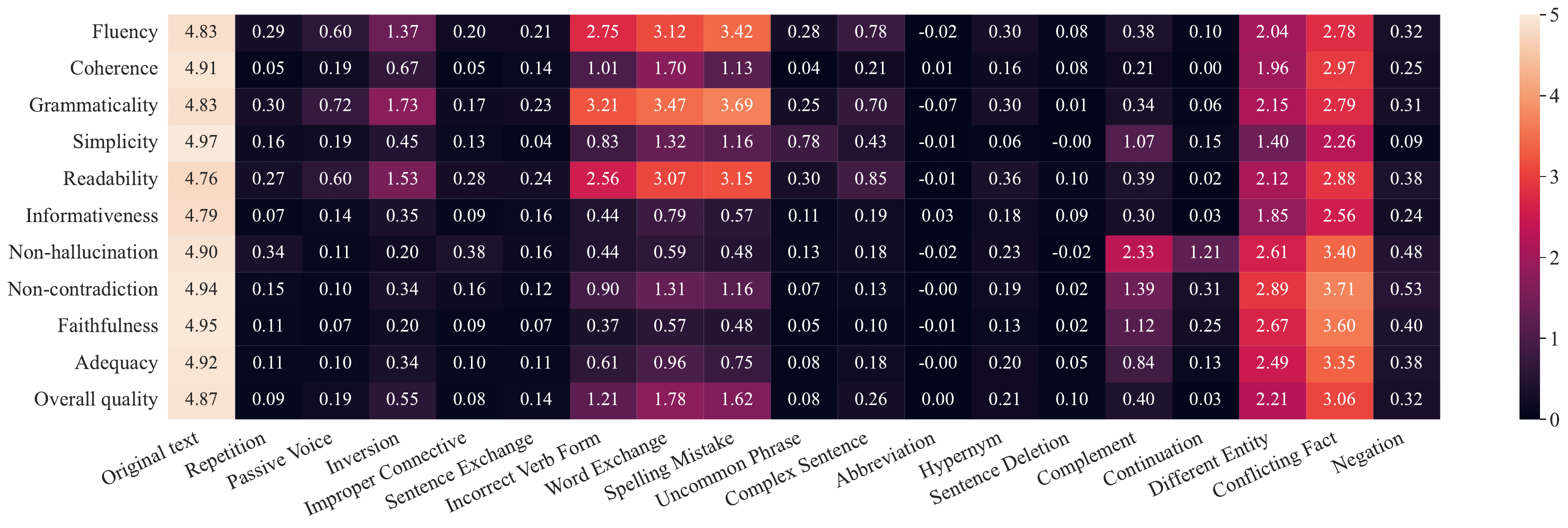}
    \caption{Results of perturbation attacks for the criteria that only retain descriptions.}
    \label{fig:63}
  \end{minipage}
  \vspace{5cm}
  \begin{minipage}{\textwidth}
    \includegraphics[width=\textwidth]{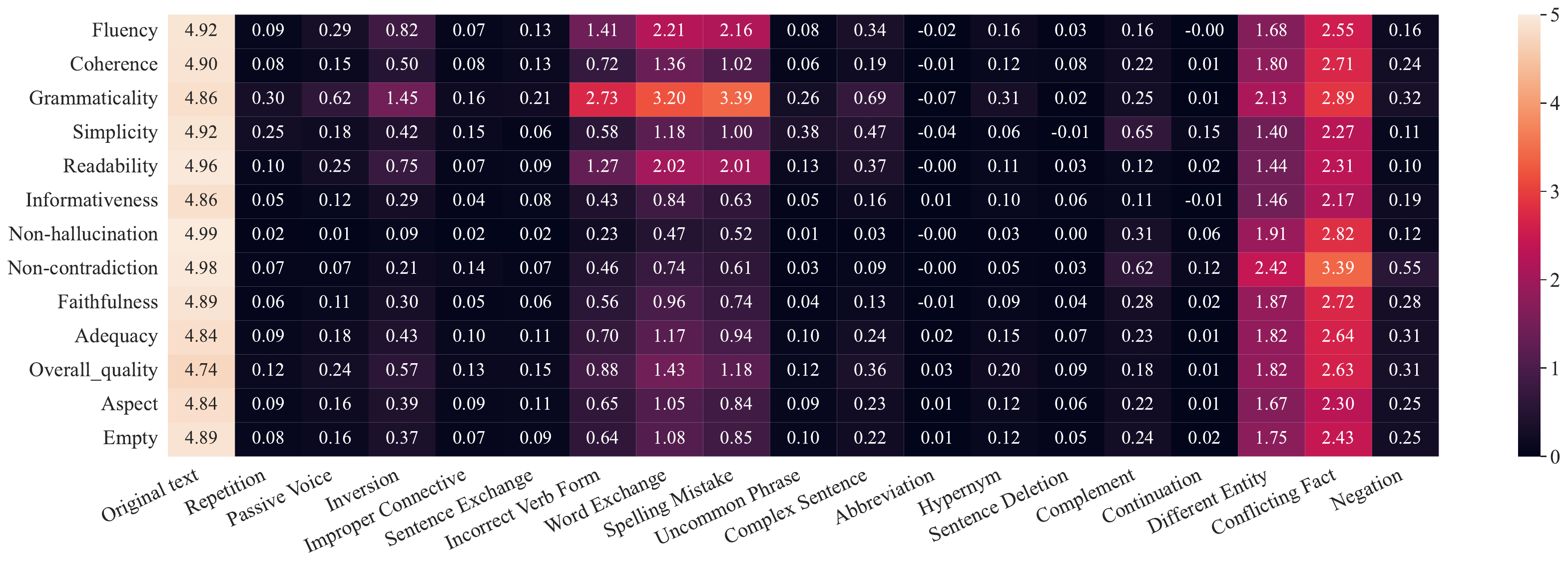}
    \caption{Results of perturbation attacks for the criteria that only retain terms, including the empty criterion and the meaningless criterion with a single word of "Aspect".}
    \label{fig:64}
  \end{minipage}
\end{figure*}

\paragraph{Deleting terms or descriptions.} Figure \ref{fig:63} and \ref{fig:64} show the perturbation results for the criteria that only retain descriptions, only retain terms, only contain a single word of "Aspect", and are empty.

\begin{table*}[]
  \centering
  \setlength{\tabcolsep}{6pt}
  \begin{tabular}{lccccc}
  \toprule
   Description Type & Annotation Consistency & Vote\_vote & Vote\_all \\ 
  \midrule
  Default & 0.8791 & 0.8586 & 0.8990 \\ 
  Simplified & 0.8472 & 0.8081 & 0.8586 \\
  Detailed & \textbf{0.9012} & \textbf{0.8788} & \textbf{0.9444} \\
  Term & 0.7181 & 0.7172 & 0.7677 \\
  List & 0.8602 & 0.8333 & 0.9091 \\
  \bottomrule
  \end{tabular}
  \caption{Annotation consistency and match rate of human annotations on five types of descriptions designed by us. \textit{vote\_vote} means that the results of 4 annotators are first taken as plurality on a single sample, and then 4 samples are taken as plurality as the final result of human annotation on a perturbation. \textit{vote\_all} means that 16 results are directly taken as plurality together as the final result.}
  \label{tab:human_designed_result}
\end{table*}

\begin{table*}[]
\centering
\small
\begin{tabular}{lccccccc}
\toprule
Perturbation & GPT-4 (Overall) & BLEU & chrF++ & ROUGE & COMET-QE & BERTScore & BLEURT \\

\midrule
Repetition & 0.37 & 1.05 & 0.20 & 0.39 & 0.00 & 0.08 & 0.60 \\
Passive Voice & 0.59 & 1.68 & 0.25 & 1.02 & 0.13 & 0.13 & 0.73 \\
Inversion & 1.22 & 1.51 & 0.17 & 0.98 & 0.33 & 0.16 & 1.06 \\
Improper Connective & 0.38 & 0.71 & 0.13 & 0.13 & 0.05 & 0.06 & 0.44 \\
Sentence Exchange & 0.21 & 0.16 & 0.01 & 1.46 & 0.08 & 0.11 & 0.24 \\
Incorrect Verb Form & 2.20 & 1.11 & 0.14 & 0.49 & 0.18 & 0.10 & 0.58 \\
Word Exchange & 2.35 & 1.35 & 0.12 & 0.46 & 0.43 & 0.16 & 1.31 \\
Spelling Mistake & 2.64 & 1.53 & 0.13 & 0.73 & 0.50 & 0.26 & 1.18 \\
Uncommon Phrase & 0.40 & 1.52 & 0.43 & 0.73 & 0.10 & 0.12 & 0.81 \\
Complex Sentence & 0.62 & 2.03 & 0.30 & 0.85 & 0.15 & 0.16 & 0.80 \\
Abbreviation & 0.56 & 2.64 & 1.21 & 1.43 & 0.20 & 0.20 & 1.05 \\
Hypernym & 0.88 & 1.44 & 0.59 & 0.78 & 0.11 & 0.15 & 0.99 \\
Sentence Deletion & 0.77 & 1.28 & 0.94 & 0.66 & 0.12 & 0.11 & 0.74 \\
Complement & 1.22 & 2.03 & 0.59 & 1.06 & -0.28 & 0.20 & 0.90 \\
Continuation & 0.82 & 1.27 & 0.40 & 0.77 & -0.19 & 0.12 & 0.54 \\
Different Entity & 2.94 & 1.11 & 0.42 & 0.54 & 0.09 & 0.09 & 0.93 \\
Conflicting Fact & 3.54 & 1.83 & 0.50 & 1.04 & 0.20 & 0.19 & 1.21 \\
Negation & 1.93 & 0.37 & 0.07 & 0.14 & 0.08 & 0.04 & 0.52 \\
\bottomrule
\end{tabular}
\caption{The variances of scores of non-LLM evaluation metrics bewteen original texts and different perturbed texts. Scores are scaled to 1-5, and the higher the better. }
\label{tab:non-LLM results}
\end{table*}

\end{document}